# PATHWISE COORDINATE OPTIMIZATION FOR SPARSE LEARNING: ALGORITHM AND THEORY[*]

By Tuo Zhao[†], Han Liu[‡] and Tong Zhang[§]

*Georgia Tech[†], Princeton University[‡], Tencent AI Lab[§]*

The pathwise coordinate optimization is one of the most important computational frameworks for high dimensional convex and nonconvex sparse learning problems. It differs from the classical coordinate optimization algorithms in three salient features: *warm start initialization*, *active set updating*, and *strong rule for coordinate preselection*. Such a complex algorithmic structure grants superior empirical performance, but also poses significant challenge to theoretical analysis. To tackle this long lasting problem, we develop a new theory showing that these three features play pivotal roles in guaranteeing the outstanding statistical and computational performance of the pathwise coordinate optimization framework. Particularly, we analyze the existing pathwise coordinate optimization algorithms and provide new theoretical insights into them. The obtained insights further motivate the development of several modifications to improve the pathwise coordinate optimization framework, which guarantees linear convergence to a unique sparse local optimum with optimal statistical properties in parameter estimation and support recovery. This is the first result on the computational and statistical guarantees of the pathwise coordinate optimization framework in high dimensions. Thorough numerical experiments are provided to support our theory.

**1. Introduction.** Modern data acquisition routinely produces massive amount of high dimensional data, where the number of variables $d$ greatly exceeds the sample size $n$, such as high-throughput genomic data (Neale et al., 2012) and image data from functional Magnetic Resonance Imaging (Eloyan et al., 2012; Liu et al., 2015). To handle high dimensionality, we often assume that only a small subset of variables are relevant in modeling (Tibshirani, 1996). Such a parsimonious assumption motivates various sparse learning approaches. Taking sparse linear regression as an example, we consider a linear model $y = X\theta^* + \varepsilon$, where $y \in \mathbb{R}^n$ is the response vector, $X \in \mathbb{R}^{n \times d}$ is the design matrix, $\theta^* = (\theta_1, ..., \theta_d)^\top \in \mathbb{R}^d$ is the unknown

---

sparse regression coefficient vector, and $\varepsilon \sim N(0, \sigma^2 I_n)$ is the random noise. Here $I_n \in \mathbb{R}^{n \times n}$ is an identity matrix. Let $\|\cdot\|_2$ denote the $\ell_2$ norm, and $\mathcal{R}_\lambda(\theta)$ denote a sparsity-inducing regularizer with a regularization parameter $\lambda > 0$. We can obtain a sparse estimator of $\theta^*$ by solving the following regularized least square optimization problem

$$(1.1) \qquad \min_{\theta \in \mathbb{R}^d} \mathcal{F}_\lambda(\theta), \text{ where } \mathcal{F}_\lambda(\theta) = \frac{1}{2n}\|y - X\theta\|_2^2 + \mathcal{R}_\lambda(\theta).$$

Popular choices of $\mathcal{R}_\lambda(\theta)$ are usually coordinate decomposable, $\mathcal{R}_\lambda(\theta) = \sum_{j=1}^d r_\lambda(\theta_j)$, including the $\ell_1$ (Lasso, Tibshirani (1996)), SCAD (Smooth Clipped Absolute Deviation, Fan and Li (2001)), and MCP (Minimax Concavity Penalty, Zhang (2010)) regularizers. For example, the $\ell_1$ regularizer takes $\mathcal{R}_\lambda(\theta) = \lambda\|\theta\|_1 = \lambda\sum_j |\theta_j|$ with $r_\lambda(|\theta_j|) = \lambda|\theta_j|$ for $j = 1, ..., d$.

The $\ell_1$ regularizer is convex and computationally tractable, but often induces large estimation bias, and requires a restrictive irrepresentable condition to attain variable selection consistency (Zhao and Yu, 2006; Meinshausen and Bühlmann, 2006; Zou, 2006). To address this issue, nonconvex regularizers such as SCAD and MCP have been proposed to obtain nearly unbiased estimators. Throughout the rest of the paper, we only consider MCP as an example due to space limit, but the extension to SCAD is straightforward. Particularly, let $\mathcal{E}$ be an event, we define $\mathbb{1}_{\{\mathcal{E}\}}$ as an indicator function with $\mathbb{1}_{\{\mathcal{E}\}} = 1$ if $\mathcal{E}$ holds and $\mathbb{1}_{\{\mathcal{E}\}} = 0$ otherwise. Given $\gamma > 1$, MCP has

$$(1.2) \qquad r_\lambda(|\theta_j|) = \lambda\left(|\theta_j| - \frac{\theta_j^2}{2\lambda\gamma}\right) \cdot \mathbb{1}_{\{|\theta_j| < \lambda\gamma\}} + \frac{\lambda^2\gamma}{2} \cdot \mathbb{1}_{\{|\theta_j| \geq \lambda\gamma\}}.$$

We call $\gamma$ the concavity parameter of MCP, since it essentially characterizes the concavity of the MCP regularizer: A larger $\gamma$ implies that the regularizer is less concave. We observe that the MCP regularizer can be written as

$$(1.3) \qquad \mathcal{R}_\lambda(\theta) = \lambda\|\theta\|_1 + \mathcal{H}_\lambda(\theta),$$

where $\mathcal{H}_\lambda(\theta) = \sum_{j=1}^d h_\lambda(|\theta_j|)$ is a smooth, concave, and also coordinate decomposable function with

$$(1.4) \qquad h_\lambda(|\theta_j|) = -\frac{\theta_j^2}{2\gamma} \cdot \mathbb{1}_{\{|\theta_j| < \lambda\gamma\}} + \frac{\lambda^2\gamma - 2\lambda|\theta_j|}{2} \cdot \mathbb{1}_{\{|\theta_j| \geq \lambda\gamma\}}.$$

We present several examples of the MCP regularizer in Figure 1. Fan and Li (2001); Zhang (2010) show that the nonconvex regularizer effectively reduces the estimation bias, and achieve better performance than the $\ell_1$ regularizer in both parameter estimation and support recovery. Particularly, given a suitable chosen $\gamma < \infty$, they show that there exits a local optimum to (1.1), which attains the oracle properties under much weaker conditions. However, they cannot provide specific algorithms that guarantee such a local optimum in polynomial time due to the nonconvexity.

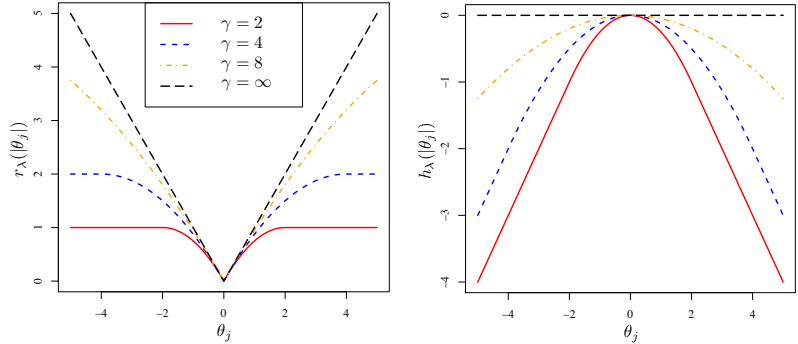

FIG 1. *Several examples of the MCP regularizer with $\lambda = 1$ and $\gamma = 2, 4, 8$, and $\infty$ (Lasso). The MCP regularizer reduces the estimation bias and achieve better performance than the $\ell_1$ regularizer in both parameter estimation and support recovery, but imposes great computational challenge.*

Typical algorithms for solving (1.1) developed in existing optimization literature include proximal gradient algorithms (Nesterov, 2013) and coordinate optimization algorithms (Luo and Tseng, 1992; Shalev-Shwartz and Tewari, 2011; Richtárik and Takáč, 2012; Lu and Xiao, 2015). The proximal gradient algorithms need to access all entries of the design matrix $X$ in each iteration for computing a full gradient and a sophisticated line search step. Thus, they are often not scalable and efficient in practice when $d$ is large. To address this issue, many researchers resort to the coordinate optimization algorithms for better computational efficiency and scalability.

The classical coordinate optimization algorithm is straightforward and much simpler than the proximal gradient algorithms in each iteration: Given $\theta^{(t)}$ at the $t$-th iteration, we select a coordinate $j$, and then take an exact coordinate minimization step

$$\theta_j^{(t+1)} = \underset{\theta_j}{\operatorname{argmin}}\ \mathcal{F}_\lambda(\theta_j, \theta_{\setminus j}^{(t)}), \tag{1.5}$$

where $\theta_{\setminus j}$ is a subvector of $\theta$ with the $j$-th entry removed. For the $\ell_1$, SCAD, and MCP regularizers, (1.5) admits a closed form solution. For notational simplicity, we denote $\theta_j^{(t+1)} = \mathcal{T}_{\lambda,j}(\theta^{(t)})$. Then (1.5) can be rewritten as

$$(1.6) \qquad \theta_j^{(t+1)} = \mathcal{T}_{\lambda,j}(\theta^{(t)}) = \underset{\theta_j}{\operatorname{argmin}} \frac{1}{2n}\|z^{(t)} - X_{*j}\theta_j\|_2^2 + r_\lambda(\theta_j),$$

where $X_{*j}$ denotes the $j$-th column of $X$ and $z^{(t)} = y - X\theta^{(t)} + X_{*j}\theta_j^{(t)}$ is the partial residual. Without loss of generality, we assume that $X$ satisfies the column normalization condition $\|X_{*j}\|_2 = \sqrt{n}$ for all $j = 1,...,d$. Let $\widetilde{\theta}_j^{(t)} = \frac{1}{n}X_{*j}^\top z^{(t)}$. Then for MCP, we obtain $\theta_j^{(t+1)}$ by

$$(1.7) \qquad \theta_j^{(t+1)} = \widetilde{\theta}_j^{(t)} \cdot \mathbb{1}_{\{|\widetilde{\theta}_j^{(t)}| \geq \gamma\lambda\}} + \frac{\mathcal{S}_\lambda(\widetilde{\theta}_j^{(t)})}{1 - 1/\gamma} \cdot \mathbb{1}_{\{|\widetilde{\theta}_j^{(t)}| < \gamma\lambda\}},$$

where $\mathcal{S}_\lambda(a) = \operatorname{sign}(a) \cdot \max\{|a| - \lambda, 0\}$. As shown in Appendix A, (1.7) can be efficiently calculated by a simple partial residual update trick, which only requires the access to one single column of the design matrix $X_{*j}$ (Recall the proximal gradient algorithms need to access the entire design matrix). Once we obtain $\theta_j^{(t+1)}$, we take $\theta_{\setminus j}^{(t+1)} = \theta_{\setminus j}^{(t)}$. Such a coordinate optimization algorithm, though simple, is not necessarily efficient in theory and practice. Existing optimization theory only shows its sublinear convergence to local optima in high dimensions if we select coordinates from 1 to $d$ in a cyclic order throughout all iterations (Razaviyayn et al., 2013; Li et al., 2016). Moreover, no theoretical guarantee has been established on statistical properties of the obtained estimators for nonconvex regularizers in parameter estimation and support recovery. Thus, the coordinate optimization algorithms were almost neglected until recent rediscovery by Friedman et al. (2007); Mazumder et al. (2011); Tibshirani et al. (2012).[1]

**Remark 1.1** (Connection between MCP and Lasso). Let $\frac{c}{\infty} = 0$ for any constant $c$. As can be seen from (1.2), for $\gamma = \infty$, MCP is reduced to the $\ell_1$ regularizer, i.e., $r_\lambda(|\theta_j|) = \lambda|\theta_j|$ with $h_\lambda(|\theta_j|) = 0$. Accordingly, (1.7) is reduced to $\theta_j^{(t+1)} = \mathcal{S}_\lambda(\widetilde{\theta}_j^{(t)})$, which is identical to the updating formula of the coordinate optimization algorithm proposed in Fu (1998) for Lasso. Thus, throughout the rest of the paper, we just simply consider the $\ell_1$ regularizer as a special case of MCP, unless we clearly specify the difference between $\gamma < \infty$ and $\gamma = \infty$ for MCP.

As illustrated in Figure 2, Friedman et al. (2010); Mazumder et al. (2011); Tibshirani et al. (2012) propose a pathwise coordinate optimization framework with three nested loops, which integrates the warm start initialization, active set updating strategy, and strong rule for coordinate preselection into the classical coordinate optimization.

---

[1] A brief history on applying coordinate optimization to sparse learning problems is presented in Hastie (2009).

Particularly, in the *outer loop*, the warm start initialization optimizes (1.1) with a sequence of decreasing regularization parameters in a multistage manner, and yields solutions from sparse to dense. Within each stage of the warm start initialization (an iteration of the outer loop), the algorithm uses the solution from the previous stage for initialization, and then adopts the active set updating strategy to exploit the solution sparsity to speed up computation. The active set updating strategy contains two consequent nested loops: In the *middle loop*, the algorithm first divides all coordinates into active ones (active set) and inactive ones (inactive set) based on some heuristic coordinate gradient thresholding rule (strong rule, Tibshirani et al. (2012)). Then within each iteration of the middle loop, an *inner loop* is called to conduct coordinate optimization. In general, the algorithm runs an inner loop on the current active coordinates until convergence, with all inactive coordinates remain zero. The algorithm then exploits some heuristic rule to identify a new active set, which further decreases the objective value and repeats the inner loops. The iteration within each stage terminates when the active set in the middle loop no longer changes. In practice, the warm start initialization, active set updating strategy, and strong rule for coordinate preselection encourage the algorithm to iterate over a small active set involving only a small number of coordinates, and therefore significantly boost the computational efficiency and scalability. Software packages such as `GLMNET`, `SparseNet`, and `HUGE` have been developed and widely applied to many research areas (Friedman et al., 2010; Zhao et al., 2012).

Despite of the popularity of the pathwise coordinate optimization framework, we are still in lack of adequate theory to justify its superior computational performance due to its complex algorithmic structure. The warm start initialization, active set updating strategy, and strong rule for coordinate preselection are only considered as engineering heuristics in existing literature. On the other hand, many experimental results have shown that the pathwise coordinate optimization framework is effective at finding local optima with good empirical performance, yet no theoretical guarantee has been established. Thus, a gap exists between theory and practice.

To bridge this gap, we propose a new algorithm, named PICASSO (PathwIse CalibrAted Sparse Shooting algOrithm), which improves the existing pathwise coordinate optimization framework. Particularly, we propose a new greedy selection rule for active set updating and a new convex relaxation based warm start initialization (for sparse learning problems using general loss functions beyond the least square loss). These modifications though simple, have a profound impact: The solution sparsity and restricted strong convexity can be ensured throughout all iterations, which allows us to establish statistical and computational guarantees of PICASSO in high dimensions (Zhang and Huang, 2008; Bickel et al., 2009; Raskutti et al., 2010). Eventually, we prove that PICASSO attains a linear convergence to a unique sparse local optimum with optimal statistical properties in parameter estimation and support recovery (See more details in Section 3). To the best of our knowledge, this is the first result on the computational and statistical guarantees for the pathwise coordinate optimization framework in high dimensions.

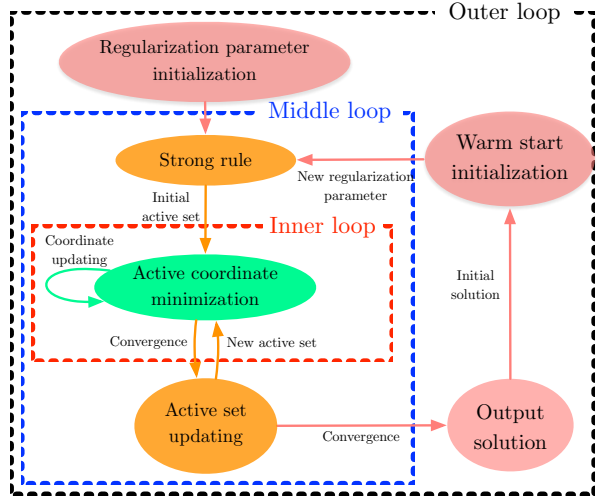

FIG 2. *The pathwise coordinate optimization framework contains 3 nested loops: (I) Warm start initialization; (II) Active set updating and strong rule for coordinate preselection; (III) Active coordinate minimization. Many empirical results have corroborated its outstanding performance. Detailed descriptions of the loops are presented in Section 2.*

Several proximal gradient algorithms are closely related to PICASSO. By exploiting similar sparsity structures of the optimization problem, Wang et al. (2014); Zhao and Liu (2016); Loh and Wainwright (2015) show that these proximal gradient algorithms also attain linear convergence to (approximate) local optima with guaranteed statistical properties. We will compare these algorithms with PICASSO in Section 6.

The rest of this paper is organized as follows: In Section 2, we present the PICASSO algorithm; In Section 3 we present a new theory for analyzing the pathwise coordinate optimization framework, and establish the computational and statistical properties of PICASSO for sparse linear regression;

In Section 4, we extend PICASSO to other sparse learning problems with general loss functions, and provide theoretical guarantees; In Section 5, we present thorough numerical experiments to support our theory; In Section 6, we discuss related work; In Section 7, we present the proofs of the theorems. Due to space limit, the proofs of all lemmas are deferred to the appendix.

**Notations:** Given a vector $v = (v_1, \ldots, v_d)^\top \in \mathbb{R}^d$, we define vector norms: $\|v\|_1 = \sum_j |v_j|$, $\|v\|_2^2 = \sum_j v_j^2$, and $\|v\|_\infty = \max_j |v_j|$. We denote the number of nonzero entries in $v$ as $\|v\|_0 = \sum_j \mathbb{1}_{\{v_j \neq 0\}}$. We define the soft-thresholding function and operator as $\mathcal{S}_\lambda(v_j) = \text{sign}(v_j) \cdot \max\{|v_j| - \lambda, 0\}$ and $\mathcal{S}_\lambda(v) = \big(\mathcal{S}_\lambda(v_1), \ldots, \mathcal{S}_\lambda(v_d)\big)^\top$. We denote $v_{\setminus j} = (v_1, \ldots, v_{j-1}, v_{j+1}, \ldots, v_d)^\top \in \mathbb{R}^{d-1}$ as the subvector of $v$ with the $j$-th entry removed. Let $\mathcal{A} \subseteq \{1, \ldots, d\}$ be an index set. We use $\overline{\mathcal{A}}$ to denote the complementary set to $\mathcal{A}$, i.e. $\overline{\mathcal{A}} = \{j \mid j \in \{1, \ldots, d\}, j \notin \mathcal{A}\}$. We use $v_\mathcal{A}$ to denote a subvector of $v$ by extracting all entries of $v$ with indices in $\mathcal{A}$. Given a matrix $A \in \mathbb{R}^{d \times d}$, we use $A_{*j} = (A_{1j}, \ldots, A_{dj})^\top$ to denote the $j$-th column of $A$, and $A_{k*} = (A_{k1}, \ldots, A_{kd})^\top$ to denote the $k$-th row of $A$. Let $\Lambda_{\max}(A)$ and $\Lambda_{\min}(A)$ be the largest and smallest eigenvalues of $A$. We define the matrix norms $\|A\|_F^2 = \sum_j \|A_{*j}\|_2^2$ and $\|A\|_2$ as the largest singular value of $A$. We denote $A_{\setminus i \setminus j}$ as the submatrix of $A$ with the $i$-th row and the $j$-th column removed. We denote $A_{i \setminus j}$ as the $i$-th row of $A$ with its $j$-th entry removed. Let $\mathcal{A} \subseteq \{1, \ldots, d\}$ be an index set. We use $A_{\mathcal{A}\mathcal{A}}$ to denote a submatrix of $A$ by extracting all entries of $A$ with both row and column indices in $\mathcal{A}$.

**2. Pathwise Calibrated Sparse Shooting Algorithm.** We introduce the PICASSO algorithm for sparse linear regression. PICASSO is a pathwise coordinate optimization algorithm and contains three nested loops (as illustrated in Figure 2). For simplicity, we first introduce its inner loop, then its middle loop, and at last its outer loop.

2.1. *Inner Loop: Iterates over Coordinates within an Active Set.* We start with the inner loop of PICASSO, which is the active coordinate minimization (ActCooMin) algorithm. The iteration index for the inner loop is $(t)$, where $t = 0, 1, 2, \ldots$. As illustrated in Algorithm 1, the ActCooMin algorithm solves (1.1) by iteratively conducting exact coordinate minimization, but it is only allowed to iterate over a subset of all coordinates, which is called "the active set". Accordingly, the complementary set to the active set is called "the inactive set", because the values of these coordinates do not change throughout all iterations of the inner loop. Since the active set usually contains a very small number of coordinates, the active set coordinate minimization algorithm is very scalable and efficient.

For notational simplicity, we denote the active and inactive sets by $\mathcal{A}$ and $\overline{\mathcal{A}}$ respectively. Here we select $\mathcal{A}$ and $\overline{\mathcal{A}}$ based on the sparsity pattern of the initial solution of the inner loop $\theta^{(0)}$,

$$\mathcal{A} = \{j \mid \theta_j^{(0)} \neq 0\} \quad \text{and} \quad \overline{\mathcal{A}} = \{j \mid \theta_j^{(0)} = 0\}.$$

The ActCooMin algorithm then minimizes (1.1) with all coordinates of $\overline{\mathcal{A}}$ staying at zero values,

$$(2.1) \qquad \min_{\theta \in \mathbb{R}^d} \mathcal{F}_\lambda(\theta) \quad \text{subject to} \quad \theta_{\overline{\mathcal{A}}} = 0.$$

The ActCooMin algorithm iterates over all active coordinates in a cyclic order at each iteration. Without loss of generality, we assume

$$|\mathcal{A}| = s \quad \text{and} \quad \mathcal{A} = \{j_1, ..., j_s\} \subseteq \{1, ..., d\},$$

where $j_1 \leq j_2 \leq ... \leq j_s$. Given a solution $\theta^{(t)}$ at the $t$-th iteration, we construct a sequence of auxiliary solutions $\{w^{(t+1,k)}\}_{k=0}^s$ to obtain $\theta^{(t+1)}$. Particularly, for $k = 0$, we take $w^{(t+1,0)} = \theta^{(t)}$; For $k = 1, ..., s$, we take

$$w_{j_k}^{(t+1,k)} = \mathcal{T}_{\lambda, j_k}(w^{(t+1,k-1)}) \quad \text{and} \quad w_{\setminus j_k}^{(t+1,k)} = w_{\setminus j_k}^{(t+1,k-1)},$$

where $\mathcal{T}_{\lambda, j_k}(\cdot)$ is defined in (1.6). We then set $\theta^{(t+1)} = w^{(t+1,s)}$ for the next iteration. Given $\tau$ as a small convergence parameter (e.g., $10^{-5}$), we terminate the ActCooMin algorithm when

$$(2.2) \qquad \|\theta^{(t+1)} - \theta^{(t)}\|_2 \leq \tau\lambda.$$

We then take the output solution as $\widehat{\theta} = \theta^{(t+1)}$.

The ActCooMin algorithm only converges to a local optimum of (2.1), which is not necessarily a local optimum of (1.1). Thus, PICASSO needs to combine this inner loop with some active set updating scheme, which allows the active set to change. This leads to the middle loop of PICASSO.

2.2. *Middle Loop: Iteratively Updates Active Sets.* We then introduce the middle loop of PICASSO, which is the iterative active set updating (IteActUpd) algorithm. The iteration index of the middle loop is $[m]$, where $m = 0, 1, 2, ...$. As illustrated in Algorithm 2, the IteActUpd algorithm simultaneously decreases the objective value and iteratively changes the active set to ensure convergence to a local optimum to (1.1). For notational simplicity, we denote the least square loss function and its gradient as $\mathcal{L}(\theta) = \frac{1}{2n}\|y - X\theta\|_2^2$ and $\nabla \mathcal{L}(\theta) = \frac{1}{n}X^\top(X\theta - y)$.

**Algorithm 1:** *The active coordinate minimization algorithm (Act-CooMin) is the* inner loop *of PICASSO. It iterates over only a small subset of all coordinates in a cyclic order. Thus, its computation is scalable and efficient. Without loss of generality, we assume $|\mathcal{A}| = s$ and $\mathcal{A} = \{j_1, ..., j_s\} \subseteq \{1, ..., d\}$, where $j_1 \leq j_2 \leq ... \leq j_s$.*

**Algorithm:** $\widehat{\theta} \leftarrow \mathsf{ActCooMin}(\lambda, \theta^{(0)}, \mathcal{A}, \tau)$
**Initialize:** $t \leftarrow 0$
**Repeat**
$\quad w^{(t+1,0)} \leftarrow \theta^{(t)}$
$\quad$ **For** $k \leftarrow 1, ..., s$
$\quad\quad w^{(t+1,k)}_{j_k} \leftarrow \mathcal{T}_{\lambda, j_k}(w^{(t+1,k-1)}), \; w^{(t+1,k)}_{\setminus j_k} \leftarrow w^{(t+1,k-1)}_{\setminus j_k}$
$\quad \theta^{(t+1)} \leftarrow w^{(t+1,s)}$
$\quad t \leftarrow t + 1$
**Until** $\|\theta^{(t+1)} - \theta^{(t)}\|_2 \leq \tau \lambda$
**Return:** $\widehat{\theta} \leftarrow \theta^{(t)}$

**(I) Active Set Initialization by Strong Rule:** We first introduce how PICASSO initializes the active set for each middle loop. Suppose an initial solution $\theta^{[0]}$ is supplied to the middle loop of PICASSO. Friedman et al. (2007) suggest a straightforward "simple rule" to initialize the active set based on the sparsity pattern of $\theta^{[0]}$,

$$(2.3) \qquad \mathcal{A}_0 = \{j \mid \theta^{[0]}_j \neq 0\} \quad \text{and} \quad \overline{\mathcal{A}}_0 = \{j \mid \theta^{[0]}_j = 0\}.$$

Tibshirani et al. (2012) further show that (2.3) is sometimes too conservative, and suggest a more aggressive active set initialization procedure using a "strong rule", which often leads to better computational performance in practice. Specifically, given an active set initialization parameter $\varphi \in (0, 1)$, the strong rule[2] for PICASSO initializes $\mathcal{A}_0$ and $\overline{\mathcal{A}}_0$ as

$$(2.4) \qquad \mathcal{A}_0 = \{j \mid \theta^{[0]}_j = 0, \; |\nabla_j \mathcal{L}(\theta^{[0]})| \geq (1-\varphi)\lambda\} \cup \{j \mid \theta^{[0]}_j \neq 0\},$$

$$(2.5) \qquad \overline{\mathcal{A}}_0 = \{j \mid \theta^{[0]}_j = 0, \; |\nabla_j \mathcal{L}(\theta^{[0]})| < (1-\varphi)\lambda\},$$

where $\nabla_j \mathcal{L}(\theta^{[0]})$ denotes the $j$-th entry of $\nabla \mathcal{L}(\theta^{[0]})$. As can be seen from (2.4), the strong rule yields an active set, which is no smaller than the simple rule. Note that we need the initialization parameter $\varphi$ to be a reasonably small value (e.g., 0.1). Otherwise, the strong rule may select too many active coordinates and compromise the solution sparsity.

---

[2] Our proposed strong rule for PICASSO is sightly different from the sequential strong rule proposed in Tibshirani et al. (2012). See more details in Remark 2.2.

**(II) Active Set Updating Strategy:** We then introduce how PICASSO updates the active set at each iteration of the middle loop. Suppose at the $m$-th iteration ($m \geq 1$), we are supplied with a solution $\theta^{[m]}$ with a pair of active and inactive sets defined as

$$\mathcal{A}_m = \{j \mid \theta_j^{[m]} \neq 0\} \quad \text{and} \quad \overline{\mathcal{A}}_m = \{j \mid \theta_j^{[m]} = 0\}.$$

Each iteration of the IteActUpd algorithm contains two stages. The first stage conducts the active coordinate minimization algorithm over the active set $\mathcal{A}_m$ until convergence, and returns a solution $\theta^{[m+0.5]}$. Note that the active coordinate minimization algorithm may yield zero values for some active coordinates. Accordingly, we remove those coordinates from the active set, and obtain a new pair of active and inactive sets as

$$\mathcal{A}_{m+0.5} = \{j \mid \theta_j^{[m+0.5]} \neq 0\} \quad \text{and} \quad \overline{\mathcal{A}}_{m+0.5} = \{j \mid \theta_j^{[m+0.5]} = 0\}.$$

The second stage checks which inactive coordinates of $\overline{\mathcal{A}}_{m+0.5}$ should be added into the active set. Existing pathwise coordinate optimization algorithms usually add inactive coordinates into the active set based on a *cyclic selection rule* (Friedman et al., 2007; Mazumder et al., 2011). Particularly, they conduct the exact coordinate minimization over all coordinates of $\overline{\mathcal{A}}_{m+0.5}$ in a cyclic order. Accordingly, an inactive coordinate is added into the active set if the corresponding exact coordinate minimization step yields a nonzero value. Such a cyclic selection rule, however, has no control over the solution sparsity. It may add too many inactive coordinates into the active set, and compromise the solution sparsity.

To address this issue, we propose a new greedy selection rule for updating the active set. Particularly, let $\nabla_j \mathcal{L}(\theta^{[m+0.5]})$ denote the $j$-th entry of $\nabla \mathcal{L}(\theta^{[m+0.5]})$. We select a coordinate by

$$k_m = \operatorname{argmax}_{k \in \overline{\mathcal{A}}_{m+0.5}} |\nabla_k \mathcal{L}(\theta^{[m+0.5]})|.$$

We then terminate the IteActUpd algorithm if

$$(2.6) \qquad |\nabla_{k_m} \mathcal{L}(\theta^{[m+0.5]})| \leq (1+\delta)\lambda,$$

where $\delta$ is a small convergence parameter (e.g., $10^{-5}$). Otherwise, we take

$$\theta_{k_m}^{[m+1]} = \mathcal{T}_{\lambda, k_m}(\theta^{[m+0.5]}) \quad \text{and} \quad \theta_{\setminus k_m}^{[m+1]} = \theta_{\setminus k_m}^{[m+0.5]},$$

and set the new active and inactive sets as

$$\mathcal{A}_{m+1} = \mathcal{A}_{m+0.5} \cup \{k_m\} \quad \text{and} \quad \overline{\mathcal{A}}_{m+1} = \overline{\mathcal{A}}_{m+0.5} \setminus \{k_m\}.$$

**Algorithm 2:** *The iterative active set updating (IteActUpd) algorithm is the middle loop of PICASSO. It simultaneously decreases the objective value and iteratively changes the active set. To encourage the sparsity of the active set, the greedy selection rule moves only one inactive coordinate to the active set in each iteration.*

**Algorithm:** $\widehat{\theta} \leftarrow \mathsf{IteActUpd}(\lambda, \theta^{[0]}, \delta, \tau, \varphi)$
**Initialize:** $m \leftarrow 0$, $\mathcal{A}_0 \leftarrow \{j \mid \theta_j^{[0]} = 0, \ |\nabla_j \mathcal{L}(\theta^{[0]})| \geq (1-\varphi)\lambda\} \cup \{j \mid \theta_j^{[0]} \neq 0\}$
**Repeat**
$\quad \theta^{[m+0.5]} \leftarrow \mathsf{ActCooMin}(\lambda, \theta^{[m]}, \mathcal{A}_m, \tau)$
$\quad \mathcal{A}_{m+0.5} \leftarrow \{j \mid \theta_j^{[m+0.5]} \neq 0\}, \overline{\mathcal{A}}_{m+0.5} \leftarrow \{j \mid \theta_j^{[m+0.5]} = 0\}$
$\quad k_m \leftarrow \mathrm{argmax}_{k \in \overline{\mathcal{A}}_{m+0.5}} |\nabla_k \mathcal{L}(\theta^{[m+0.5]})|$
$\quad \theta_{k_m}^{[m+1]} \leftarrow \mathcal{T}_{\lambda, k_m}(\theta^{[m+0.5]}), \theta_{\setminus k_m}^{[m+1]} \leftarrow \theta_{\setminus k_m}^{[m+0.5]}$
$\quad \mathcal{A}_{m+1} \leftarrow \mathcal{A}_{m+0.5} \cup \{k_m\}, \overline{\mathcal{A}}_{m+1} \leftarrow \overline{\mathcal{A}}_{m+0.5} \setminus \{k_m\}$
$\quad m \leftarrow m+1$
**Until** $|\nabla_{k_m} \mathcal{L}(\theta^{[m+0.5]})| \leq (1+\delta)\lambda$
**Return:** $\widehat{\theta} \leftarrow \theta^{[m]}$

**Remark 2.1.** Beside the greedy selection rule, we also propose a randomized selection rule and a truncated cyclic selection rule for updating the active set. Due to space limit, we defer the details to Appendix G.

The IteActUpd algorithm, though equipped with the proposed greedy selection rule and strong rule for coordinate preselection, ensures the solution sparsity throughout iterations only for a sufficiently large regularization parameter[3]. Otherwise, given an insufficiently large regularization parameter, the IteActUpd algorithm may still overselect the active coordinates. To address this issue, we combine the IteActUpd algorithm with a sequence of decreasing regularization parameters, which leads to the outer loop of PICASSO.

2.3. *Outer Loop: Iterates over Regularization Parameters.* The outer loop of PICASSO is the warm start initialization (WarmStartInt). The iteration index of the outer loop is $\{K\}$, where $K = 1, ..., N$. As illustrated in Algorithm 3, the warm start initialization solves (1.1) indexed by a geometrically decreasing sequence of regularization parameters $\{\lambda_K = \lambda_0 \eta^K\}_{K=0}^N$ with a common ratio $\eta \in (0, 1)$, and outputs a sequence of $N+1$ solutions $\{\widehat{\theta}^{\{K\}}\}_{K=0}^N$, which is also called the solution path.

For sparse linear regression[4], the warm start initialization chooses the leading regularization parameter $\lambda_0$ as $\lambda_0 = \|\nabla\mathcal{L}(0)\|_\infty = \|\frac{1}{n}X^\top y\|_\infty$. Recall $\mathcal{H}_\lambda(\theta)$ is defined in (1.3). By verifying the KKT condition, we have

$$\min_{\xi\in\partial\|0\|_1} \|\nabla\mathcal{L}(0) + \nabla\mathcal{H}_{\lambda_0}(0) + \lambda_0\xi\|_\infty = \min_{\xi\in\partial\|0\|_1} \|\nabla\mathcal{L}(0) + \lambda_0\xi\|_\infty = 0,$$

where the first equality comes from $\nabla\mathcal{H}_{\lambda_0}(0) = 0$ for the MCP regularizer (See more details in Appendix B). This indicates that 0 is a local optimum of (1.1). Accordingly, we set $\widehat{\theta}^{\{0\}} = 0$. Then for $K = 1, 2, ..., N$, we solve (1.1) for $\lambda_K$ using $\widehat{\theta}^{\{K-1\}}$ as initialization.

---

[3]As will be shown in Section 3, the choice of $\lambda$ is determined by the initial solution of the middle loop.

[4]When dealing with general loss functions, we need a new convex relaxation based

The warm start initialization starts with large regularization parameters to suppress the overselection of the irrelevant coordinates $\{j \mid \theta_j^* = 0\}$ (in conjunction with the IteActUpd algorithm). Thus, the solution sparsity ensures the restricted convexity throughout all iterations, making the algorithm behave as if minimizing a strongly convex function. Though large regularization parameters may also yield zero values for many relevant coordinates $\{j \mid \theta_j^* \neq 0\}$ and result in larger estimation errors, this can be compensated by the decreasing regularization sequence. Eventually, PICASSO gradually recovers the relevant coordinates, reduces the estimation error of each output solution, and attains a sparse output solution with optimal statistical properties in parameter estimation and support recovery.

**Remark 2.2** (Connection to the sequential strong rule). Tibshirani et al. (2012) propose a sequential strong rule for coordinate preselection, which initializes the active set for $\lambda_K$ as

$$(2.7) \qquad \mathcal{A}_0 = \{j \mid \theta_j^{[0]} = 0, \ |\nabla_j\mathcal{L}(\theta^{[0]})| \geq 2\lambda_K - \lambda_{K-1}\} \cup \{j \mid \theta_j^{[0]} \neq 0\},$$

$$(2.8) \qquad \overline{\mathcal{A}}_0 = \{j \mid \theta_j^{[0]} = 0, \ |\nabla_j\mathcal{L}(\theta^{[0]})| < 2\lambda_K - \lambda_{K-1}\}.$$

Recall $\lambda_K = \eta\lambda_{K-1}$. Then we have $2\lambda_K - \lambda_{K-1} = (1 - (1-\eta)/\eta)\,\lambda_K$. This indicates that the sequential strong rule is a special case of our strong rule for PICASSO with $\varphi = (1-\eta)/\eta$.

**3. Computational and Statistical Theory.** We develop a new theory to analyze the pathwise coordinate optimization framework, and establish the computational and statistical properties of PICASSO for sparse linear regression. Recall our linear model assumption is $y = X\theta^* + \varepsilon$, where $\varepsilon \sim N(0, \sigma^2 I_n)$[5]. Moreover, in (1.3), we rewrite the nonconvex regularizer as $\mathcal{R}_\lambda(\theta) = \lambda\|\theta\|_1 + \mathcal{H}_\lambda(\theta)$, where $\mathcal{H}_\lambda(\theta) = \sum_{j=1}^{d} h_\lambda(|\theta_j|)$ is a smooth, concave,

---

warm start initialization approach, which will be introduced in Section 4.2.

[5] For simplicity, we only consider the Gaussian noise setting, but it is straight forward to extend our analysis to the sub-Gaussian noise setting.

PATHWISE COORDINATE OPTIMIZATION: ALGORITHM AND THEORY 13...

**Algorithm 3:** *The warm start initialization is the **outer loop** of PI-CASSO. It solves* (1.1) *with respect to a decreasing sequence of regularization parameters* $\{\lambda_K\}_{K=0}^N$. *The leading regularization parameter* $\lambda_0$ *is chosen as* $\lambda_0 = \|\nabla\mathcal{L}(0)\|_\infty$, *which yields an all zero output solution* $\widehat{\theta}^{\{0\}} = 0$. *For* $K = 1, ..., N$, *we solve* (1.1) *for* $\lambda_K$ *using* $\widehat{\theta}^{\{K-1\}}$ *as an initial solution.* $\{\tau_K\}_{K=1}^N$ *and* $\{\delta_K\}_{K=1}^N$ *are two sequences of small convergence parameters, where* $\tau_K$ *and* $\delta_K$ *correspond to the* $K$-*th outer loop iteration with the regularization parameter* $\lambda_K$.

**Algorithm:** $\{\widehat{\theta}^{\{K\}}\}_{K=0}^N \leftarrow \mathsf{WarmStartInt}(\{\lambda_K\}_{K=0}^N)$
**Parameter:** $\eta$, $\varphi$, $\{\tau_K\}_{K=1}^N$, $\{\delta_K\}_{K=1}^N$
**Initialize:** $\lambda_0 \leftarrow \|\nabla\mathcal{L}(0)\|_\infty$, $\widehat{\theta}^{\{0\}} \leftarrow 0$
**For** $K \leftarrow 1, 2, ..., N$
 $\quad \lambda_K \leftarrow \eta\lambda_{K-1}$
 $\quad \widehat{\theta}^{\{K\}} \leftarrow \mathsf{IteActUpd}(\lambda_K, \widehat{\theta}^{\{K-1\}}, \delta_K, \tau_K, \varphi)$
**Return:** $\{\widehat{\theta}^{\{K\}}\}_{K=0}^N$

and coordinate decomposable function. For notational simplicity, we define $\widetilde{\mathcal{L}}_\lambda(\theta) = \mathcal{L}(\theta) + \mathcal{H}_\lambda(\theta)$. Accordingly, we rewrite $\mathcal{F}_\lambda(\theta)$ as

$$\mathcal{F}_\lambda(\theta) = \mathcal{L}(\theta) + \mathcal{R}_\lambda(\theta) = \widetilde{\mathcal{L}}_\lambda(\theta) + \lambda\|\theta\|_1.$$

3.1. *Computational Theory.* We first introduce three assumptions. The first assumption requires $\lambda_N$ to be sufficiently large.

**Assumption 3.1.** We require that the regularization sequence satisfies

(3.1) $$\lambda_N = 8\sigma\sqrt{\frac{\log d}{n}} \geq 4\|\nabla\mathcal{L}(\theta^*)\|_\infty = \frac{4}{n}\|X^\top\varepsilon\|_\infty.$$

Moreover, we require $\eta \in [0.96, 1)$.

Assumption 3.1 ensures that all regularization parameters are sufficiently large to eliminate the irrelevant coordinates for PICASSO.

**Remark 3.2.** Note that Assumption 3.1 is a deterministic bound for our chosen $\lambda_N$. As will be shown in Lemma 3.13, since $\|X^\top\varepsilon\|_\infty$ is random, we need to verify that (3.1) holds with high probability when applying PICASSO to sparse linear regression.

Before we present the second assumption, we define the largest and smallest $s$ sparse eigenvalues of the Hessian matrix $\nabla^2\mathcal{L}(\theta) = \frac{1}{n}X^\top X$ as follows.

**Definition 3.3.** Given an integer $s \geq 1$, we define the largest and smallest $s$ sparse eigenvalues as

$$\rho_+(s) = \sup_{\|v\|_0 \leq s} \frac{v^\top \nabla^2 \mathcal{L}(\theta) v}{\|v\|_2^2} \quad \text{and} \quad \rho_-(s) = \inf_{\|v\|_0 \leq s} \frac{v^\top \nabla^2 \mathcal{L}(\theta) v}{\|v\|_2^2}.$$

The next lemma connects the largest and smallest $s$ sparse eigenvalues to the restricted strong convexity and smoothness.

**Lemma 3.4.** Suppose there exists an integer $s$ such that $0 < \rho_-(s) \leq \rho_+(s) < \infty$. For any $\theta, \theta' \in \mathbb{R}^d$ satisfying $\|\theta - \theta'\|_0 \leq s$, $\mathcal{L}(\theta)$ is restricted strongly convex and smooth,

$$(3.2) \quad \frac{\rho_-(s)}{2}\|\theta' - \theta\|_2^2 \leq \mathcal{L}(\theta') - \mathcal{L}(\theta) - (\theta' - \theta)^\top \nabla \mathcal{L}(\theta) \leq \frac{\rho_+(s)}{2}\|\theta' - \theta\|_2^2.$$

Moreover, given $\alpha = 1/\gamma \leq \rho_-(s)$ and $\widetilde{\rho}_-(s) = \rho_-(s) - \alpha > 0$, where $\gamma$ is the concavity parameter of MCP defined in (1.2), for any $\theta, \theta' \in \mathbb{R}^d$ satisfying $\|\theta - \theta'\|_0 \leq s$, $\widetilde{\mathcal{L}}_\lambda(\theta)$ is restricted strongly convex and smooth,

$$\frac{\widetilde{\rho}_-(s)}{2}\|\theta' - \theta\|_2^2 \leq \widetilde{\mathcal{L}}_\lambda(\theta') - \widetilde{\mathcal{L}}_\lambda(\theta) - (\theta' - \theta)^\top \nabla \widetilde{\mathcal{L}}_\lambda(\theta) \leq \frac{\rho_+(s)}{2}\|\theta' - \theta\|_2^2.$$

Meanwhile, for any $\xi \in \partial \|\theta\|_1$, $\mathcal{F}_\lambda(\theta)$ is restricted strongly convex,

$$\frac{\widetilde{\rho}_-(s)}{2}\|\theta' - \theta\|_2^2 \leq \mathcal{F}_\lambda(\theta') - \mathcal{F}_\lambda(\theta) - (\theta' - \theta)^\top (\nabla \widetilde{\mathcal{L}}_\lambda(\theta) + \lambda \xi).$$

Lemma 3.4 indicates the importance of the solution sparsity: When $\theta$ is sufficiently sparse, the restricted strong convexity of $\mathcal{L}(\theta)$ dominates the concavity of $\mathcal{H}_\lambda(\theta)$. Thus, if an algorithm ensures the solution sparsity throughout all iterations, it will behave like minimizing a strongly convex optimization problem. Accordingly, a linear convergence can be established. Note that Lemma 3.4 is also applicable to Lasso, since Lasso satisfies $\alpha = 1/\gamma = 1/\infty = 0$. Now we introduce the second assumption.

**Assumption 3.5.** Given $\|\theta^*\|_0 \leq s^*$, there exists an integer $\widetilde{s}$ such that

$$\widetilde{s} \geq (484\kappa^2 + 100\kappa)s^*, \ \rho_+(s^* + 2\widetilde{s}) < \infty, \ \text{and} \ \widetilde{\rho}_-(s^* + 2\widetilde{s}) > 0,$$

where $\kappa$ is defined as $\kappa = \rho_+(s^* + 2\widetilde{s})/\widetilde{\rho}_-(s^* + 2\widetilde{s})$.

Assumption 3.5 guarantees that the optimization problem satisfies the restricted strong convexity as long as the number of active irrelevant coordinates never exceeds $\widetilde{s}$ throughout all iterations.

**Remark 3.6.** Assumptions 3.1 and 3.5 are closely related to high dimensional statistical theories for sparse linear regression in existing literature. See more details in Zhang and Huang (2008); Bickel et al. (2009); Zhang (2010); Negahban et al. (2012).

Now we introduce the last assumption on the computational parameters.

**Assumption 3.7.** Recall the convergence parameters $\delta_K$'s and $\tau_K$'s are defined in Algorithm 3, and the active set initialization parameter $\varphi$ is defined in (2.4). We require for all $K = 1, ..., N$,

$$\delta_K \leq \frac{1}{8}, \quad \tau_K \leq \frac{\delta_K}{\rho_+(s^* + 2\widetilde{s})}\sqrt{\frac{\widetilde{\rho}_-(1)}{\rho_+(1)(s^* + 2\widetilde{s})}}, \quad \text{and} \quad \varphi \leq \frac{1}{8}.$$

Assumption 3.7 guarantees that all middle and inner loops of PICASSO attain adequate precisions such that their output solutions satisfy the desired computational and statistical properties.

**Remark 3.8.** All constants in our technical assumptions and proof are for providing insights of PICASSO. We do not make efforts on optimizing any of these constants. Taking Assumption 3.1 as an example, we choose $\eta \in [0.96, 1)$ just for easing our analysis. We can also choose $\eta$ to be any other constant, e.g., 0.95, as long as it is sufficiently close to 1. Such a change in $\eta$ only affects the required sample complexity, iteration complexity, and statistical rates of convergence up to small constant factors.

Now, we start with the convergence analysis for the inner loop of PICASSO. The following theorem presents the convergence rate in terms of the objective value. For notational simplicity, we omit the outer loop index $K$, and denote $\lambda_K$ and $\tau_K$ by $\lambda$ and $\tau$ respectively.

**Theorem 3.9.** [Inner Loop] Suppose Assumption 3.5 holds. If the initial active set satisfies $|\mathcal{A}| = s \leq s^* + 2\widetilde{s}$, then (2.1) is essentially strongly convex. For $t = 1, 2...$, we have

$$\mathcal{F}_\lambda(\theta^{(t)}) - \mathcal{F}_\lambda(\overline{\theta}) \leq \left(\frac{s\rho_+^2(s)}{s\rho_+^2(s) + \widetilde{\rho}_-(s)\widetilde{\rho}_-(1)}\right)^t [\mathcal{F}_\lambda(\theta^{(0)}) - \mathcal{F}_\lambda(\overline{\theta})],$$

where $\overline{\theta}$ is a unique global optimum to (2.1). Moreover, we need at most

$$\left(1 + \frac{s\rho_+^2(s)}{\widetilde{\rho}_-(s)\widetilde{\rho}_-(1)}\right) \cdot \log\left(\frac{2[\mathcal{F}_\lambda(\theta^{(0)}) - \mathcal{F}_\lambda(\overline{\theta})]}{\widetilde{\rho}_-(1)\tau^2\lambda^2}\right)$$

iterations to terminate the ActCooMin algorithm, where $\tau$ is defined in (2.2).

Theorem 3.9 guarantees that given a sufficiently sparse active set, Algorithm 1 essentially minimizes a strongly convex optimization problem, though (1.1) is globally nonconvex. Thus, it attains a linear convergence to a unique global optimum.

Then, we proceed with the convergence analysis for the middle loop of PICASSO. The following theorem presents the convergence rate in terms of the objective value. For notational simplicity, we omit the outer loop index $K$, and denote $\lambda_K$ and $\delta_K$ by $\lambda$ and $\delta$ respectively. Moreover, we define

$$(3.3) \quad \triangle_\lambda = \frac{4\lambda^2 s^*}{\widetilde{\rho}_-(s^* + \widetilde{s})}, \quad \mathcal{S} = \{j \mid \theta_j^* \neq 0\}, \quad \text{and} \quad \overline{\mathcal{S}} = \{j \mid \theta_j^* = 0\}.$$

**Theorem 3.10.** [Middle Loop] Suppose Assumptions 3.1, 3.5, and 3.7 hold. For any $\lambda \geq \lambda_N$, if the initial solution $\theta^{[0]}$ satisfies $\|\theta^{[0]}_{\overline{\mathcal{S}}}\|_0 \leq \widetilde{s}$ and $\mathcal{F}_\lambda(\theta^{[0]}) \leq \mathcal{F}_\lambda(\theta^*) + \triangle_\lambda$, then regardless the active set initialized by either the strong rule or simple rule, we have $|\mathcal{A}_0 \cap \overline{\mathcal{S}}| \leq \widetilde{s}$. Meanwhile, for $m = 0, 1, 2, ...$, we also have $\|\theta^{[m]}_{\overline{\mathcal{S}}}\|_0 \leq \widetilde{s} + 1$, $\|\theta^{[m+0.5]}_{\overline{\mathcal{S}}}\|_0 \leq \widetilde{s}$, and

$$\mathcal{F}_\lambda(\theta^{[m]}) - \mathcal{F}_\lambda(\overline{\theta}^\lambda) \leq \left(1 - \frac{\widetilde{\rho}_-(s^* + 2\widetilde{s})}{(s^* + 2\widetilde{s})\rho_+(1)}\right)^m [\mathcal{F}_\lambda(\theta^{(0)}) - \mathcal{F}_\lambda(\overline{\theta}^\lambda)],$$

where $\overline{\theta}^\lambda$ is a unique sparse local optimum of (1.1) satisfying

$$(3.4) \qquad \mathcal{K}_\lambda(\overline{\theta}^\lambda) = \min_{\xi \in \partial \|\overline{\theta}^\lambda\|_1} \|\nabla \widetilde{\mathcal{L}}_\lambda(\overline{\theta}^\lambda) + \lambda \xi\|_\infty = 0 \quad \text{and} \quad \|\overline{\theta}^\lambda_{\overline{\mathcal{S}}}\|_0 \leq \widetilde{s}.$$

Moreover, recall $\delta$ is defined in (2.6), we need at most

$$\frac{(s^* + 2\widetilde{s})\rho_+(1)}{\widetilde{\rho}_-(s^* + 2\widetilde{s})} \cdot \log\left(\frac{3\rho_+(1)[\mathcal{F}_\lambda(\theta^{[0]}) - \mathcal{F}_\lambda(\overline{\theta}^\lambda)]}{\delta^2 \lambda^2}\right)$$

active set updating iterations to terminate the IteActUpd algorithm. Meanwhile, we have the output solution $\widehat{\theta}^\lambda$ satisfying $\mathcal{K}_\lambda(\widehat{\theta}^\lambda) \leq \delta\lambda$.

Theorem 3.10 guarantees that when supplied a proper initial solution, the middle loop of PICASSO attains a linear convergence to a unique sparse local optimum. Moreover, Theorem 3.10 has three important implications:

(I) The greedy rule is conservative and only select one coordinate each time. This mechanism prevents the overselection of the irrelevant coordinates and encourages the solution sparsity. In contrast, the cyclic selection rule in Mazumder et al. (2011) may overselect the irrelevant coordinates and compromise the restricted convexity. An illustration is provided in Figure 3.

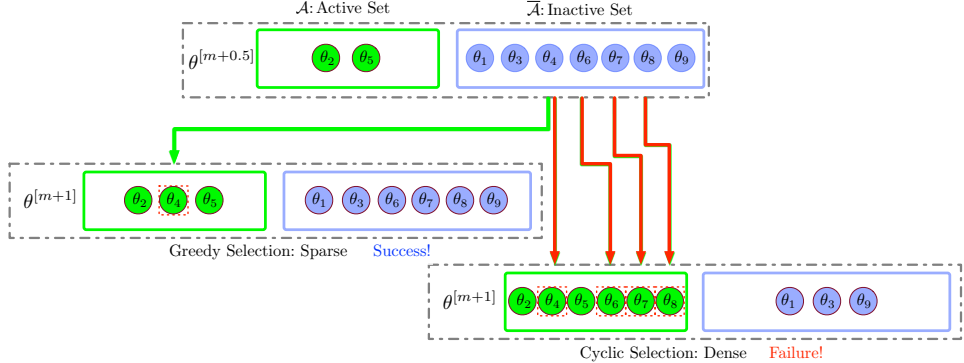

FIG 3. *An illustration of the failure of the cyclic selection rule. The green and blue circles denote the active and inactive coordinates respectively. Suppose we have 9 coordinates and the maximum number of active coordinates we can tolerate is* 4. *The greedy selection rule is conservative, and only add one coordinate to the active set each time. Thus, it eventually increases the number of active coordinates from* 2 *to* 3, *and prevents from overselecting coordinates. In contrast, the cyclic selection rule used in* Friedman et al. (2007); Mazumder et al. (2011) *leads to overselecting coordinates, which eventually increases the number of active coordinates to* 6. *Thus, it fails to preserve the restricted strong convexity.*

(II) Besides decreasing the objective value, the active coordinate minimization algorithm can remove some irrelevant coordinates from the active set. Thus, in conjunction with the greedy selection rule, the solution sparsity is ensured throughout all iterations. An illustration is provided in Figure 4. To the best of our knowledge, such a "forward-backward" phenomenon has not been discovered and rigorously characterized in existing literature.

(III) The strong rule for coordinate preselection in PICASSO put some coordinates with zero values to the active set, only when their corresponding coordinate gradients have sufficiently large magnitudes. Thus, it can also prevent the overselection of the irrelevant coordinates and ensure the solution sparsity.

Next, we proceed with the convergence analysis for the outer loop of PICASSO. As has been shown in Theorem 3.10, each middle loop of PICASSO requires a proper initialization. Since $\theta^*$ and $\mathcal{S}$ are unknown in practice, it is difficult to manually pick such an initial solution. The next theorem shows that the warm start initialization guides PICASSO to attain such a proper initialization for every middle loop without any prior knowledge.

**Lemma 3.11.** [Outer Loop] Recall $\triangle_{\lambda_K}$ and $\mathcal{K}_{\lambda_K}(\theta)$ are defined in (3.3) and (3.4) respectively. Suppose Assumptions 3.1, 3.5, and 3.7 hold. If $\theta$ satisfies $\|\theta_{\overline{S}}\|_0 \leq \widetilde{s}$ and $\mathcal{K}_{\lambda_{K-1}}(\theta) \leq \delta_{K-1}\lambda_{K-1}$, then we have

$$\|\widehat{\Delta}\|_1 \leq 11\|\widehat{\Delta}_{\mathcal{S}}\|_1 \leq 11\sqrt{s^*}\|\widehat{\Delta}\|_2,\ \mathcal{K}_{\lambda_K}(\theta) \leq \frac{\lambda_K}{4},\ \mathcal{F}_{\lambda_K}(\theta) \leq \mathcal{F}_{\lambda_K}(\theta^*) + \triangle_{\lambda_K}.$$

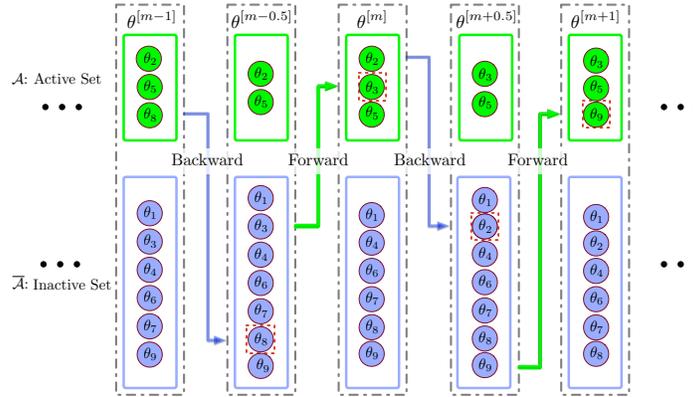

FIG 4. *An illustration of the active set updating algorithm. The green and blue circles denote the active and inactive coordinates respectively. Suppose we have 9 coordinates, and the maximum number of active coordinates we can tolerate is* 4. *The active set updating iteration first removes some active coordinates from the active set, then add some inactive coordinates into the active set. Thus, the number of active coordinates is ensured to never exceed* 4 *throughout all iterations. To the best of our knowledge, such a "forward-backward" phenomenon has not been discovered and rigorously characterized in existing literature.*

The warm start initialization starts with an all zero local optimum and a sufficiently large $\lambda_0$, which naturally satisfy all requirements

$$\|0_{\overline{S}}\|_0 \leq \widetilde{s} \quad \text{and} \quad \mathcal{K}_{\lambda_0}(0) = 0.$$

Thus, $\theta^{[0]} = 0$ is a proper initial solution for $\lambda_1$. Then combining Theorems 3.10 and 3.11, we show by induction that the output solution of each middle loop is always a proper initial solution for the next middle loop. The warm start initialization is illustrated in Figure 5.

Combining Theorems 3.9 and 3.10 with Lemma 3.11, we establish the global convergence in terms of the objective value for PICASSO.

**Theorem 3.12.** [Main Theorem] Suppose Assumptions 3.1, 3.5, and 3.7 hold. Recall $\alpha = 1/\gamma$ and $\gamma$ is the concavity parameter defined in (1.2), $\delta_K$'s and $\tau_K$'s are the convergence parameters for the middle and inner loops within the $K$-th iteration of the outer loop, and $\kappa$ and $\widetilde{s}$ are defined in Assumption 3.5. For $K = 1, \cdots, N$, we have:

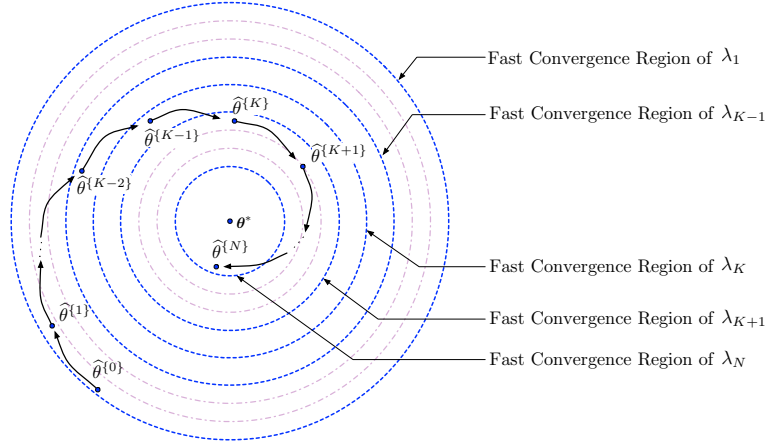

FIG 5. *An illustration of the warm start initialization (the outer loop). From an intuitive geometric perspective, the warm start initialization yields a sequence of nested fast convergence regions. We start with large regularization parameters. This suppresses the overselection of the irrelevant coordinates $\{j \mid \theta_j^* = 0\}$ and yields highly sparse solutions. With the decrease of the regularization parameter, PICASSO gradually recovers the relevant coordinates, and eventually obtains a sparse estimator $\widehat{\theta}^{\{N\}}$ with optimal statistical properties in both parameter estimation and support recovery.*

(I) At the $K$-th iteration of the outer loop, the number of exact coordinate minimization iterations within each inner loop is at most

$$\left(s^* + 2\widetilde{s} + \frac{(s^* + 2\widetilde{s})^2 \rho_+^2(s^* + 2\widetilde{s})}{\widetilde{\rho}_-(s^* + 2\widetilde{s})\widetilde{\rho}_-(1)}\right) \cdot \log\left(\frac{50 s^*}{\widetilde{\rho}_-(1)\tau_K^2 \widetilde{\rho}_-(s^* + \widetilde{s})}\right);$$

(II) At the $K$-th iteration of the outer loop, the number of active set updating iterations is at most

$$\frac{(s^* + 2\widetilde{s})\rho_+(1)}{\widetilde{\rho}_-(s^* + 2\widetilde{s})} \cdot \log\left(\frac{75 s^* \rho_+(1)}{\delta_K^2 \widetilde{\rho}_-(s^* + \widetilde{s})}\right);$$

(III) At the $K$-th iteration of the outer loop, we have

$$\mathcal{F}_{\lambda_N}(\widehat{\theta}^{\{K\}}) - \mathcal{F}_{\lambda_N}(\overline{\theta}^{\lambda_N}) \leq \left[\mathbb{1}_{\{K<N\}} + \mathbb{1}_{\{K=N\}} \cdot \delta_N\right] \frac{50 \lambda_K^2 s^*}{\widetilde{\rho}_-(s^* + \widetilde{s})}.$$

Theorem 3.12 guarantees that PICASSO attains a linear convergence to a unique sparse local optimum, which is a significant improvement over sublinear convergence of the randomized coordinate minimization algorithms established in existing literature. To the best of our knowledge, this is the first result establishing the convergence properties of the pathwise coordinate optimization framework in high dimensions.

3.2. *Statistical Theory.* Finally, we analyze the statistical properties of the estimator obtained by PICASSO for sparse linear regression. We assume $\|\theta^*\|_0 \leq s^*$, and for any $v \neq 0$, the design matrix $X$ satisfies

$$(3.5) \qquad \psi_\ell \|v\|_2^2 - \gamma_\ell \cdot \frac{\log d}{n} \|v\|_1^2 \leq \frac{\|Xv\|_2^2}{n} \leq \psi_u \|v\|_2^2 + \gamma_u \cdot \frac{\log d}{n} \|v\|_1^2,$$

where $\gamma_\ell$, $\gamma_u$, $\psi_\ell$, and $\psi_u$ are positive constants, and do not scale with $(s^*, n, d)$. Existing literature has shown that (3.5) is satisfied by many common examples of sub-Gaussian random design with high probability (Raskutti et al., 2010; Negahban et al., 2012).

We then verify Assumptions 3.1 and 3.5 by the following lemma.

**Lemma 3.13.** Suppose $\varepsilon \sim N(0, \sigma^2 I_n)$ and (3.5) holds. Given $\lambda_N = 8\sigma \sqrt{\log d/n}$, we have

$$\mathbb{P}\left(\lambda_N \geq 4\|\nabla \mathcal{L}(\theta^*)\|_\infty = \frac{4}{n}\|X^\top \varepsilon\|_\infty\right) \geq 1 - 2d^{-2}.$$

Moreover, given $\|\frac{1}{n}X^\top X\|_1 = \mathcal{O}(d)$, $\|\theta^*\|_\infty = \mathcal{O}(d)$, $\gamma \geq 4/\psi_\ell$, and large enough $n$, there exists a generic constant $C_1$ such that we have $N = \mathcal{O}_P(\log d)$,

$$\widetilde{s} = C_1 s^* \geq [484\kappa^2 + 100\kappa] \cdot s^*, \ \widetilde{\rho}_-(s^* + 2\widetilde{s}) \geq \frac{\psi_\ell}{4}, \ \text{and} \ \rho_+(s^* + 2\widetilde{s}) \leq \frac{5\psi_u}{4}.$$

Lemma 3.13 guarantees that the regularization sequence satisfies Assumption 3.1 with high probability, and Assumption 3.5 holds when the design matrix satisfies (3.5). Thus, by Theorem 3.12, we know that with high probability, PICASSO attains a linear convergence to a unique sparse local optimum for sparse linear regression. Moreover, Lemma 3.13 also implies that the number of regularization parameters only needs to be the order of $\log d$. Thus, solving the optimization problem with a sequence of regularization parameters does not require much additional efforts.

We then characterize the statistical rate of convergence in parameter estimation for the estimator obtained by PICASSO.

**Theorem 3.14** (Parameter Estimation). Suppose $\varepsilon \sim N(0, \sigma^2 I_n)$ and (3.5) holds. Given $\gamma \geq 4/\psi_\ell$ and $\lambda_N = 8\sigma \sqrt{\log d/n}$, for small enough $\delta_N$ and large enough $n$ such that $n \geq C_2 s^* \log d$ for a generic constant $C_2$, we have

$$\|\widehat{\theta}^{\{N\}} - \theta^*\|_2 = \mathcal{O}_P\left(\sigma \sqrt{\frac{s_1^*}{n}} + \sigma \sqrt{\frac{s_2^* \log d}{n}}\right),$$

where $s_1^* = |\{j \mid |\theta_j^*| \geq \gamma \lambda_N\}|$ and $s_2^* = |\{j \mid 0 < |\theta_j^*| < \gamma \lambda_N\}|$.

By dividing all nonzero $\theta_j^*$'s into strong signals and weak signals by their magnitudes, Theorem 3.14 shows that the MCP regularizer reduces the estimation error for strong signal with magnitudes larger than $\gamma \lambda_N$, and therefore attains a faster statistical rate of convergence than Lasso.

**Remark 3.15** (Parameter Estimation for Lasso). Theorem 3.14 is also applicable to Lasso with $\gamma = \infty$. As a result, all nonzero $\theta_j^*$'s are considered as weak signals $|\theta_j^*| < \infty$ for all $j = 1, .., d$, i.e., $s_1^* = 0$ and $s_2^* = s^*$. Theorem 3.14 only guarantees a slower statistical rate of convergence for Lasso,

$$\|\widehat{\theta}^{\{N\}} - \theta^*\|_2 = \mathcal{O}_P\left(\sigma\sqrt{\frac{s_2^* \log d}{n}}\right) = \mathcal{O}_P\left(\sigma\sqrt{\frac{s^* \log d}{n}}\right) \quad \text{for} \quad \gamma = \infty.$$

We then proceed to show that the statistical rate of convergence in Theorem 3.14 is minimax optimal in parameter estimation for a suitably chosen $\gamma < \infty$. Particularly, we consider a class of sparse vectors:

(3.6) $\Theta(s_1^*, s_2^*, d) = \{\theta^* \mid \theta^* \in \mathbb{R}^d, \ \sum_{j=1}^d \mathbb{1}_{\{|\theta_j^*| \geq \theta_{\min}\}} \leq s_1^*,$
$$\sum_{j=1}^d \mathbb{1}_{\{0 < |\theta_j^*| < \theta_{\min}\}} \leq s_2^*\},$$

where $\theta_{\min} = \frac{8\gamma\sigma}{\sqrt{C_2(s_1^* + s_2^*)}}$ is the threshold between strong and week signals for some generic constant $C_2$ and $\gamma < \infty$. Given $s^* = s_1^* + s_2^*$ and $n \geq C_2 s^* \log d$, we have

$$\theta_{\min} = \frac{8\gamma\sigma}{\sqrt{C_2(s_1^* + s_2^*)}} \geq 8\gamma\sigma\sqrt{\frac{\log d}{n}} = \gamma\lambda_N,$$

which matches the threshold for dividing signals in Theorem 3.14. The next theorem establishes a lower bound for parameter estimation.

**Theorem 3.16** (Lower Bound). Let $\widehat{\theta}$ denote any estimator of $\theta^*$ based on $y \sim N(X\theta^*, \sigma^2 I_n)$, where $\theta^* \in \Theta(s_1^*, s_2^*, d)$. Then there exists a generic constant $C_4$ such that

$$\inf_{\widehat{\theta}} \sup_{\theta \in \Theta(s_1^*, s_2^*, d)} \mathbb{E}\|\widehat{\theta} - \theta^*\|_2 \geq C_4\left(\sigma\sqrt{\frac{s_1^*}{n}} + \sigma\sqrt{\frac{s_2^* \log d}{n}}\right).$$

Theorem 3.16 guarantees that the estimator obtained by PICASSO attains the minimax optimal rates of convergence over $\Theta(s_1^*, s_2^*, d)$. The convex $\ell_1$ regularizer, however, only attains a suboptimal statistical rate of convergence due to the universal estimation bias regardless the signal strength. See more details in Zhang and Huang (2008); Bickel et al. (2009).

To analyze the support recovery performance for the estimator obtained by PICASSO, we define the oracle least square estimator $\widehat{\theta}^{\mathrm{o}}$ as

$$(3.7) \qquad \widehat{\theta}^{\mathrm{o}}_{\mathcal{S}} = \operatorname*{argmin}_{\theta_{\mathcal{S}}} \frac{1}{2n}\|y - X_{*\mathcal{S}}\theta_{\mathcal{S}}\|_2^2 \quad \text{and} \quad \widehat{\theta}^{\mathrm{o}}_{\overline{\mathcal{S}}} = 0,$$

where $\mathcal{S}$ and $\overline{\mathcal{S}}$ are defined in (3.3). Recall $\overline{\theta}^{\lambda_N}$ is the unique sparse local minimizer to (1.1) with $\lambda_N$. The following theorem shows that $\overline{\theta}^{\lambda_N}$ is identical to the oracle least square estimator $\widehat{\theta}^{\mathrm{o}}$ with high probability.

**Theorem 3.17** (Support Recovery). Suppose (3.5) holds,

$$(3.8) \qquad \varepsilon \sim N(0, \sigma^2 I_n), \quad \text{and} \quad \min_{j \in \mathcal{S}} |\theta_j^*| \geq C_5 \gamma \sigma \sqrt{\frac{\log d}{n}}$$

for a generic constant $C_5$. Given $4/\psi_\ell \leq \gamma < \infty$ and $\lambda_N = 8\sigma\sqrt{\log d/n}$, for large enough $n$, there exits a generic constant $C_3$ such that $\mathbb{P}(\overline{\theta}^{\lambda_N} = \widehat{\theta}^{\mathrm{o}}) \geq 1 - 4d^{-2}$. Meanwhile, with probability at least $1 - 4d^{-2}$, we also have

$$\|\widehat{\theta}^{\{N\}} - \theta^*\|_2 \leq C_3 \sigma \sqrt{\frac{s^*}{n}} \quad \text{and} \quad \operatorname{supp}(\widehat{\theta}^{\{N\}}) = \operatorname{supp}(\theta^*).$$

Theorem 3.17 guarantees that PICASSO converges to $\widehat{\theta}^{\mathrm{o}}$ with high probability, which is often referred to the oracle property in existing literature (Fan and Li, 2001). Besides, we also guarantee that the estimator $\widehat{\theta}^{\{N\}}$ obtained by PICASSO is nearly unbiased and correctly identifies the true support with high probability. Although the $\ell_1$ regularizer can be viewed as a special case of MCP, such an oracle property does not hold Lasso. This is because we require $\gamma < \infty$ such that the estimation bias can be eliminated for strong signals. Thus Lasso cannot guarantee the correct support recovery (unless the design matrix satisfies a restrictive irrepresentable condition–see more details in Zhao and Yu (2006); Meinshausen and Bühlmann (2006); Zou (2006)). We present an illustration of Theorems 3.14 and 3.17 in Figure 6.

**4. Extension to General Loss Functions.** PICASSO can be further extended to other regularized M-estimation problems. Taking sparse logistic regression as an example[6], we denote the binary response vector by $y = (y_1, ..., y_n)^\top \in \mathbb{R}^n$, and the design matrix by $X \in \mathbb{R}^{n \times d}$. We consider a logistic model with $\mathbb{P}(y_i = 1) = \pi_i(\theta^*)$ and $\mathbb{P}(y_i = -1) = 1 - \pi_i(\theta^*)$, where

$$(4.1) \qquad \pi_i(\theta) = \frac{1}{1 + \exp(-X_{i*}^\top \theta)} \text{ for } i = 1, ..., n.$$

---

[6]Due to space limit, we only present sparse logistic regression as an example. Please see more details on sparse robust regression using the huber loss function in Appendix F.

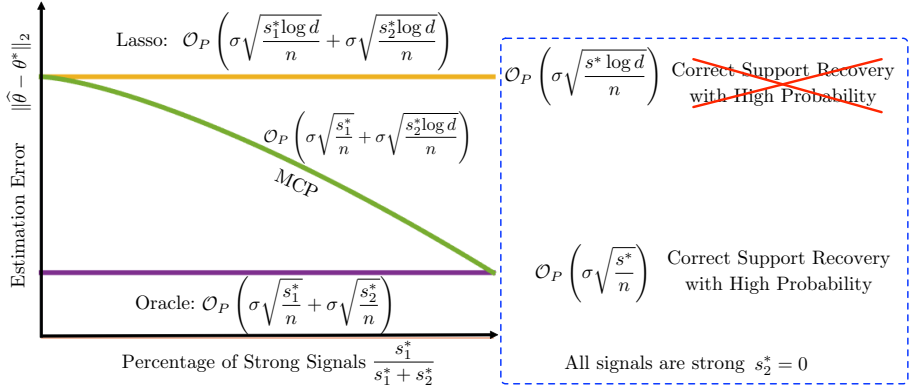

FIG 6. *An illustration of the statistical rates of convergence in parameter estimation and support recovery for the Lasso, MCP, and oracle estimators. Recall $s_1^*$ and $s_2^*$ are defined in (3.6), and $s^* = s_1^* + s_2^*$. When all the signals are weak ($s_1^* = 0, s^* = s_2^*$), both the Lasso and MCP estimators attain the same estimation error bound $\mathcal{O}_P(\sigma\sqrt{s^* \log d/n})$. When some signals are strong, the MCP-regularized estimator attains a better estimation error bound $\mathcal{O}_P(\sigma\sqrt{s_1^*/n} + \sigma\sqrt{s_2^* \log d/n})$ than Lasso, because it reduces the estimation bias for the strong signals. Eventually, when all the signals are strong ($s_2^* = 0, s^* = s_1^*$), the MCP estimator attains the same estimation error bound as the oracle estimator $\mathcal{O}_P(\sigma\sqrt{s^*/n})$, and also correctly identify the support true support with high probability.*

When $\theta^*$ is sparse, we consider the optimization problem

$$(4.2) \quad \min_{\theta \in \mathbb{R}^d} \mathcal{L}(\theta) + \mathcal{R}_\lambda(\theta), \quad \text{where } \mathcal{L}(\theta) = \frac{1}{n} \sum_{i=1}^n \left[ \log\left(1 + \exp(-y_i X_{i*}^\top \theta)\right) \right].$$

For notational simplicity, we denote the logistic loss function in (4.2) as $\mathcal{L}(\theta)$, and define $\widetilde{\mathcal{L}}_\lambda(\theta) = \mathcal{L}(\theta) + \mathcal{H}_\lambda(\theta)$. Then similar to sparse linear regression, we write $\mathcal{F}_\lambda(\theta)$ as

$$\mathcal{F}_\lambda(\theta) = \mathcal{L}(\theta) + \mathcal{R}_\lambda(\theta) = \widetilde{\mathcal{L}}_\lambda(\theta) + \lambda\|\theta\|_1.$$

The logistic loss function is twice differentiable with

$$\nabla \mathcal{L}(\theta) = \frac{1}{n} \sum_{i=1}^n [1 - \pi_i(\theta)] y_i X_{i*} \quad \text{and} \quad \nabla^2 \mathcal{L}(\theta) = \frac{1}{n} X^\top P X,$$

where $P = \text{diag}([1 - \pi_1(\theta)]\pi_1(\theta), ..., [1 - \pi_n(\theta)]\pi_n(\theta)) \in \mathbb{R}^{n \times n}$. Similar to sparse linear regression, we also assume that the design matrix $X$ satisfies the column normalization condition $\|X_{*j}\|_2 = \sqrt{n}$ for all $j = 1, ..., d$.

4.1. *Proximal Coordinate Gradient Descent.* For sparse logistic regression, directly taking the minimum with respect to a selected coordinate does not admit a closed form solution, and therefore may involve some sophisticated algorithm such as the root-finding method.

To address this issue, Razaviyayn et al. (2013) suggest a more convenient approach, which takes a proximal coordinate gradient descent iteration. For example, we select a coordinate $j$ at the $t$-th iteration and consider a quadratic approximation of $\mathcal{F}_\lambda(\theta_j; \theta_{\setminus j}^{(t)})$,

$$\mathcal{Q}_{\lambda,j,L}(\theta_j; \theta^{(t)}) = \mathcal{V}_{\lambda,j,L}(\theta_j; \theta^{(t)}) + \lambda|\theta_j| + \lambda\|\theta_{\setminus j}^{(t)}\|_1,$$

where $L > 0$ is a step size parameter, and $\mathcal{V}_{\lambda,j,L}(\theta_j; \theta^{(t)})$ is defined as

$$\mathcal{V}_{\lambda,j,L}(\theta_j; \theta^{(t)}) = \widetilde{\mathcal{L}}_\lambda(\theta^{(t)}) + (\theta_j - \theta_j^{(t)})\nabla_j\widetilde{\mathcal{L}}_\lambda(\theta^{(t)}) + \frac{L}{2}(\theta_j - \theta_j^{(t)})^2.$$

Here we choose the step size parameter $L$ such that $\mathcal{Q}_{\lambda,j,L}(\theta_j; \theta^{(t)}) \geq \mathcal{F}_\lambda(\theta_j, \theta_{\setminus j}^{(t)})$ for all $j = 1,...d$. We then take

$$(4.3) \qquad \theta_j^{(t+1)} = \underset{\theta_j}{\operatorname{argmin}}\, \mathcal{Q}_{\lambda,j,L}(\theta_j; \theta^{(t)}) = \underset{\theta_j}{\operatorname{argmin}}\, \mathcal{V}_{\lambda,j,L}(\theta_j; \theta^{(t)}) + \lambda|\theta_j|.$$

Different from the exact coordinate minimization, (4.3) always has a closed form solution obtained by soft thresholding. Particularly, we define $\widetilde{\theta}_j^{(t)} = \theta_j^{(t)} - \nabla_j\widetilde{\mathcal{L}}_\lambda(\theta^{(t)})/L$. Then we have

$$\theta_j^{(t+1)} = \underset{\theta_j}{\operatorname{argmin}}\, \frac{1}{2}(\theta_j - \widetilde{\theta}_j^{(t)})^2 + \frac{\lambda}{L}|\theta_j| = \mathcal{S}_{\lambda/L}(\widetilde{\theta}_j^{(t)}) \quad \text{and} \quad \theta_{\setminus j}^{(t+1)} = \theta_{\setminus j}^{(t)}.$$

For notational convenience, we write $\theta_j^{(t+1)} = \mathcal{T}_{\lambda,j,L}(\theta^{(t)})$. When applying PICASSO to solve sparse logistic regression, we only need to replace $\mathcal{T}_{\lambda,j}(\cdot)$ with $\mathcal{T}_{\lambda,j,L}(\cdot)$ in Algorithms 1-3.

**Remark 4.1.** For sparse logistic regression, we have $\nabla_{jj}^2\mathcal{L}(\theta) = \frac{1}{n}X_{*j}^\top P X_{*j}$. Since $P$ is a diagonal matrix, and $\pi_i(\theta) \in (0,1)$ for any $\theta \in \mathbb{R}^d$, we have $\|P\|_2 = \max_i P_{ii} \in (0, 1/4]$ for all $i = 1,...,n$. Then we have $X_{*j}^\top P X_{*j} \leq \|P\|_2\|X_{*j}\|_2^2 \leq n/4$, where the last inequality comes from the column normalization condition of $X$. Thus, we choose $L = \sup_\theta \max_j \nabla_{jj}^2\mathcal{L}(\theta) = 1/4$.

We then analyze the computational and statistical properties of the estimator obtained by PICASSO for sparse logistic regression.

4.2. *Convex Relaxation based Warm Start Initialization.* We assume that $\|\theta^*\|_0 \leq s^*$, and for any $v \neq 0$ and any $\theta$ such that $\|\theta - \theta^*\|_2 \leq R$, we have

$$(4.4) \quad \psi_\ell \|v\|_2^2 - \gamma_\ell \sqrt{\frac{\log d}{n}} \|v\|_1^2 \leq v^\top \nabla^2 \mathcal{L}(\theta) v \leq \psi_u \|v\|_2^2 + \gamma_u \sqrt{\frac{\log d}{n}} \|v\|_1^2,$$

where $\gamma_\ell$, $\gamma_u$, $\psi_\ell$, $\psi_u$, and $R$ are positive constants, and do not scale with $(s^*, n, d)$. Existing literature has shown that many common examples of sub-Gaussian random design satisfy (4.4) with high probability (Raskutti et al., 2010; Negahban et al., 2012; Loh and Wainwright, 2015).

Similar to sparse linear regression, we need to verify Assumptions 3.1 and 3.5 for sparse logistic regression by the following lemma.

**Lemma 4.2.** Suppose (4.4) holds. Given $\lambda_N = 16\sqrt{\log d/n}$, we have

$$\mathbb{P}\left(\lambda_N \geq 4\|\nabla \mathcal{L}(\theta^*)\|_\infty = \frac{4}{n}\|X^\top w\|_\infty\right) \geq 1 - d^{-7},$$

where $w = ([1 - \pi_1(\theta^*)]y_1, ..., [1 - \pi_n(\theta^*)]y_n)^\top$ with $\pi_i(\theta)$'s defined in (4.1). Moreover, given $\gamma \geq 4/\psi_\ell$ and $\|\theta - \theta^*\|_2 \leq R$, there exists some generic constant $C_1$ such that for large enough $n$, we have

$$\widetilde{s} = C_1 s^* \geq [484\kappa^2 + 100\kappa]s^*, \ \widetilde{\rho}_-(s^* + 2\widetilde{s}) \geq \frac{\psi_\ell}{2}, \ \rho_+(s^* + 2\widetilde{s}) \leq \frac{5\psi_u}{4}.$$

The proof of Lemma 4.2 directly follows Appendix E.2 and Loh and Wainwright (2015), and therefore is omitted. Lemma 4.2 guarantees that the regularization sequence satisfies Assumption 3.1 with high probability, and Assumption 3.5 holds when the design matrix satisfies (4.4).

Different from sparse linear regression, however, the restricted convexity and smoothness only hold over an $\ell_2$ ball centered at $\theta^*$ for sparse logistic regression. Thus, directly choosing $\widehat{\theta}^{\{0\}} = 0$ may violate the restricted strong convexity. A simple counter example is $\|\theta^*\|_2 > R$, which results in $\|0 - \theta^*\|_2 > R$. To address this issue, we propose a new convex relaxation based warm start initialization to obtain an initial solution for $\lambda_0$. Particularly, we solve the following convex relaxation of (1.1):

$$(4.5) \qquad \min_{\theta \in \mathbb{R}^d} \widetilde{\mathcal{F}}_{\lambda_0}(\theta), \ \text{ where } \widetilde{\mathcal{F}}_{\lambda_0}(\theta) = \mathcal{L}(\theta) + \lambda_0 \|\theta\|_1$$

up to an adequate precision. For example, we choose $\theta^{\mathsf{relax}}$ satisfying the approximate KKT condition of (4.5) as follows,

$$(4.6) \qquad \min_{\xi \in \partial \|\theta^{\mathsf{relax}}\|_1} \|\nabla \mathcal{L}(\theta^{\mathsf{relax}}) + \lambda_0 \xi\|_\infty \leq \delta_0 \lambda_0,$$

where $\delta_0 \in (0,1)$ is the initial precision parameter for $\lambda_0$. Since $\delta_0$ in (4.6) can be chosen as a sufficiently large value (e.g., $\delta_0 = 1/8$), computing $\theta^{\text{relax}}$ becomes very efficient even for algorithms with only sublinear rates of convergence to global optima, e.g., classical coordinate minimization and proximal gradient algorithms. For notational convenience, we call the above initialization procedure the convex relaxation based warm initialization.

**Lemma 4.3.** Suppose Assumption 3.5 holds only for $\|\theta - \theta^*\|_2 \leq R$. Given $\rho_-(s^* + \widetilde{s})R \geq 9\lambda_0\sqrt{s^*} \geq 18\lambda_N\sqrt{s^*}$ and $\delta_0 = 1/8$, we have

$$\|\theta^{\text{relax}}_{\overline{\mathcal{S}}}\|_0 \leq \widetilde{s}, \quad \|\theta^{\text{relax}} - \theta^*\|_2 \leq R, \quad \text{and} \quad \mathcal{F}_{\lambda_0}(\theta^{\text{relax}}) \leq \mathcal{F}_{\lambda_0}(\theta^*) + \triangle_{\lambda_0}.$$

Lemma 4.3 guarantees that $\theta^{\text{relax}}$ is a proper initial solution for $\lambda_0$. Thus, all convergence analysis in Theorem 3.12 directly follows, and PICASSO attains a linear convergence to a unique sparse local optimum with high probability. The statistical properties can also be established accordingly. An illustration of the convex relaxation based warm start initialization is provided in Figure 7.

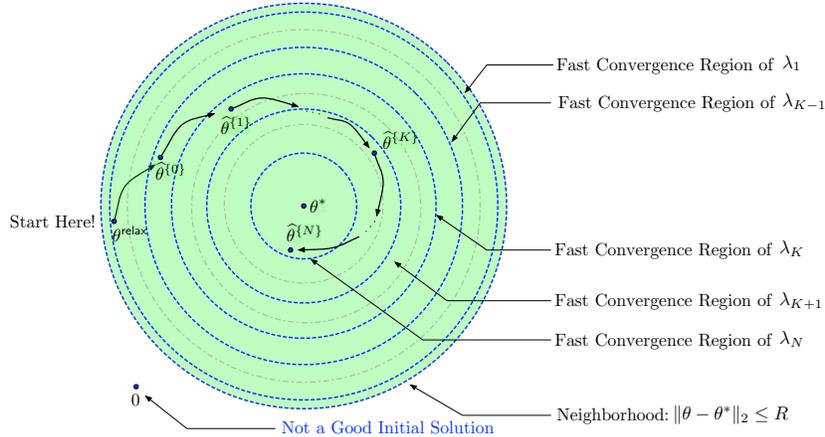

Fig 7. *An illustration of the convex relaxation based warm start initialization. When the restricted convexity and smoothness only hold over a neighborhood around $\theta^*$ (Green Region). Directly choosing $0$ as the initial solution may violate the restricted strong convexity. Thus, we adopt a convex relaxation approach to obtain an initial solution, which is ensured to be sparse and belong to the desired neighborhood.*

**5. Numerical Experiments.** We evaluate the computational and statistical performance of PICASSO through numerical simulations. We compare PICASSO with five competitors: (1) SparseNet (Mazumder et al., 2011); (2) Path-following Iterative Shrinkage Thresholding Algorithm (PISTA, Wang et al. (2014)); (3) Accelerated PISTA (A-PISTA, Zhao and Liu (2016));

(4) Multistage Convex Relaxation Method (Mcvx, Zhang (2013)); (5) Local Linear Approximation (LLA, Zou and Li (2008)). Note that each subproblem of Mcvx and LLA is solved by proximal gradient algorithms with backtracking line search.

All experiments are conducted on a PC with Intel Core i5 3.3 GHz and 16GB memory. All programs are coded in double precision C, called from an R wrapper. We optimize the computation by exploiting the vector and matrix sparsity, which gains a significant speedup in vector and matrix manipulations (e.g., computing the gradient and evaluating the objective value). We apply PICASSO to sparse linear regression with the MCP regularizer.

We generate each row of the design matrix $X_{i*}$ independently from a $d$-dimensional Gaussian distribution with mean 0 and covariance matrix $\Sigma \in \mathbb{R}^{d \times d}$, where $\Sigma_{kj} = 0.75$ and $\Sigma_{kk} = 1$ for all $j, k = 1, ..., d$ and $k \neq j$. We then normalize each column of the design matrix $X_{*j}$ such that $\|X_{*j}\|_2 = \sqrt{n}$. The response vector is generated from the linear model $y = X\theta^* + \varepsilon$, where $\theta^* \in \mathbb{R}^d$ is the regression coefficient vector, and $\varepsilon$ is generated from a $n$-dimensional Gaussian distribution with mean 0 and covariance matrix $\sigma^2 I_n$. We set $n = 300$, $d = 18000$, $s^* = 18$, and $\sigma^2 = 4$. $\theta^*$ has 18 nonzero entries, which are $\theta^*_{1000} = \theta^*_{7000} = \theta^*_{13000} = 3$, $\theta^*_{2000} = \theta^*_{8000} = \theta^*_{14000} = 2$, $\theta^*_{3000} = \theta^*_{9000} = \theta^*_{15000} = 1.5$, $\theta^*_{4000} = \theta^*_{10000} = \theta^*_{16000} = -3$, $\theta^*_{5000} = \theta^*_{11000} = \theta^*_{17000} = -2$, and $\theta_{6000} = \theta^*_{12000} = \theta_{18000} = -1.5$ for $k = 0, ..., 2$. We then set $\gamma = 1.25$, $N = 70$, $\lambda_N = 0.25\sigma\sqrt{\log d/n}$, $\varphi = 0.05$, $\delta_K = 10^{-3}$, and $\tau_K = 10^{-6}$ for all $1 \leq K \leq N$.

We present the numerical results averaged over 1000 simulations. Specifically, we create a validation set using the same design matrix as the training set for regularization parameter selection. We then tune the regularization parameter over the selected regularization sequence. We denote the response vector of the validation set as $\widetilde{y} \in \mathbb{R}^n$. Let $\widehat{\theta}^\lambda$ denote the obtained estimator using the regularization parameter $\lambda$. We then choose the optimal regularization parameter $\widehat{\lambda}$ by

$$\widehat{\lambda} = \operatorname{argmin}_{\lambda \in \{\lambda_1,...,\lambda_N\}} \|\widetilde{y} - X\widehat{\theta}^\lambda\|_2^2.$$

We repeat the simulation for 1000 times and summarize the averaged results in Table 1. In terms of timing performance, PICASSO slightly outperforms SparseNet, outperforms A-PISTA, and greatly outperforms PISTA, LLA, and Mcvx. In terms of support recovery and parameter estimation, PICASSO slightly outperforms A-PISTA, PISTA, and Mcvx, and greatly outperforms SparseNet and LLA.

To further demonstrate the superiority of PICASSO, we present a typical failure example of SparseNet using the heuristic cyclic selection rule. This

Table 1

*Quantitative comparison on the simulated data set ($n = 300$, $d = 18000$, $s^* = 18$, $\sigma^2 = 4$). In terms of timing performance, PICASSO slightly outperforms SparseNet, outperforms A-PISTA, and greatly outperforms PISTA, LLA, and Mcvx. In terms of support recovery and parameter estimation, PICASSO slightly outperforms A-PISTA, PISTA, and Mcvx, and greatly outperforms SparseNet and LLA.*

| Method | $\|\widehat{\theta} - \theta^*\|_2$ | $\|\widehat{\theta}_{\mathcal{S}}\|_0$ | $\|\widehat{\theta}_{\mathcal{S}^c}\|_0$ | Correct | Timing |
|---|---|---|---|---|---|
| PICASSO | 1.258(0.515) | 17.79(0.54) | 0.48(0.52) | 616/1000 | 1.062(0.084) |
| SparseNet | 1.602(0.791) | 17.64(0.85) | 2.07(1.41) | 248/1000 | 1.109(0.088) |
| PISTA | 1.267(0.528) | 17.76(0.54) | 0.55(0.51) | 614/1000 | 52.358(5.920) |
| A-PISTA | 1.276(0.530) | 17.76(0.54) | 0.57(0.57) | 613/1000 | 6.358(0.865) |
| Mcvx | 1.293(0.529) | 17.76(0.52) | 0.58(0.52) | 615/1000 | 67.247(7.128) |
| LLA | 1.517(0.949) | 17.50(0.61) | 1.28(0.85) | 365/1000 | 31.247(3.870) |

example is chosen from our 1000 simulations, and illustrated in Figure 8. We see that the heuristic cyclic selection rule in SparseNet always needs to iterate over many irrelevant variables before getting to the relevant variable when identifying a new active set. Since these irrelevant variables are highly correlated with the relevant variables in our experiment, the heuristic cyclic selection rule tends to overselect the irrelevant variables and miss some relevant variables. In contrast, PICASSO, PISTA, and A-PISTA have mechanisms to prevent from overselecting the irrelevant coordinates when identifying active sets. This eventually makes them outperform SparseNet in both parameter estimation and support recovery. Moreover, we also see that PISTA is much slower than other algorithms, because PISTA needs to calculate a full gradient and conduct a sophisticated line search in every iteration, which are computationally expensive. Though A-PISTA adopts the coordinate minimization to further accelerate PISTA, it still suffers from the computationally expensive line search when identifying active sets. This eventually leads to less competitive timing performance than PICASSO.

**6. Discussions and Future Work.** Here we discuss several existing methods related to PICASSO, including the multistage convex relaxation method (Mcvx), local linear approximation method (LLA), path-following iterative shrinkage thresholding algorithm (PISTA), accelerated path-following iterative shrinkage thresholding algorithm (A-PISTA), and proximal gradient algorithm.

The multistage convex relaxation method is proposed in Zhang (2013). It solves a sequence of convex relaxation problems of (1.1). Zhang (2013) show that the obtained estimator enjoys similar statistical guarantees to those of PICASSO for sparse linear regression. However, there is only sublinear guarantee on its convergence rate to a local optimum. Moreover, since each relaxed problem is still lack of strong convexity, the multistage convex relaxation method needs to be combined with some efficient computational algorithms such as PICASSO.

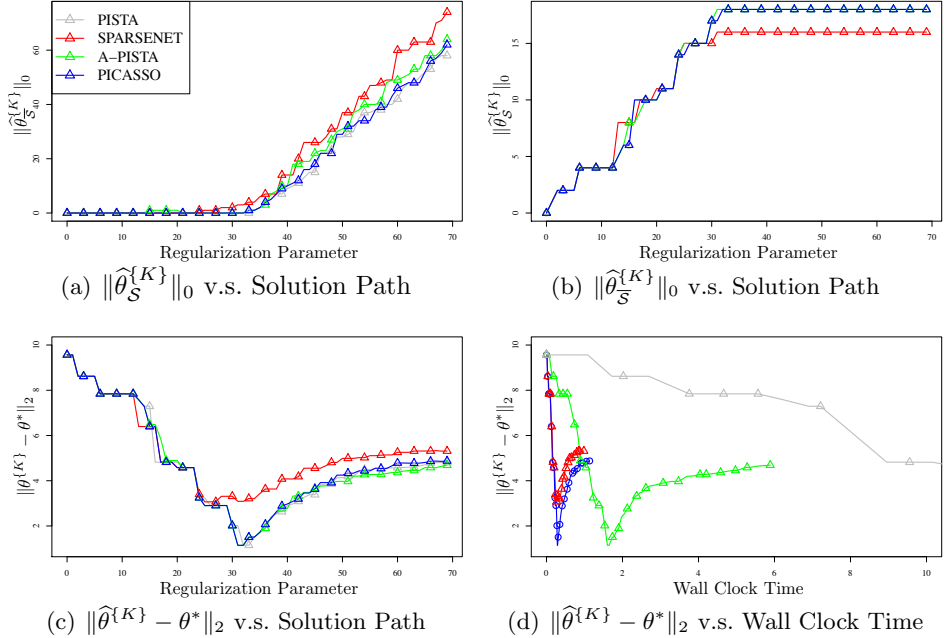

FIG 8. *A typical failure example of SparseNet using the heuristic cyclic selection rule, which is chosen from our 1000 simulations. We see that cyclic selection rule tends to overselect the irrelevant coordinates and miss some relevant coordinates when updating the active set. Thus SparseNet eventually yields denser solutions with worse performance in parameter estimation and support recovery than PICASSO, PISTA, and A-PISTA.*

The local linear approximation method is proposed in Zou and Li (2008); Wang et al. (2013); Fan et al. (2014). It is essentially a special case of the multistage convex relaxation with only two iterations. Similar to the multistage convex relaxation method, it also needs an efficient computational algorithm to solve each relaxed problem. Moreover, in order to obtain the variable selection consistency, the local linear approximation method requires a stronger minimum signal strength. Taking sparse linear regression as an example, Wang et al. (2013); Fan et al. (2014) requires a minimum signal strength of order of $\sigma\sqrt{s^* \log d/n}$, while PICASSO only requires a minimum signal strength of order of $\sigma\sqrt{\log d/n}$.

The path-following iterative shrinkage thresholding algorithm (PISTA)

is proposed in Wang et al. (2014). PISTA is essentially a proximal gradient algorithm combined with the warm start initialization. PISTA needs to calculate the entire ($d$-dimensional) gradient vector and requires a sophisticated backtracking line search procedure in every iteration. Thus, PICASSO is computationally much more efficient and scalable than PISTA in practice, although PISTA and PICASSO enjoy similar theoretical guarantees. Besides, the implementation of PISTA requires subtle control over the step size, and often yield slow empirical convergence. An accelerated PISTA algorithm (A-PISTA) is proposed in Zhao and Liu (2016), which uses coordinate minimization algorithms to accelerated PISTA. It shows an improved computational performance over PISTA in our numerical simulations, but not as competitive as PICASSO.

Moreover, when extending PISTA to general loss functions, Wang et al. (2014) propose a contained formulation. Particularly, they solve (1.1) with an additional constraint

$$(6.1) \quad \min_{\theta \in \mathbb{R}^d} \mathcal{L}(\theta) + \mathcal{R}_\lambda(\theta) \quad \text{subject to } \|\theta\|_2 \leq R/2.$$

The additional constraint guarantees that the solution always stays in the restricted strongly convex region (a small neighborhood around $\theta^*$), only under the assumption $\|\theta^*\|_2 \leq R/2$, where $R$ is a constant and cannot scale with $(n, s^*, d)$. This assumption is very restrictive, and also introduces an additional tuning parameter. In contrast, our proposed convex relaxation based warm start initialization avoids this assumption, and allows $\|\theta^*\|_2$ to be arbitrarily large. Furthermore, we want to emphasize that PISTA exploits an explicit soft-thresholding procedure to directly control the solution sparsity in each iteration, while PICASSO adopts an algorithmic strategy to control the sparsity of the active set.

Other researchers focus on solving (1.1) with an additional constraint,

$$(6.2) \quad \min_{\theta \in \mathbb{R}^d} \mathcal{L}(\theta) + \mathcal{R}_\lambda(\theta) \quad \text{subject to } \|\theta\|_1 \leq M,$$

where $M > 0$ is an extra tuning parameter. Loh and Wainwright (2015) show that the proximal gradient algorithm attains a linear convergence to a ball centered at $\theta^*$ to (6.2) with a radius approximately equal to the statistical error. However, the analysis of Loh and Wainwright (2015) does not justify the advantage of nonconvex regularization: They only provides a slower statistical rate of convergence than PICASSO in parameter estimation for their obtained estimator, and no support recovery guarantee is established. Besides, their analysis for general loss functions also requires the restrictive assumption: $\|\theta^*\|_2 \leq R/2$, where $R$ is a constant and does not scale with $(n, s^*, d)$. Nevertheless, PICASSO does not require this assumption.

For future work, we are interested in possible extensions: (I) Extension to more complicated regularizers such as grouping regularizers for variable clustering; (II) Extension to more complicated (possibly nonconvex) loss functions such as sparse phase retrieval and sparse coding problems; (III) Extension to asynchronous parallel optimization setting with shared memory or communication-efficient distributed optimization setting; (IV) Extension to second order algorithms such as the regularized iterative reweighed least square optimization algorithm for sparse generalized linear model estimation (proximal Newton). These extensions will lead to more efficient and scalable coordinate optimization algorithms for more sophisticated nonconvex optimization problems.

**7. Proof of Main Results.** We present the proof sketch of our computational and statistical theories. Some lemmas are deferred to the appendix. To unify the convergence analysis of PICASSO using the exact coordinate minimization (1.6) and proximal coordinate gradient descent (4.3), we define two auxiliary parameters $\nu_+(1)$ and $\nu_-(1)$. Specifically, we choose $\nu_+(1) = \nu_-(1) = L$ for the proximal coordinate gradient descent, and $\nu_+(1) = \rho_+(1)$ and $\nu_-(1) = \widetilde{\rho}_-(1)$ for the exact coordinate minimization.

7.1. *Proof of Theorem 3.9.*

PROOF. Since $\|\theta^{(0)}\|_0 = s \leq s^* + 2\widetilde{s}$, by Assumption 3.5 and Lemma 3.4, we know that (2.1) is a strongly convex optimization problem. Thus, its minimizer $\overline{\theta}$ is unique. We then introduce the following lemmas.

**Lemma 7.1.** Suppose Assumption and 3.5 holds, and $|\mathcal{A}| = s \leq s^* + 2\widetilde{s}$. For $t = 0, 1, 2, ...$, we have $\mathcal{F}_\lambda(\theta^{(t)}) - \mathcal{F}_\lambda(\theta^{(t+1)}) \geq \frac{\nu_-(1)}{2}\|\theta^{(t)} - \theta^{(t+1)}\|_2^2$.

**Lemma 7.2.** Suppose Assumption and 3.5 holds, and $|\mathcal{A}| = s \leq s^* + 2\widetilde{s}$. For $t = 0, 1, 2, ...$, we have $\mathcal{F}_\lambda(\theta^{(t+1)}) - \mathcal{F}_\lambda(\overline{\theta}) \leq \frac{s\rho_+^2(s)}{2\widetilde{\rho}_-(s)}\|\theta^{(t+1)} - \theta^{(t)}\|_2^2$.

Lemmas 7.1 and 7.2 characterize the successive descent and the gap towards the optimal objective value after each iteration respectively.

[Linear Convergence] Combining Lemmas 7.1 and 7.2, we obtain

$$(7.1) \quad \mathcal{F}_\lambda(\theta^{(t+1)}) - \mathcal{F}_\lambda(\overline{\theta}) \leq \frac{s\rho_+^2(s)}{\widetilde{\rho}_-(s)\nu_-(1)}[\mathcal{F}_\lambda(\theta^{(t)}) - \mathcal{F}_\lambda(\overline{\theta})] \\ - \frac{s\rho_+^2(s)}{\widetilde{\rho}_-(s)\nu_-(1)}[\mathcal{F}_\lambda(\theta^{(t+1)}) - \mathcal{F}_\lambda(\overline{\theta})].$$

By simple manipulation, (7.1) implies

$$(7.2) \quad \mathcal{F}_\lambda(\theta^{(t+1)}) - \mathcal{F}_\lambda(\bar{\theta}) \overset{(i)}{\leq} \left(\frac{s\rho_+^2(s)}{\widetilde{\rho}_-(s)\nu_-(1) + s\rho_+^2(s)}\right)[\mathcal{F}_\lambda(\theta^{(t)}) - \mathcal{F}_\lambda(\bar{\theta})]$$

$$\overset{(ii)}{\leq} \left(\frac{s\rho_+^2(s)}{\widetilde{\rho}_-(s)\nu_-(1) + s\rho_+^2(s)}\right)^{t+1}[\mathcal{F}_\lambda(\theta^{(0)}) - \mathcal{F}_\lambda(\bar{\theta})],$$

where (ii) comes from recursively using (i).

[Number of Iterations] Combining (7.2) with Lemma 7.1, we obtain

$$\|\theta^{(t)} - \theta^{(t+1)}\|_2^2 \overset{(i)}{\leq} \frac{2[\mathcal{F}_\lambda(\theta^{(t)}) - \mathcal{F}_\lambda(\bar{\theta})]}{\nu_-(1)}$$

$$\leq \left(\frac{s\rho_+^2(s)}{\widetilde{\rho}_-(s)\nu_-(1) + s\rho_+^2(s)}\right)^t \frac{2[\mathcal{F}_\lambda(\theta^{(0)}) - \mathcal{F}_\lambda(\bar{\theta})]}{\nu_-(1)},$$

where (i) comes from $\mathcal{F}_\lambda(\theta^{(t)}) \geq \mathcal{F}_\lambda(\bar{\theta})$. Thus, we need at most

$$t = \log^{-1}\left(\frac{s\rho_+^2(s)}{\widetilde{\rho}_-(s)\nu_-(1) + s\rho_+^2(s)}\right) \log\left(\frac{\nu_-(1)\tau^2\lambda^2}{2[\mathcal{F}_\lambda(\theta^{(0)}) - \mathcal{F}_\lambda(\bar{\theta})]}\right)$$

iterations such that

$$\|\theta^{(t+1)} - \theta^{(t)}\|_2^2 \leq \left(\frac{s\rho_+^2(s)}{\widetilde{\rho}_-(s)\nu_-(1) + s\rho_+^2(s)}\right)^t \frac{2[\mathcal{F}_\lambda(\theta^{(0)}) - \mathcal{F}_\lambda(\bar{\theta})]}{\nu_-(1)} \leq \tau^2\lambda^2.$$

□

7.2. *Proof of Theorem 3.10.*

PROOF. Before we start the proof, we introduce the following lemmas.

**Lemma 7.3.** Suppose Assumptions 3.1, 3.5, and 3.7 hold. There exists a unique sparse local optimum $\bar{\theta}^\lambda$ satisfying $\|\bar{\theta}_{\overline{\mathcal{S}}}^\lambda\|_0 \leq \widetilde{s}$ and $\mathcal{K}_\lambda(\bar{\theta}^\lambda) = 0$.

**Lemma 7.4.** Suppose Assumptions 3.1, 3.5, and 3.7 hold. If the initial solution $\theta^{(0)}$ in Algorithm 1 satisfies $\|\theta_{\overline{\mathcal{S}}}^{(0)}\|_0 \leq 2\widetilde{s}$ and $\mathcal{F}_\lambda(\theta^{(0)}) \leq \mathcal{F}_\lambda(\theta^*) + \triangle_\lambda$, the output solution $\widehat{\theta}$ satisfies

$$(7.3) \quad \min_{\xi_\mathcal{A} \in \partial\|\widehat{\theta}_\mathcal{A}\|_1} \|\nabla_\mathcal{A}\widetilde{\mathcal{L}}_\lambda(\widehat{\theta}) + \lambda\xi_\mathcal{A}\|_\infty \leq \delta\lambda \quad \text{and} \quad \|\widehat{\theta}_{\overline{\mathcal{S}}}\|_0 \leq \widetilde{s}.$$

**Lemma 7.5.** Suppose Assumptions 3.1, 3.5, and 3.7 hold. If the initial solution $\theta^{[0]}$ satisfies $\|\theta_{\overline{\mathcal{S}}}^{[0]}\|_0 \leq \widetilde{s}$ and $\mathcal{F}_\lambda(\theta^{[0]}) \leq \mathcal{F}_\lambda(\theta^*) + \triangle_\lambda$. Then regardless the simple rule or strong rule, we have $|\mathcal{A}_0 \cap \overline{\mathcal{S}}| \leq \widetilde{s}$.

The proofs of Lemmas 7.3, 7.4, and 7.5 are provided in Appendices C.5, C.6, and C.8 respectively. Lemma 7.3 verifies the existence of the unique sparse local optimum. Lemma 7.4 implies that the inner loop of PICASSO removes some irrelevant coordinates, and encourages the output solution sparsity. Lemma 7.5 implies that the initial active set is sufficiently sparse for both simple and strong rules.

[Solution Sparsity] Since the objective always decreases, we have

$$(7.4) \quad \mathcal{F}_\lambda(\theta^{[m+1]}) \leq \mathcal{F}_\lambda(\theta^{[m+0.5]}) \leq \mathcal{F}_\lambda(\theta^{[0]}) \leq \mathcal{F}_\lambda(\theta^*) + \triangle_\lambda$$

for all $m = 0, 1, 2, \ldots$. Since $\theta^{[0]}$ satisfies $\|\theta^{[0]}_{\overline{\mathcal{S}}}\|_0 \leq \widetilde{s}$, by Lemma 7.5, we have $|\mathcal{A}_0 \cap \overline{\mathcal{S}}| \leq \widetilde{s}$. Then by Lemma 7.4, we have $\|\theta^{[0.5]}_{\overline{\mathcal{S}}}\|_0 \leq \widetilde{s}$. Moreover, the greedy selection rule moves only one inactive coordinate to the active set, and therefore guarantees $\|\theta^{[1]}_{\overline{\mathcal{S}}}\|_0 \leq \widetilde{s}+1$. By induction, we prove $\|\theta^{[m]}_{\overline{\mathcal{S}}}\|_0 \leq \widetilde{s}+1$ and $\|\theta^{[m+0.5]}_{\overline{\mathcal{S}}}\|_0 \leq \widetilde{s}$ for all $m = 0, 1, 2, \ldots$.

[Linear Convergence] We first prove the linear convergence for the proximal coordinate gradient descent. We need to construct an auxiliary solution

$$\begin{aligned} w^{[m+1]} &= \underset{w \in \mathbb{R}^d}{\operatorname{argmin}} \, \mathcal{J}_{\lambda,L}(w; \theta^{[m+0.5]}) \\ &= \underset{w \in \mathbb{R}^d}{\operatorname{argmin}} \, \widetilde{\mathcal{L}}_\lambda(\theta^{[m+0.5]}) + (w - \theta^{[m+0.5]})^\top \nabla \widetilde{\mathcal{L}}_\lambda(\theta^{[m+0.5]}) \\ &\qquad\qquad + \frac{L}{2}\|w - \theta^{[m+0.5]}\|_2^2 + \lambda\|w\|_1. \end{aligned}$$

We can verify $w^{[m+1]}_k = \operatorname{argmin}_{\theta_k} \mathcal{Q}_{\lambda,k,L}(\theta_k; \theta^{[m+0.5]})$ for $j = 1, \ldots, d$. For notational simplicity, we define $w^{[m+1]} = \mathcal{T}_{\lambda,L}(\theta^{[m+0.5]})$. Before we proceed, we introduce the following lemmas.

**Lemma 7.6.** Suppose Assumptions 3.1, 3.5, and 3.7 hold. For the proximal coordinate gradient descent and $m = 0, 1, 2\ldots$, we have

$$\mathcal{F}_\lambda(\theta^{[m+0.5]}) - \mathcal{F}_\lambda(\theta^{[m+1]}) \geq \frac{1}{s^* + 2\widetilde{s}} \left[ \mathcal{F}_\lambda(\theta^{[m+0.5]}) - \mathcal{J}_{\lambda,L}(w^{[m+1]}; \theta^{[m+0.5]}) \right].$$

**Lemma 7.7.** Suppose Assumptions 3.1, 3.5, and 3.7 hold. For the proximal coordinate gradient descent and $m = 0, 1, 2\ldots$, we have

$$\mathcal{F}_\lambda(\theta^{[m+0.5]}) - \mathcal{F}_\lambda(\overline{\theta}^\lambda) \leq \frac{L}{\widetilde{\rho}_-(s^* + 2\widetilde{s})} \left[ \mathcal{F}_\lambda(\theta^{[m+0.5]}) - \mathcal{J}_{\lambda,L}(w^{[m+1]}; \theta^{[m+0.5]}) \right].$$

The proofs of Lemmas 7.6 and 7.7 are presented in Appendices C.9 and C.12 respectively. Lemmas 7.6 and 7.7 characterize the successive descent in each iteration and the gap towards the optimal objective value after each iteration respectively. Combining Lemmas 7.6 and 7.7, we obtain

$$(7.5) \quad \mathcal{F}_\lambda(\theta^{[m+0.5]}) - \mathcal{F}_\lambda(\overline{\theta}^\lambda)$$
$$\leq \frac{(s^* + 2\widetilde{s})L}{\widetilde{\rho}_-(s^* + 2\widetilde{s})} \left( [\mathcal{F}_\lambda(\theta^{[m+0.5]}) - \mathcal{F}_\lambda(\overline{\theta}^\lambda)] - [\mathcal{F}_\lambda(\theta^{[m+1]}) - \mathcal{F}_\lambda(\overline{\theta}^\lambda)] \right).$$

By simple manipulation, (7.5) implies

$$(7.6) \quad \mathcal{F}_\lambda(\theta^{[m+1]}) - \mathcal{F}_\lambda(\overline{\theta}^\lambda) \leq \left(1 - \frac{\widetilde{\rho}_-(s^* + 2\widetilde{s})}{(s^2 + 2\widetilde{s})L}\right) [\mathcal{F}_\lambda(\theta^{[m+0.5]}) - \mathcal{F}_\lambda(\overline{\theta}^\lambda)]$$
$$\overset{(i)}{\leq} \left(1 - \frac{\widetilde{\rho}_-(s^* + 2\widetilde{s})}{(s^* + 2\widetilde{s})L}\right) [\mathcal{F}_\lambda(\theta^{[m]}) - \mathcal{F}_\lambda(\overline{\theta}^\lambda)]$$
$$\overset{(ii)}{\leq} \left(1 - \frac{\widetilde{\rho}_-(s^* + 2\widetilde{s})}{(s^* + 2\widetilde{s})L}\right)^{m+1} [\mathcal{F}_\lambda(\theta^{[0]}) - \mathcal{F}_\lambda(\overline{\theta}^\lambda)],$$

where (i) comes from (7.4), and (ii) comes from recursively applying (i).

For the exact coordinate minimization, at the $m$-th iteration, we only need to conduct a proximal coordinate gradient descent iteration with $L = \rho_+(1)$, and obtain an auxiliary solution $\widetilde{\theta}^{[m+1]}$. Since $\mathcal{F}_\lambda(\theta^{[m+1]}) \leq \mathcal{F}_\lambda(\widetilde{\theta}^{[m+1]})$, by (7.6), we further have

$$(7.7) \quad \mathcal{F}_\lambda(\theta^{[m+1]}) - \mathcal{F}_\lambda(\overline{\theta}^\lambda) \leq \left(1 - \frac{\widetilde{\rho}_-(s^* + 2\widetilde{s})}{(s^* + 2\widetilde{s})\rho_+(1)}\right) \left[\mathcal{F}_\lambda(\theta^{[m]}) - \mathcal{F}_\lambda(\overline{\theta}^\lambda)\right].$$

[Number of Iterations] Before we proceed, we introduce the following lemma.

**Lemma 7.8.** Suppose Assumption 3.5 holds. For any $\theta$, we conduct an exact coordinate minimization or proximal coordinate gradient descent iteration over a coordinate $k$, and obtain $w$. Then we have $\mathcal{F}_\lambda(\theta) - \mathcal{F}_\lambda(w) \geq \frac{\nu_-(1)}{2}(w_k - \theta_k)^2$. Moreover, if $\theta_k = 0$ and $|\nabla_k \mathcal{L}(\theta)| \geq (1+\delta)\lambda$, we have

$$|w_k| \geq \frac{\delta\lambda}{L} \quad \text{and} \quad \mathcal{F}_\lambda(\theta) - \mathcal{F}_\lambda(w) \geq \frac{\delta^2\lambda^2}{2\nu_+(1)}.$$

Lemma 7.8 characterizes the sufficient descent when adding the selected inactive coordinate $k$ into the active set. Assume that the selected coordinate $k_m$ satisfies $|\nabla_{k_m}\mathcal{L}(\theta^{[m+0.5]})| \geq (1+\delta)\lambda$. Then by Lemma 7.8, we have

$$(7.8) \quad \mathcal{F}_\lambda(\theta^{[m+0.5]}) - \mathcal{F}_\lambda(\overline{\theta}^\lambda) \geq \mathcal{F}_\lambda(\theta^{[m+0.5]}) - \mathcal{F}_\lambda(\theta^{[m+1]}) \geq \frac{\delta^2\lambda^2}{2\nu_+(1)}.$$

Moreover, by (7.6) and (7.7), we need at most

$$m = \log^{-1}\left(1 - \frac{\widetilde{\rho}_-(s^* + 2\widetilde{s})}{(s^* + 2\widetilde{s})\nu_+(1)}\right)\log\left(\frac{\delta^2\lambda^2}{3\nu_+(1)[\mathcal{F}_\lambda(\theta^{[0]}) - \mathcal{F}_\lambda(\overline{\theta}^\lambda)]}\right)$$

iterations such that $\mathcal{F}_\lambda(\theta^{[m+0.5]}) - \mathcal{F}_\lambda(\overline{\theta}^\lambda) \leq \frac{\delta^2\lambda^2}{3\nu_+(1)}$, which is contradicted by (7.8). Thus, we must have $\max_{k \in \overline{\mathcal{A}}_m} |\nabla_k \mathcal{L}(\theta^{[m+0.5]})| \leq (1+\delta)\lambda$, and the algorithm is terminated.

[Approximately Optimal Output Solution] By Lemma 7.4, we know that when every inner loop terminates, the approximate KKT condition must hold over the active set. Since $\nabla_{\overline{\mathcal{A}}_m}\mathcal{H}_\lambda(\theta^{[m+0.5]}) = 0$, the stopping criterion $\max_{k \in \overline{\mathcal{A}}_m} |\nabla_k \mathcal{L}(\theta^{[m+0.5]})| \leq (1+\delta)\lambda$ implies that the approximate KKT condition holds over the inactive set,

$$\min_{\xi_{\overline{\mathcal{A}}_m} \in \partial\|\theta^{[m+0.5]}_{\overline{\mathcal{A}}_m}\|_1} \|\nabla_{\overline{\mathcal{A}}_m}\widetilde{\mathcal{L}}_\lambda(\theta^{[m+0.5]}) + \lambda\xi_{\overline{\mathcal{A}}_m}\|_\infty \leq \delta\lambda.$$

The above two approximate KKT conditions implies that $\theta^{[m+0.5]}$ satisfies the approximate KKT condition $\mathcal{K}_\lambda(\theta^{[m+0.5]}) \leq \delta\lambda$. □

7.3. *Proof of Theorem 3.12.*

PROOF. [Result (I)] Before we proceed, we introduce the following lemma.

**Lemma 7.9.** Suppose Assumptions 3.1, 3.5, and 3.7 hold. For any $\lambda \geq \lambda_N$, if $\theta$ satisfies $\|\theta_{\overline{S}}\|_0 \leq \widetilde{s}$ and $\mathcal{K}_\lambda(\theta) \leq \delta\lambda$, where $\delta \leq 1/8$, then for any $\lambda' \in [\lambda_N, \lambda]$, we have

$$\mathcal{F}_{\lambda'}(\theta) - \mathcal{F}_{\lambda'}(\overline{\theta}^{\lambda'}) \leq \frac{40(\mathcal{K}_\lambda(\theta) + 3(\lambda - \lambda'))(\lambda + \lambda')s^*}{\widetilde{\rho}_-(s^* + \widetilde{s})}.$$

The proof of Lemma 7.9 is provided in Appendix D.2. If we take $\lambda = \lambda' = \lambda_K$ and $\theta = \widehat{\theta}^{\{K-1\}}$, then Lemma 7.9 implies

(7.9) $$\mathcal{F}_{\lambda_K}(\widehat{\theta}^{\{K-1\}}) - \mathcal{F}_{\lambda_K}(\overline{\theta}^{\lambda_K}) \leq \frac{25s^*\lambda_K^2}{\widetilde{\rho}_-(s^* + \widetilde{s})}.$$

Since the objective value always decreases within each middle loop, for any inner loop with $\lambda_K$, we have $\mathcal{F}_{\lambda_K}(\theta^{(0)}) - \mathcal{F}_{\lambda_K}(\overline{\theta}) \leq \mathcal{F}_{\lambda_K}(\widehat{\theta}^{\{K-1\}}) - \mathcal{F}_{\lambda_K}(\overline{\theta}^{\lambda_K})$. Thus, by Theorem 3.9 and (7.9), we know that the number of iterations within each inner loop is at most

$$\log^{-1}\left(\frac{\widetilde{\rho}_-(s)\nu_-(1) + s\rho_+^2(s)}{s\rho_+^2(s)}\right)\log\left(\frac{\nu_-(1)\tau_K^2\widetilde{\rho}_-(s^* + \widetilde{s})}{25s^*}\right).$$

[Results (II)] Combining Theorem 3.10 with (7.9), we know that the number of active set updating iterations within each middle loop is at most

$$\log^{-1}\left(1 - \frac{\widetilde{\rho}_-(s^* + 2\widetilde{s})}{(s^* + 2\widetilde{s})\nu_+(1)}\right) \log\left(\frac{\delta_K^2 \widetilde{\rho}_-(s^* + \widetilde{s})}{75\nu_+(1)s^*}\right).$$

[Results (III)] For $K < N$, we take $\lambda' = \lambda_N$, $\lambda = \lambda_K$, and $\theta = \widehat{\theta}^{\{K\}}$. Then by Lemma 7.9, we have

$$\mathcal{F}_{\lambda_N}(\widehat{\theta}^{\{K\}}) - \mathcal{F}_{\lambda_N}(\overline{\theta}^{\lambda_N}) \leq \frac{25(\lambda_K + \lambda_N)(\mathcal{K}_{\lambda_K}(\widehat{\theta}^{\{K\}}) + 3(\lambda_K - \lambda_N))s^*}{\widetilde{\rho}_-(s^* + \widetilde{s})},$$

which completes the proof due to $\lambda_K > \lambda_N$ for $K = 0, ..., N - 1$. $\square$

7.4. *Proof of Theorem 3.16.*

PROOF. For any $\theta^*$, we consider a partition of $\mathbb{R}^d$ as

$$\mathcal{S}_1 = \left\{j \;\Big|\; \theta_j^* \geq \frac{C_2 \sigma}{\sqrt{s_1^* + s_2^*}}\right\}, \text{ and } \mathcal{S}_0 = \left\{j \;\Big|\; \theta_j^* < \frac{C_2 \sigma}{\sqrt{s_1^* + s_2^*}}\right\}.$$

We consider the first scenario, where $\mathcal{S}_0 = \emptyset$. Then we establish the lower bound for estimating $\theta_{\mathcal{S}_1}^*$ only. Let $\widetilde{\theta}_{\mathcal{S}_1}$ denote any estimator of $\theta_{\mathcal{S}_1}^*$ based on $y \sim N(X_{*\mathcal{S}_1}\theta_{\mathcal{S}_1}^*, \sigma^2 I_n)$. This is essentially a low dimensional linear regression problem since $s_1^* < n$. By the minimax lower bound for standard linear regression model in Duchi (2015), we have

$$\inf_{\widetilde{\theta}_{\mathcal{S}_1}} \sup_{\theta \in \Theta(s_1^*, s_2^*, d)} \mathbb{E}\|\widetilde{\theta}_{\mathcal{S}_1} - \theta_{\mathcal{S}_1}^*\|_2 \geq C_6 \sigma \sqrt{\frac{s_1^*}{n}}$$

for a generic constant $C_6$. We then consider a second scenario, where $\mathcal{S}_1 = \emptyset$. Then we establish the lower bound for estimating $\theta_{\mathcal{S}_0}^*$ only. Let $\widetilde{\theta}_{\mathcal{S}_0}$ denote any estimator of $\theta_{\mathcal{S}_0}^*$ based on $y \sim N(X_{*\mathcal{S}_0}\theta_{\mathcal{S}_0}^*, \sigma^2 I_n)$. This is essentially a high dimensional sparse linear regression problem. By the lower bound for sparse linear regression model established in Raskutti et al. (2011), we have

$$\inf_{\widetilde{\theta}_{\mathcal{S}_0}} \sup_{\theta \in \Theta(s_1^*, s_2^*, d)} \mathbb{E}\|\widetilde{\theta}_{\mathcal{S}_0} - \theta_{\mathcal{S}_0}^*\|_2 \geq 2C_7 \sigma \sqrt{\frac{s_2^* \log(d - s_2^*)}{n}} \geq C_7 \sigma \sqrt{\frac{s_2^* \log d}{n}},$$

where $C_7$ is a generic constant and the last inequality comes from the fact $s_2^* \ll d$. Combining two scenarios, we have

$$\inf_{\widehat{\theta}} \sup_{\theta \in \Theta(s_1^*, s_2^*, d)} \mathbb{E}\|\widehat{\theta} - \theta^*\|_2$$

$$\geq \max\left\{ \inf_{\widetilde{\theta}_{\mathcal{S}_1}} \sup_{\theta \in \Theta(s_1^*, s_2^*, d)} \mathbb{E}\|\widetilde{\theta}_{\mathcal{S}_1} - \theta^*_{\mathcal{S}_1}\|_2, \inf_{\widetilde{\theta}_{\mathcal{S}_0}} \sup_{\theta \in \Theta(s_1^*, s_2^*, d)} \mathbb{E}\|\widetilde{\theta}_{\mathcal{S}_0} - \theta^*_{\mathcal{S}_0}\|_2 \right\}$$

$$\geq \frac{1}{2} \inf_{\widetilde{\theta}_{\mathcal{S}_1}} \sup_{\theta \in \Theta(s_1^*, s_2^*, d)} \mathbb{E}\|\widetilde{\theta}_{\mathcal{S}_1} - \theta^*_{\mathcal{S}_1}\|_2 + \frac{1}{2} \inf_{\widetilde{\theta}_{\mathcal{S}_0}} \sup_{\theta \in \Theta(s_1^*, s_2^*, d)} \mathbb{E}\|\widetilde{\theta}_{\mathcal{S}_0} - \theta^*_{\mathcal{S}_0}\|_2$$

$$\geq \frac{C_6}{2}\sigma\sqrt{\frac{s_1^*}{n}} + \frac{C_7}{2}\sigma\sqrt{\frac{s_2^* \log d}{n}} \geq C_4\left(\sigma\sqrt{\frac{s_1^*}{n}} + \sigma\sqrt{\frac{s_2^* \log d}{n}}\right),$$

where $C_4 = \min\{\frac{C_6}{2}, \frac{C_7}{2}\}$. $\square$

7.5. *Proof of Theorem 3.17.*

PROOF. For notational simplicity, we denote $\lambda_N$, $\widehat{\theta}^{\{N\}}$, and $\overline{\theta}^{\lambda_N}$ by $\lambda$, $\widehat{\theta}$, and $\overline{\theta}^\lambda$ respectively. Before we proceed, we introduce the following lemmas.

**Lemma 7.10.** Suppose $\varepsilon \sim N(0, \sigma^2 I_n)$ and $\|X_{*j}\|_2 = \sqrt{n}$ for $j = 1, ..., d$. Then we have

$$\mathbb{P}\left(\frac{1}{n}\|X^\top \varepsilon\|_\infty \geq 2\sigma\sqrt{\frac{\log d}{n}}\right) \leq 2d^{-2}.$$

**Lemma 7.11.** Suppose Assumptions 3.1 and 3.5, and the following event

$$\mathcal{E}_1 = \left\{\frac{1}{n}\|X^\top \varepsilon\|_\infty \geq 2\sigma\sqrt{\frac{\log d}{n}}\right\}$$

hold. We have

$$\frac{1}{n}X_{*\mathcal{S}}(y - X\widehat{\theta}^{\text{o}}) + \nabla_{\mathcal{S}}\mathcal{H}_\lambda(\widehat{\theta}^{\text{o}}) + \lambda\nabla\|\widehat{\theta}^{\text{o}}_{\mathcal{S}}\|_1 = 0.$$

**Lemma 7.12.** Suppose Assumptions 3.1, and 3.5, and the following event

$$\mathcal{E}_2 = \left\{\frac{1}{n}\|U^\top \varepsilon\|_\infty \geq 2\sigma\sqrt{\frac{\log d}{n}}\right\}$$

hold, where $U = X^\top(I_n - X_{*\mathcal{S}}(X_{*\mathcal{S}}^\top X_{*\mathcal{S}})^{-1}X_{*\mathcal{S}}^\top)$. There exists some $\widehat{\xi}^{\text{o}}_{\overline{\mathcal{S}}} \in \partial\|\widehat{\theta}^{\text{o}}_{\overline{\mathcal{S}}}\|_1$ such that

$$\frac{1}{n}X^\top_{*\overline{\mathcal{S}}}(y - X\widehat{\theta}^{\text{o}}) + \nabla_{\overline{\mathcal{S}}}\mathcal{H}_\lambda(\widehat{\theta}^{\text{o}}) + \lambda\widehat{\xi}^{\text{o}}_{\overline{\mathcal{S}}} = 0.$$

The proof of Lemma 7.10 is provided in Negahban et al. (2012), therefore is omitted. The proofs of Lemmas 7.11 and 7.12 are presented in Appendices E.4 and E.5 respectively. Lemmas 7.11 and 7.12 imply that $\widehat{\theta}^{\text{o}}$ satisfies the KKT condition of (1.1) over $\mathcal{S}$ and $\overline{\mathcal{S}}$ respectively. Note that the above results only depend on Conditions $\mathcal{E}_1$ and $\mathcal{E}_2$. Meanwhile, we also have

$$\|U_{*j}\|_2 = \|X_{*j}^\top (I_n - X_{*\mathcal{S}}(X_{*\mathcal{S}}^\top X_{*\mathcal{S}})^{-1} X_{*\mathcal{S}}^\top)\|_2$$
$$(7.10) \qquad \leq \|I_n - X_{*\mathcal{S}}(X_{*\mathcal{S}}^\top X_{*\mathcal{S}})^{-1} X_{*\mathcal{S}}^\top\|_2 \|X_{*j}\|_2 \leq \|X_{*j}\|_2 = \sqrt{n},$$

where the last inequality comes from $\|I_n - X_{*\mathcal{S}}(X_{*\mathcal{S}}^\top X_{*\mathcal{S}})^{-1} X_{*\mathcal{S}}^\top\|_2 \leq 1$. Thus, (7.10) implies that Lemma 7.10 is also applicable to $\mathcal{E}_2$. Moreover, since both $\widehat{\theta}^{\{N\}}$ and $\widehat{\theta}^{\text{o}}$ are sparse local optima, by Lemma C.1, we further have $\mathbb{P}(\widehat{\theta}^{\text{o}} = \overline{\theta}^\lambda) \geq 1 - 4d^{-2}$.

Moreover, since $\widehat{\theta}$ converges to $\overline{\theta}^\lambda$, given a sufficiently small $\delta_N$, we have

$$\|\nabla \widetilde{\mathcal{L}}_\lambda(\overline{\theta}^\lambda) - \nabla \widetilde{\mathcal{L}}_\lambda(\widehat{\theta})\|_\infty \leq \|\widetilde{\mathcal{L}}_\lambda(\overline{\theta}^\lambda) - \widetilde{\mathcal{L}}_\lambda(\widehat{\theta})\|_2 \leq \rho_+(s^*)\|\overline{\theta}^\lambda - \widehat{\theta}\|_2 \leq \omega \ll \frac{\lambda}{4}.$$

Since we have proved $\|\nabla_{\overline{\mathcal{S}}} \widetilde{\mathcal{L}}_\lambda(\overline{\theta}^\lambda)\|_\infty \leq \lambda/4$ in Lemma 7.12, we have

$$\|\widetilde{\mathcal{L}}_\lambda(\widehat{\theta})\|_\infty \leq \|\nabla_{\overline{\mathcal{S}}} \widetilde{\mathcal{L}}_\lambda(\overline{\theta}^\lambda)\|_\infty + \|\nabla \widetilde{\mathcal{L}}_\lambda(\overline{\theta}^\lambda) - \nabla \widetilde{\mathcal{L}}_\lambda(\widehat{\theta})\|_\infty \leq \frac{\lambda}{4} + \omega.$$

Since $\widehat{\theta}$ also satisfies the approximate KKT condition and $\delta \leq 1/8$, then we must have $\widehat{\theta}_{\overline{\mathcal{S}}} = 0$. Moreover, since we have also proved that there exists some constant $C_8$ such that $\min_{j \in \mathcal{S}} |\overline{\theta}_j^\lambda| \geq C_8 \sigma \sqrt{\log d/n}$ in Lemma 7.11, then for $\omega/\rho_-(s^*) \ll C_8 \sigma \sqrt{\log d/n}$, we have

$$\min_{j \in \mathcal{S}} |\widehat{\theta}_j| = \min_{j \in \mathcal{S}} |\overline{\theta}_j^\lambda| - \omega \geq C_8 \sigma \sqrt{\frac{\log d}{n}} > 0.$$

Combining with the fact $\widehat{\theta}_{\overline{\mathcal{S}}} = 0$, we have $\text{supp}(\widehat{\theta}) = \text{supp}(\overline{\theta}^\lambda) = \text{supp}(\theta^*)$. Meanwhile, since all signals are strong enough, then by Theorem 3.14, we also have $\|\widehat{\theta} - \theta^*\|_2 \leq C_3 \sigma \sqrt{\frac{s^*}{n}}$. □

**Acknowledgements.** We sincerely thank Zhaoran Wang, Xingguo Li, Jason Ge, and Qiang Sun for their helpful personal communications. We are grateful to the Editor, Associate Editor and referees for their insightful comments.

This work is supported by NSF IIS 1250985, NSF IIS 1407939, NSF DMS1454377 CAREER, NSF IIS 1546482 BIGDATA, NIH R01MH102339, NSF IIS 1408910, NSF IIS 1332109, NIH R01GM083084.

# SUPPLEMENTARY MATERIALS FOR "PATHWISE COORDINATE OPTIMIZATION FOR SPARSE LEARNING: ALGORITHM AND THEORY"

By Tuo Zhao, Han Liu, and Tong Zhang

The supplementary materials contain the supplementary proofs of the theoretical lemmas in the paper "Pathwise Coordinate Optimization for Nonconvex Sparse Learning: Algorithm and Theory".

## APPENDIX A: COMPUTATIONAL COST COMPARISON

We first show that the computational cost of each proximal gradient iteration is $\mathcal{O}(nd)$. At the $t$-th iteration, we calculate

$$\theta^{(t+1)} = \mathcal{S}_{\lambda/L}\left(\theta^{(t)} - \frac{1}{Ln}X^\top\left(y^{(t)} - X\theta^{(t)}\right)\right),$$

where $L$ is the step size parameter. Thus, the computational cost is $\mathcal{O}(ns + nd + d + d) = \mathcal{O}(nd)$, where $s = \|\theta^{(t)}\|_0 \leq d$.

We then show that the computational cost of each coordinate minimization iteration is only $\mathcal{O}(n)$. Suppose we maintain $\widetilde{y}^{(t)} = X_{*\backslash j}\theta^{(t)}_{\backslash j}$ for the $t$-th iteration. Then we calculate $\theta^{(t+1)}_j$ by

$$(A.1) \qquad \theta^{(t+1)}_j = \widetilde{\theta}^{(t)}_j \cdot \mathbb{1}_{\{|\widetilde{\theta}^{(t)}_j| \geq \gamma\lambda\}} + \frac{\mathcal{S}_\lambda(\widetilde{\theta}^{(t)}_j)}{1 - 1/\gamma} \cdot \mathbb{1}_{\{|\widetilde{\theta}^{(t)}_j| < \gamma\lambda\}},$$

where $\widetilde{\theta}^{(t)}_j = \frac{1}{n}X^\top_{*j}(y - \widetilde{y}^{(t)})$. Thus, the computational cost of (A.1) is $\mathcal{O}(n)$. Once we have $\widetilde{\theta}^{(t)}_j$, we obtain $\widetilde{y}^{(t+1)}$ for the $(t+1)$ iteration by

$$\widetilde{y}^{(t+1)} = \widetilde{y}^{(t)} + X_{*j}(\theta^{(t+1)}_j - \theta^{(t)}_j),$$

and the computational cost is also $\mathcal{O}(n)$. Thus the overall computational cost is $\mathcal{O}(n)$. For the proximal coordinate gradient algorithm, the coordinate gradient can be computed using a similar strategy, and therefore its overall computational cost is also $\mathcal{O}(n)$ for each iteration.

## APPENDIX B: THE MCP REGULARIZER

Throughout our analysis, we frequently use the following properties of the MCP regularizer.

**Lemma B.1.** For the MCP regularizer, $h(\cdot)$ and $h'(\cdot)$ satisfy:

(R.1) For any $a > b \geq 0$, we have
$$-\alpha(a-b) \leq h'_\lambda(a) - h'_\lambda(b) \leq 0,$$
where $\alpha = 1/\gamma \geq 0$;

(R.2) For some $\gamma > 0$ and $\forall\, a \geq 0$, we have $h'_\lambda(a) \in [-\lambda, 0]$ if $a \leq \lambda\gamma$, and $h'_\lambda(a) = -\lambda$ otherwise;

(R.3) $h_\lambda(\cdot)$ and $h'_\lambda(\cdot)$ pass through the origin, i.e., $h_\lambda(0) = 0$ and $h'_\lambda(0) = 0$;

(R.4) For $\forall\, a \geq 0$, we have $|h'_{\lambda_1}(a) - h'_{\lambda_2}(a)| \leq |\lambda_1 - \lambda_2|$.

The proof of Lemma B.1 is straightforward, and therefore omitted. Note that all above properties also hold for Lasso, i.e., $\gamma = \infty$ and $h_\lambda(\cdot) = 0$.

## APPENDIX C: LEMMAS FOR COMPUTATIONAL THEORY

### C.1. Proof of Lemma 3.4.

PROOF. Since $\mathcal{L}(\theta)$ is twice differentiable and $\|\theta - \theta'\|_0 \leq s$, by the mean value theorem, we have

$$(C.1) \quad \mathcal{L}(\theta') - \mathcal{L}(\theta) - (\theta' - \theta)^\top \nabla \mathcal{L}(\theta) = \frac{1}{2}(\theta' - \theta)^\top \nabla^2 \mathcal{L}(\widetilde{\theta})(\theta' - \theta),$$

where $\widetilde{\theta} = (1-\beta)\theta' + \beta\theta$ for some $\beta \in (0,1)$. By Definition 3.3, we have

$$(C.2) \quad \frac{\rho_-(s)}{2}\|\theta' - \theta\|_2^2 \leq \frac{1}{2}(\theta' - \theta)^\top \nabla^2 \mathcal{L}(\widetilde{\theta})(\theta' - \theta) \leq \frac{\rho_+(s)}{2}\|\theta' - \theta\|_2^2.$$

Combining (C.1) with (C.2), we have

$$(C.3) \quad \frac{\rho_-(s)}{2}\|\theta' - \theta\|_2^2 \leq \mathcal{L}(\theta') - \mathcal{L}(\theta) - (\theta' - \theta)^\top \nabla \mathcal{L}(\theta) \leq \frac{\rho_+(s)}{2}\|\theta' - \theta\|_2^2.$$

By (R.1) in Assumption B.1, we have

$$(C.4) \quad -\frac{\alpha}{2}\|\theta' - \theta\|_2^2 \leq \mathcal{H}_\lambda(\theta') - \mathcal{H}_\lambda(\theta) - (\theta' - \theta)^\top \nabla \mathcal{H}_\lambda(\theta) \leq 0.$$

Combining (C.3) with (C.4), we have

$$(C.5) \quad \frac{\rho_-(s) - \alpha}{2}\|\theta' - \theta\|_2^2 \leq \widetilde{\mathcal{L}}_\lambda(\theta') - \widetilde{\mathcal{L}}_\lambda(\theta) - (\theta' - \theta)^\top \nabla \widetilde{\mathcal{L}}_\lambda(\theta)$$
$$\leq \frac{\rho_+(s)}{2}\|\theta' - \theta\|_2^2.$$

By the convexity of $\|\theta\|_1$, we have

(C.6) $$\|\theta'\|_1 \geq \|\theta\|_1 + (\theta' - \theta)^\top \xi$$

for any $\xi \in \partial \|\theta\|_1$. Combining (C.6) with (C.5), we obtain

$$\mathcal{F}_\lambda(\theta') \geq \mathcal{F}_\lambda(\theta) + (\theta' - \theta)^\top (\nabla \widetilde{\mathcal{L}}_\lambda(\theta) + \lambda \xi) + \frac{\rho_-(s)}{2}\|\theta' - \theta\|_2^2.$$

□

### C.2. Proof of Lemma 7.1.

PROOF. By Lemma 7.8, we have

$$\mathcal{F}_\lambda(w^{(t+1,k-1)}) - \mathcal{F}_\lambda(w^{(t+1,k)}) \geq \frac{\nu_-(1)}{2}(w_k^{(t+1,k-1)} - w_k^{(t+1,k)})^2$$
$$= \frac{\nu_-(1)}{2}(\theta_k^{(t+1)} - \theta_k^{(t)})^2,$$

which further implies

$$\mathcal{F}_\lambda(\theta^{(t)}) - \mathcal{F}_\lambda(\theta^{(t+1)}) = \sum_{k=1}^{s}[\mathcal{F}_\lambda(w^{(t+1,k-1)}) - \mathcal{F}_\lambda(w^{(t+1,k)})]$$
$$\geq \frac{\nu_-(1)}{2}\|\theta^{(t)} - \theta^{(t+1)}\|_2^2.$$

□

### C.3. Proof of Lemma 7.2.

PROOF. We first analyze the gap for the proximal coordinate gradient descent. Let $\theta \in \mathbb{R}^d$ be a vector satisfying $\theta_{\overline{\mathcal{A}}} = 0$. By the restricted convexity of $\mathcal{F}_\lambda(\theta)$, we have

(C.7) $$\mathcal{F}_\lambda(\theta) \geq \mathcal{F}_\lambda(\theta^{(t+1)}) + (\nabla_{\mathcal{A}} \widetilde{\mathcal{L}}_\lambda(\theta^{(t+1)}) + \lambda \xi_{\mathcal{A}}^{(t+1)})^\top (\theta - \theta^{(t+1)})$$
$$+ \frac{\widetilde{\rho}_-(s)}{2}\|\theta - \theta^{(t+1)}\|_2^2,$$

where $\xi_{\mathcal{A}}^{(t+1)}$ satisfies the optimality condition of the proximal coordinate gradient descent,

(C.8) $$\nabla \mathcal{V}_{\lambda,k,L}(\theta_k^{(t+1)}; w^{(t+1,k-1)}) + \lambda \xi_k^{(t+1)} = 0 \text{ for any } k \in \mathcal{A}.$$

By setting $\theta_{\overline{\mathcal{A}}} = 0$ and minimizing both sides of (C.7) over $\theta_{\mathcal{A}}$, we obtain

$$\text{(C.9)} \quad \mathcal{F}_\lambda(\theta^{(t+1)}) - \mathcal{F}_\lambda(\bar{\theta}) \leq \frac{1}{2\widetilde{\rho}_-(s)} \|\nabla_{\mathcal{A}} \widetilde{\mathcal{L}}_\lambda(\theta^{(t+1)}) + \lambda \xi_{\mathcal{A}}^{(t+1)}\|_2^2$$

$$\overset{\text{(i)}}{=} \frac{1}{2\widetilde{\rho}_-(s)} \sum_{k=1}^{s} \|\nabla_k \widetilde{\mathcal{L}}_\lambda(\theta^{(t+1)}) - \nabla \mathcal{V}_{\lambda,k,L}(\theta_k^{(t+1)}; w^{(t+1,k-1)})\|_2^2$$

$$\overset{\text{(ii)}}{\leq} \frac{\rho_+^2(s)}{2\widetilde{\rho}_-(s)} \sum_{k=1}^{s} \|\theta^{(t+1)} - w^{(t+1,k-1)}\|_2^2 \leq \frac{s\rho_+^2(s)}{2\widetilde{\rho}_-(s)} \|\theta^{(t+1)} - \theta^{(t)}\|_2^2,$$

where (i) comes from (C.8), and (ii) comes from $\nabla \mathcal{V}_{\lambda,k,L}(\theta_k^{(t+1)}; w^{(t+1,k-1)}) = \nabla \widetilde{\mathcal{L}}_\lambda(w^{(t+1,k-1)})$ and the restricted smoothness of $\widetilde{\mathcal{L}}_\lambda(\theta)$.

For the exact coordinate minimization, we have $\nabla \mathcal{V}_{\lambda,k,L}(\theta_k^{(t+1)}; w^{(t+1,k-1)}) = \nabla \mathcal{Y}_{\lambda,k}(\theta_k^{(t+1)}; w^{(t+1,k-1)})$. Thus, (C.9) also holds.

$\square$

### C.4. Proof of Lemma 7.8.

PROOF. For the proximal coordinate gradient descent, we have

$$\text{(C.10)} \qquad \mathcal{F}_\lambda(\theta) = \mathcal{V}_{\lambda,k,L}(\theta_k; \theta) + \lambda|\theta_k| + \lambda\|\theta_{\setminus k}\|_1,$$

$$\text{(C.11)} \qquad \mathcal{F}_\lambda(w) \leq \mathcal{V}_{\lambda,k,L}(w_k; \theta) + \lambda|\theta_k'| + \lambda\|\theta_{\setminus k}\|_1.$$

Since $\mathcal{V}_{\lambda,k,L}(\theta_k; \theta)$ is strongly convex in $\theta_k$, we have

$$\text{(C.12)} \quad \mathcal{V}_{\lambda,k,L}(\theta_k; \theta) - \mathcal{V}_{\lambda,k,L}(w_k; \theta)$$
$$\geq (\theta_k - w_k) \nabla \mathcal{V}_{\lambda,k,L}(w_k; \theta) + \frac{L}{2}(w_k - \theta_k)^2.$$

By the convexity of the absolute value function, we have

$$\text{(C.13)} \qquad |\theta_k| - |w_k| \geq (\theta_k - w_k)\xi_k,$$

where $\xi_k \in \partial |w_k|$ satisfies the optimality condition of the proximal coordinate gradient descent,

$$\text{(C.14)} \qquad \nabla \mathcal{V}_{\lambda,k,L}(w_k; \theta) + \lambda \xi_k = 0.$$

Subtracting (C.10) by (C.11), we have

$$\mathcal{F}_\lambda(\theta) - \mathcal{F}_\lambda(w) \geq \mathcal{V}_{\lambda,k,L}(\theta_k; \theta) - \mathcal{V}_{\lambda,k,L}(w_k; \theta) + \lambda|\theta_k| - \lambda|w_k|$$
$$\overset{\text{(i)}}{\geq} (\theta_k - w_k)(\nabla \mathcal{V}_{\lambda,k,L}(w_k; \theta) + \lambda \xi_k) + \frac{L}{2}(w_k - \theta_k)^2 \overset{\text{(ii)}}{\geq} \frac{L}{2}(w_k - \theta_k)^2.$$

where (i) comes from (C.12) and (C.13), and (ii) comes from (C.14).

For the exact coordinate minimization, we only need to slightly trim the above analysis. Specifically, we replace $\mathcal{V}_{\lambda,k,L}(w_k;\theta)$ with

$$\mathcal{Y}_{\lambda,k}(w_k;\theta) = \widetilde{\mathcal{L}}_\lambda(w_k, \theta_{\setminus k}).$$

Since $\widetilde{\mathcal{L}}_\lambda(\theta)$ is restrictedly convex, we have

$$\mathcal{Y}_{\lambda,k}(\theta_k;\theta) - \mathcal{Y}_{\lambda,k}(w_k;\theta) \geq (\theta_k - w_k)\nabla \mathcal{Y}_{\lambda,k}(\theta'_k;\theta) + \frac{\widetilde{\rho}_-(1)}{2}(w_k - \theta_k)^2.$$

Eventually, we obtain

$$\mathcal{F}_\lambda(w) - \mathcal{F}_\lambda(\theta) \geq \frac{\widetilde{\rho}_-(1)}{2}(w_k - \theta_k)^2.$$

We then proceed to analyze the descent for the proximal coordinate gradient descent when $\theta_k = 0$ and $|\nabla_k \widetilde{\mathcal{L}}_\lambda(\theta)| \geq (1+\delta)\lambda$. Then we have

$$|w_k| = |\mathcal{S}_{\lambda/L}(-\nabla_k \widetilde{\mathcal{L}}_\lambda(\theta)/L)| \geq \frac{\delta\lambda}{L},$$

where the last inequality comes from the definition of the soft thresholding function. Thus, we obtain

$$\mathcal{F}_\lambda(\theta) - \mathcal{F}_\lambda(w) \geq \frac{L}{2}w_k^2 \geq \frac{\delta^2\lambda^2}{2L}.$$

For the exact coordinate minimization, we construct an auxiliary solution $w'$ by a proximal coordinate gradient descent iteration using $L = \rho_+(1)$. Since $w$ is obtained by the exact minimization, we have

$$\mathcal{F}_\lambda(\theta) - \mathcal{F}_\lambda(w) \geq \mathcal{F}_\lambda(\theta) - \mathcal{F}_\lambda(w') \geq \frac{\delta^2\lambda^2}{2\rho_+(1)}.$$

$\square$

### C.5. Proof of Lemma 7.3.

PROOF. Before we proceed, we first introduce the following lemma.

**Lemma C.1.** Suppose Assumption (3.5) holds. If $\overline{\theta}^\lambda$ satisfies

$$\|\overline{\theta}^\lambda_{\overline{\mathcal{S}}}\|_0 \leq \widetilde{s} \quad \text{and} \quad \mathcal{K}_\lambda(\overline{\theta}^\lambda) = 0,$$

then $\overline{\theta}^\lambda$ is a unique sparse local optimum to (1.1).

The proof of Lemma is provided in Appendix C.13. We then proceed with the proof. We consider a sequence of auxiliary solutions obtained by the proximal gradient algorithm. The details for generating such a sequence are provided in Wang et al. (2014). By Theorem 5.1 in Wang et al. (2014), we know that such a sequence of solutions converges to a sparse local optimum $\overline{\theta}^\lambda$. By Lemma C.1, we know that the sparse local optimum is unique. $\square$

### C.6. Proof of Lemma 7.4.

Proof. Before we proceed, we first introduce the following lemma.

**Lemma C.2.** Suppose Assumptions 3.1, 3.5, and 3.7 hold. For any $\lambda \geq \lambda_N$, if $\theta$ satisfies

(C.15) $\qquad \|\theta_{\overline{S}}\|_0 \leq s \quad \text{and} \quad \mathcal{F}_\lambda(\theta) \leq \mathcal{F}_\lambda(\theta^*) + \frac{4\lambda^2 s^*}{\widetilde{\rho}_-(s^* + s)},$

where $s \leq 2\widetilde{s}$, then we have

$$\|\theta - \theta^*\|_2 \leq \frac{9\lambda\sqrt{s^*}}{\widetilde{\rho}_-(s^* + s)} \quad \text{and} \quad \|\theta - \theta^*\|_1 \leq \frac{25\lambda s^*}{\widetilde{\rho}_-(s^* + s)}.$$

The proof of Lemma C.2 is provided in Appendix C.7. Lemma C.2 characterizes the estimation errors of any sufficiently sparse solution with a sufficiently small objective value.

When the inner loop terminates, we have the output solution as $\widehat{\theta} = \theta^{(t+1)}$. Since both the exact coordinate minimization and proximal coordinate gradient descent iterations always decrease the objective value, we have

(C.16) $\qquad \mathcal{F}_\lambda(\theta^{(t+1)}) \leq \mathcal{F}_\lambda(\theta^*) + \frac{4\lambda^2 s^*}{\widetilde{\rho}_-(s^* + 2\widetilde{s})}.$

By (C.9) in Appendix C.3, we have shown

(C.17) $\ \|\nabla_\mathcal{A}\widetilde{\mathcal{L}}_\lambda(\theta^{(t+1)}) + \lambda\xi_\mathcal{A}^{(t+1)}\|_2^2 \leq (s^* + 2\widetilde{s})\rho_+^2(s^* + 2\widetilde{s})\|\theta^{(t+1)} - \theta^{(t)}\|_2^2.$

Since Assumption 3.7 holds and $\widetilde{\rho}_-(1) \leq \nu_+(1)$, we have

(C.18) $\qquad \|\theta^{(t+1)} - \theta^{(t)}\|_2^2 \leq \tau^2\lambda^2 \leq \frac{\delta^2\lambda^2}{(s^* + 2\widetilde{s})\rho_+^2(s^* + 2\widetilde{s})}.$

Combining (C.17) with (C.18), we have $\theta^{(t+1)}$ satisfying the approximate KKT condition over the active set,

$\min_{\xi_\mathcal{A} \in \partial\|\theta_\mathcal{A}^{(t+1)}\|_1} \|\nabla_\mathcal{A}\widetilde{\mathcal{L}}_\lambda(\theta^{(t+1)}) + \lambda\xi_\mathcal{A}\|_\infty \leq \|\nabla_\mathcal{A}\widetilde{\mathcal{L}}_\lambda(\theta^{(t+1)}) + \lambda\xi_\mathcal{A}^{(t+1)}\|_2 \leq \delta\lambda.$

We now proceed to characterize the sparsity of $\widehat{\theta} = \theta^{(t+1)}$ by exploiting the above approximate KKT condition. By Assumption 3.1, we have $\lambda \geq 4\|\nabla\widetilde{\mathcal{L}}_\lambda(\theta^*)\|_\infty$, which implies

$$\text{(C.19)} \quad \left|\{j \mid |\nabla_j\widetilde{\mathcal{L}}_\lambda(\theta^*)| \geq \lambda/4, \ j \in \overline{\mathcal{S}} \cap \mathcal{A}\}\right| = 0.$$

We then consider an arbitrary set $\mathcal{S}'$ such that

$$\mathcal{S}' = \{j \mid |\nabla_j\widetilde{\mathcal{L}}_\lambda(\widehat{\theta}) - \nabla_j\widetilde{\mathcal{L}}_\lambda(\theta^*)| \geq \lambda/2, \ j \in \overline{\mathcal{S}} \cap \mathcal{A}\}.$$

Let $s' = |\mathcal{S}'|$. There exists a $v \in \mathbb{R}^d$ such that

$$\text{(C.20)} \quad \|v\|_\infty = 1, \quad \|v\|_0 \leq s', \quad \text{and} \quad s'\lambda/2 \leq v^\top(\nabla\widetilde{\mathcal{L}}_\lambda(\widehat{\theta}) - \nabla\widetilde{\mathcal{L}}_\lambda(\theta^*)).$$

By Cauchy-Schwarz inequality, (C.20) implies

$$\text{(C.21)} \quad \frac{s'\lambda}{2} \leq \|v\|_2 \|\nabla\widetilde{\mathcal{L}}_\lambda(\widehat{\theta}) - \nabla\widetilde{\mathcal{L}}_\lambda(\theta^*)\|_2 \leq \sqrt{s'}\|\nabla\widetilde{\mathcal{L}}_\lambda(\widehat{\theta}) - \nabla\widetilde{\mathcal{L}}_\lambda(\theta^*)\|_2$$
$$\overset{\text{(i)}}{\leq} \rho_+(s^* + 2\widetilde{s})\sqrt{s'}\|\widehat{\theta} - \theta^*\|_2 \overset{\text{(ii)}}{\leq} \rho_+(s^* + 2\widetilde{s})\sqrt{s'}\frac{9\lambda\sqrt{s^*}}{\widetilde{\rho}_-(s^* + 2\widetilde{s})},$$

where (i) comes from the restricted smoothness of $\widetilde{\mathcal{L}}_\lambda(\theta)$, and (ii) comes from (C.16) and Lemma C.2. (C.21) further implies

$$\text{(C.22)} \quad \sqrt{s'} \leq \frac{18\rho_+(s^* + 2\widetilde{s})\sqrt{s^*}}{\widetilde{\rho}_-(s^* + 2\widetilde{s})}.$$

Since $\mathcal{S}'$ is arbitrary defined, by simple manipulation, (C.22) implies

$$\text{(C.23)} \quad \left|\{j \mid |\nabla_j\widetilde{\mathcal{L}}_\lambda(\widehat{\theta}) - \nabla_j\widetilde{\mathcal{L}}_\lambda(\theta^*)| \geq \lambda/2, \ j \in \overline{\mathcal{S}} \cap \mathcal{A}\}\right| \leq 364\kappa^2 s^*.$$

Combining (C.19) with (C.23), we have

$$\text{(C.24)} \quad \left|\{j \mid |\nabla_j\widetilde{\mathcal{L}}_\lambda(\widehat{\theta})| \geq 3\lambda/4, \ j \in \overline{\mathcal{S}} \cap \mathcal{A}\}\right|$$
$$\leq \left|\{j \mid |\nabla_j\widetilde{\mathcal{L}}_\lambda(\theta^*)| \geq \lambda/4, \ j \in \overline{\mathcal{S}} \cap \mathcal{A}\}\right|$$
$$+ \left|\{j \mid |\nabla_j\widetilde{\mathcal{L}}_\lambda(\widehat{\theta}) - \nabla_j\widetilde{\mathcal{L}}_\lambda(\theta^*)| \geq \lambda/2, \ j \in \overline{\mathcal{S}} \cap \mathcal{A}\}\right| \leq 364\kappa^2 s^* < \widetilde{s},$$

where the last inequality comes from Assumption 3.5. Since we require $\delta \leq 1/8$ in Assumption 3.7, (C.24) implies that for any $u \in \mathbb{R}^d$ satisfying $\|u\|_\infty \leq 1$, we have

$$\left|\{j \mid |\nabla_j\widetilde{\mathcal{L}}_\lambda(\widehat{\theta}) + \delta\lambda u_j| \geq 7\lambda/8, \ j \in \overline{\mathcal{S}} \cap \mathcal{A}\}\right| \leq \widetilde{s}.$$

Then for any $j \in \overline{\mathcal{S}} \cap \mathcal{A}$ satisfying $|\nabla_j\widetilde{\mathcal{L}}_\lambda(\widehat{\theta}) + \delta\lambda u_j| \leq 7\lambda/8$, there exists a $\xi_j$ such that

$$|\xi_j| \leq 1 \quad \text{and} \quad \nabla_j\widetilde{\mathcal{L}}_\lambda(\widehat{\theta}) + \delta\lambda u_j + \lambda\xi_j = 0,$$

which further implies $\widehat{\theta}_j = 0$. Thus, we must have $\|\widehat{\theta}_{\overline{\mathcal{S}}}\|_0 \leq \widetilde{s}$. $\square$

### C.7. Proof of Lemma C.2.

PROOF. For notational simplicity, we define $\Delta = \theta - \theta^*$. We first rewrite (C.15) as

$$(C.25) \qquad \lambda\|\theta^*\|_1 - \lambda\|\theta\|_1 + \frac{4\lambda^2 s^*}{\widetilde{\rho}_-(s^* + s)} \geq \widetilde{\mathcal{L}}_\lambda(\theta) - \widetilde{\mathcal{L}}_\lambda(\theta^*).$$

By the restricted convexity of $\widetilde{\mathcal{L}}_\lambda(\theta)$, we have

$$(C.26) \quad \widetilde{\mathcal{L}}_\lambda(\theta) - \widetilde{\mathcal{L}}_\lambda(\theta^*) - \frac{\widetilde{\rho}_-(s^* + s)}{2}\|\Delta\|_2^2$$
$$\overset{(i)}{\geq} \Delta_{\mathcal{S}}^\top[\nabla_{\mathcal{S}}\mathcal{L}(\theta^*) + \nabla_{\mathcal{S}}\mathcal{H}_\lambda(\theta^*)] + \Delta_{\overline{\mathcal{S}}}^\top \nabla_{\overline{\mathcal{S}}}\mathcal{L}(\theta^*)$$
$$\overset{(ii)}{\geq} -\|\Delta_{\mathcal{S}}\|_1\|\nabla\mathcal{L}(\theta^*)\|_\infty - \|\Delta_{\overline{\mathcal{S}}}\|_1\|\nabla\mathcal{L}(\theta^*)\|_\infty - \|\Delta_{\mathcal{S}}\|_1\|\nabla_{\mathcal{S}}\mathcal{H}_\lambda(\theta^*)\|_\infty,$$

where (i) comes from $\nabla_{\overline{\mathcal{S}}}\mathcal{H}_\lambda(\theta^*) = 0$ by (R.3) of Lemma B.1, and (ii) comes from Hölder's inequality. Assumption 3.1 and (R.2) of Lemma B.1 imply

$$(C.27) \qquad \|\nabla\mathcal{L}(\theta^*)\|_\infty \leq \frac{\lambda}{4} \quad \text{and} \quad \|\nabla_{\mathcal{S}}\mathcal{H}_\lambda(\theta^*)\|_\infty \leq \lambda.$$

Combining (C.26) with (C.27), we obtain

$$(C.28) \qquad \widetilde{\mathcal{L}}_\lambda(\theta) - \widetilde{\mathcal{L}}_\lambda(\theta^*) \geq -\frac{5\lambda}{4}\|\Delta_{\mathcal{S}}\|_1 - \frac{\lambda}{4}\|\Delta_{\overline{\mathcal{S}}}\|_1 + \frac{\widetilde{\rho}_-(s^* + s)}{2}\|\Delta\|_2^2.$$

Plugging (C.28) and

$$\|\theta^*\|_1 - \|\theta\|_1 = \|\theta^*_{\mathcal{S}}\|_1 - (\|\theta_{\mathcal{S}}\|_1 + \|\Delta_{\overline{\mathcal{S}}}\|_1) \leq \|\Delta_{\mathcal{S}}\|_1 - \|\Delta_{\overline{\mathcal{S}}}\|_1$$

into (C.25), we obtain

$$(C.29) \qquad \frac{9\lambda}{4}\|\Delta_{\mathcal{S}}\|_1 + \frac{4\lambda^2 s^*}{\widetilde{\rho}_-(s^* + s)} \geq \frac{3\lambda}{4}\|\Delta_{\overline{\mathcal{S}}}\|_1 + \frac{\widetilde{\rho}_-(s^* + s)}{2}\|\Delta\|_2^2.$$

We consider the first case: $\widetilde{\rho}_-(s^* + s)\|\Delta\|_1 > 16\lambda s^*$. Then we have

$$(C.30) \qquad \frac{5\lambda}{2}\|\Delta_{\mathcal{S}}\|_1 \geq \frac{\lambda}{2}\|\Delta_{\overline{\mathcal{S}}}\|_1 + \frac{\widetilde{\rho}_-(s^* + s)}{2}\|\Delta\|_2^2.$$

By simple manipulation, (C.30) implies

$$(C.31) \qquad \frac{\widetilde{\rho}_-(s^* + s)}{2}\|\Delta\|_2^2 \leq \frac{5\lambda}{2}\|\Delta_{\mathcal{S}}\|_1 \leq \frac{5\lambda}{2}\sqrt{s^*}\|\Delta_{\mathcal{S}}\|_2 \leq \frac{5\lambda}{2}\sqrt{s^*}\|\Delta\|_2,$$

where the second inequality comes from the fact that $\Delta_{\mathcal{S}}$ only contains $s^*$ entries. By simple manipulation, (C.31) further implies

$$\text{(C.32)} \qquad \|\Delta\|_2 \leq \frac{5\lambda\sqrt{s^*}}{\widetilde{\rho}_-(s^* + s)}.$$

Meanwhile, (C.30) also implies

$$\text{(C.33)} \qquad \|\Delta_{\overline{\mathcal{S}}}\|_1 \leq 5\|\Delta_{\mathcal{S}}\|_1.$$

Combining (C.32) with (C.33), we obtain

$$\text{(C.34)} \qquad \|\Delta\|_1 \leq 5\|\Delta_{\mathcal{S}}\|_1 \leq 5\sqrt{s^*}\|\Delta_{\mathcal{S}}\|_2 \leq 5\sqrt{s^*}\|\Delta\|_2 \leq \frac{25\lambda s^*}{\widetilde{\rho}_-(s^* + s)}.$$

We consider the second case: $\widetilde{\rho}_-(s^* + s)\|\Delta\|_1 \leq 16\lambda s^*$. Then (C.29) implies

$$\|\Delta\|_2 \leq \frac{9\lambda\sqrt{s^*}}{\widetilde{\rho}_-(s^* + s)}.$$

Combining two cases, we obtain

$$\|\Delta\|_2 \leq \frac{9\lambda\sqrt{s^*}}{\widetilde{\rho}_-(s^* + s)} \quad \text{and} \quad \|\Delta\|_1 \leq \frac{25\lambda s^*}{\widetilde{\rho}_-(s^* + s)}.$$

$\square$

### C.8. Proof of Lemma 7.5.

PROOF. By Assumption 3.1, we have $\lambda \geq 4\|\nabla\widetilde{\mathcal{L}}_\lambda(\theta^*)\|_\infty$, which implies

$$\text{(C.35)} \qquad \left|\{j \mid |\nabla_j\widetilde{\mathcal{L}}_\lambda(\theta^*)| \geq \lambda/4,\ j \in \overline{\mathcal{S}} \cap \mathcal{A}\}\right| = 0.$$

We then consider an arbitrary set $\mathcal{S}'$ such that

$$\mathcal{S}' = \{j \mid |\nabla_j\widetilde{\mathcal{L}}_\lambda(\theta^{[0]}) - \nabla_j\widetilde{\mathcal{L}}_\lambda(\theta^*)| \geq \lambda/2,\ j \in \overline{\mathcal{S}}\}.$$

Let $s' = |\mathcal{S}'|$. Then there exists a $v \in \mathbb{R}^d$ such that

$$\text{(C.36)} \quad \|v\|_\infty = 1, \quad \|v\|_0 \leq s', \quad \text{and} \quad s'\lambda/2 \leq v^\top(\nabla\widetilde{\mathcal{L}}_\lambda(\theta^{[0]}) - \nabla\widetilde{\mathcal{L}}_\lambda(\theta^*)).$$

By Cauchy-Schwarz inequality, (C.36) implies

$$\text{(C.37)} \quad \frac{s'\lambda}{2} \leq \|v\|_2\|\nabla\widetilde{\mathcal{L}}_\lambda(\theta^{[0]}) - \nabla\widetilde{\mathcal{L}}_\lambda(\theta^*)\|_2 \leq \sqrt{s'}\|\nabla\widetilde{\mathcal{L}}_\lambda(\theta^{[0]}) - \nabla\widetilde{\mathcal{L}}_\lambda(\theta^*)\|_2$$

$$\overset{\text{(i)}}{\leq} \rho_+(s^* + 2\widetilde{s})\sqrt{s'}\|\theta^{[0]} - \theta^*\|_2 \overset{\text{(ii)}}{\leq} \rho_+(s^* + 2\widetilde{s})\sqrt{s'}\frac{9\lambda\sqrt{s^*}}{\widetilde{\rho}_-(s^* + 2\widetilde{s})},$$

where (i) comes from the restricted smoothness of $\widetilde{\mathcal{L}}_\lambda(\theta)$, and (ii) comes from Lemma C.2. By simple manipulation, (C.37) is rewritten as

$$\text{(C.38)} \qquad \sqrt{s'} \leq \frac{18\rho_+(s^* + 2\widetilde{s})\sqrt{s^*}}{\widetilde{\rho}_-(s^* + 2\widetilde{s})}.$$

Since $\mathcal{S}'$ is arbitrary defined, by simple manipulation, (C.22) implies

$$\text{(C.39)} \qquad \left|\{j \mid |\nabla_j \widetilde{\mathcal{L}}_\lambda(\theta^{[0]}) - \nabla_j \widetilde{\mathcal{L}}_\lambda(\theta^*)| \geq \lambda/2, \ j \in \overline{\mathcal{S}} \cap \mathcal{A}\}\right| \leq 364\kappa^2 s^*.$$

Combining (C.35) with (C.39), we have

$$\text{(C.40)} \quad \left|\{j \mid |\nabla_j \widetilde{\mathcal{L}}_\lambda(\theta^{[0]})| \geq 3\lambda/4, \ j \in \overline{\mathcal{S}} \cap \mathcal{A}\}\right|$$
$$\leq \left|\{j \mid |\nabla_j \widetilde{\mathcal{L}}_\lambda(\theta^*)| \geq \lambda/4, \ j \in \overline{\mathcal{S}} \cap \mathcal{A}\}\right|$$
$$+ \left|\{j \mid |\nabla_j \widetilde{\mathcal{L}}_\lambda(\theta^{[0]}) - \nabla_j \widetilde{\mathcal{L}}_\lambda(\theta^*)| \geq \lambda/2, \ j \in \overline{\mathcal{S}} \cap \mathcal{A}\}\right| \leq 364\kappa^2 s^* < \widetilde{s},$$

where the last inequality comes from Assumption 3.5. Since Assumption 3.7 requires $\varphi \leq 1/8$, we have $(1-\varphi)\lambda > 3\lambda/4$. Thus, (C.40) implies that the strong rule selects at most $\widetilde{s}$ irrelevant coordinates. □

### C.9. Proof of Lemma 7.6.

PROOF. Before we proceed, we first introduce the following lemmas.

**Lemma C.3.** Suppose Assumptions 3.1, 3.5, and 3.7 hold. For any $\lambda \geq \lambda_N$, if $\theta$ satisfies

$$\text{(C.41)} \qquad \|\theta_{\overline{\mathcal{S}}}\|_0 \leq \widetilde{s} \quad \text{and} \quad \mathcal{F}_\lambda(\theta) \leq \mathcal{F}_\lambda(\theta^*) + \frac{4\lambda^2 s^*}{\widetilde{\rho}_-(s^* + \widetilde{s})},$$

then we have $\|[\mathcal{T}_{\lambda,L}(\theta)]_{\overline{\mathcal{S}}}\|_0 \leq \widetilde{s}$.

The proof of Lemma C.3 is provided in Appendix C.10. Since $\theta^{[m+0.5]}$ satisfies (C.41) for all $m = 0, 1, 2, ...$, by Lemma C.3, we have $\|w_{\overline{\mathcal{S}}}^{[m+0.5]}\|_0 \leq \widetilde{s}$ for all $m = 0, 1, 2, ...$.

**Lemma C.4.** Suppose Assumptions 3.1, 3.5, and 3.7 hold. For every active set updating iteration, if we select a coordinate as

$$k_m = \operatorname{argmax}_{k \in \overline{\mathcal{A}}_m} |\nabla_k \widetilde{\mathcal{L}}_\lambda(\theta^{[m+0.5]})|,$$

then we have

$$k_m = \underset{k}{\operatorname{argmin}} \, \mathcal{Q}_{\lambda,k,L}(\mathcal{T}_{\lambda,k,L}(\theta^{[m+0.5]}); \theta^{[m+0.5]}).$$

The proof of Lemma C.4 is provided in Appendix C.11. Lemma C.4 guarantees that our selected coordinate $k_m$ leads to a sufficient descent in the objective value. Thus, we have

$$\text{(C.42)} \quad \mathcal{F}_\lambda(\theta^{[m+0.5]}) - \mathcal{F}_\lambda(\theta^{[m+1]})$$
$$\geq \mathcal{F}_\lambda(\theta^{[m+0.5]}) - \mathcal{Q}_{\lambda,k_m,L}(\theta_{k_m}^{[m+1]}; \theta^{[m+0.5]})$$
$$\geq \mathcal{F}_\lambda(\theta^{[m+0.5]}) - \frac{1}{|\mathcal{B}_m|} \sum_{k \in \mathcal{B}_m} \mathcal{Q}_{\lambda,k,L}(w_k^{[m+0.5]}; \theta^{[m+0.5]}),$$

where $\mathcal{B}_m = \{k \mid w_k^{[m+1]} \neq 0 \text{ or } \theta_k^{[m+0.5]} \neq 0\}$ and $|\mathcal{B}_m| \leq s^* + 2\widetilde{s}$. By rearranging (C.42), we obtain

$$\mathcal{F}_\lambda(\theta^{[m+0.5]}) - \mathcal{F}_\lambda(\theta^{[m+1]}) \geq \frac{1}{s^* + 2\widetilde{s}} \left[ \mathcal{F}_\lambda(\theta^{[m+0.5]}) - \mathcal{J}_{\lambda,L}(w^{[m+1]}; \theta^{[m+0.5]}) \right].$$

□

### C.10. Proof of Lemma C.3.

PROOF. We define an auxiliary solution

$$\widetilde{\theta} = \theta - \frac{1}{L} \nabla \widetilde{\mathcal{L}}_\lambda(\theta) = \theta - \frac{1}{L} \nabla \widetilde{\mathcal{L}}_\lambda(\theta^*) + \frac{1}{L}(\nabla \widetilde{\mathcal{L}}_\lambda(\theta) - \nabla \widetilde{\mathcal{L}}_\lambda(\theta^*)).$$

For notational simplicity, we denote $\Delta = \theta - \theta^*$. We first consider

$$\text{(C.43)} \quad \left|\{j \in \overline{\mathcal{S}} \mid |\theta_j| \geq L^{-1}\lambda/4\}\right| \leq \left|\{j \in \overline{\mathcal{S}} \mid |\Delta_j| \geq L^{-1}\lambda/4\}\right|$$
$$\leq \frac{4L}{\lambda} \|\Delta_{\overline{\mathcal{S}}}\|_1 \leq \frac{4L}{\lambda} \|\Delta\|_1 \leq \frac{100Ls^*}{\widetilde{\rho}_-(s^* + \widetilde{s})},$$

where the last inequality comes from Lemma C.2. By Assumption 3.1, we have $\|\nabla \widetilde{\mathcal{L}}_\lambda(\theta^*)\|_{\infty,2} \leq \lambda/4$, which implies

$$\text{(C.44)} \quad \left|\{j \in \overline{\mathcal{S}} \mid |\nabla_j \widetilde{\mathcal{L}}_\lambda(\theta^*)| \geq \lambda/4\}\right| = 0.$$

Recall in Appendix C.6, we have shown that

$$\text{(C.45)} \quad \left|\{j \mid |\nabla_j \widetilde{\mathcal{L}}_\lambda(\theta)| \geq \frac{\lambda}{2}, \ j \in \overline{\mathcal{S}} \cap \mathcal{A}\}\right| \leq 364\kappa^2 s^*.$$

Combining (C.43) and (C.44) with (C.45), we have

$$\text{(C.46)} \quad \left|\{j \in \overline{\mathcal{S}} \mid |\widetilde{\theta}_j| \geq L^{-1}\lambda\}\right| \leq \left|\{j \in \overline{\mathcal{S}} \mid |\theta_j| \geq L^{-1}\lambda/4\}\right|$$
$$+ \left|\{j \in \overline{\mathcal{S}} \mid |\nabla_j \widetilde{\mathcal{L}}_\lambda(\theta^*)| \geq \lambda/4\}\right| + \left|\{j \mid |\nabla_j \widetilde{\mathcal{L}}_\lambda(\theta)| \geq \lambda/2, \ j \in \overline{\mathcal{S}} \cap \mathcal{A}\}\right|$$
$$\leq \left(364\kappa^2 + \frac{100Ls^*}{\widetilde{\rho}_-(s^* + \widetilde{s})}\right) s^* \leq \widetilde{s},$$

where the last inequality comes from $L \leq \rho_+(s^* + 2\widetilde{s})$ and Assumption 3.5. By definition of the soft thresholding operator, we have $[\mathcal{T}_{\lambda,L}(\theta)]_j = \mathcal{S}_{\lambda/L}(\widetilde{\theta}_j)$. Thus, (C.46) further implies $\|[\mathcal{T}_{\lambda,L}(\theta)]_{\overline{\mathcal{S}}}\|_0 \leq \widetilde{s}$. $\square$

### C.11. Proof of Lemma C.4.

PROOF. Suppose there exists a coordinate $k$ such that

(C.47) $\quad \theta_k^{[m+0.5]} = 0 \quad \text{and} \quad |\nabla_k \widetilde{\mathcal{L}}_\lambda(\theta^{[m+0.5]})| \geq (1+\delta)\lambda.$

We conduct a proximal coordinate gradient descent iteration over the coordinate $k$, and obtain an auxiliary solution $w_k^{[m+1]}$. Since $w_k^{[m+1]}$ is obtained by the proximal coordinate gradient descent over the coordinate $k$, we have

(C.48) $\quad w_k^{[m+1]} = \underset{w_k}{\operatorname{argmin}}\, \mathcal{Q}_{\lambda,k,L}(w_k; \theta^{[m+0.5]}).$

We then derive an upper bound for $\mathcal{Q}_{\lambda,k,L}(w_k^{[m+1]}; \theta^{[m+0.5]})$. We consider

(C.49) $\quad \mathcal{Q}_{\lambda,k,L}(w_k^{[m+1]}; \theta^{[m+0.5]}) = \lambda|w_k^{[m+1]}| + \lambda\|\theta_{\setminus k}^{[m+0.5]}\|_1 + \widetilde{\mathcal{L}}_\lambda(\theta^{[m+0.5]})$
$\quad + (w_k^{[m+1]} - \theta_k^{[m+0.5]})\nabla_k \widetilde{\mathcal{L}}_\lambda(\theta^{[m+0.5]}) + \frac{L}{2}(w_k^{[m+1]} - \theta_k^{[m+0.5]})^2.$

By the convexity of the absolute value function, we have

(C.50) $\quad |\theta_k^{[m+0.5]}| \geq |w_k^{[m+1]}| + (\theta_k^{[m+0.5]} - w_k^{[m+1]})\xi_k,$

where $\xi_k \in \partial|w_k^{[m+1]}|$ satisfies the optimality condition of (C.48), i.e.,

(C.51) $\quad w_k^{[m+1]} - \theta_k^{[m+0.5]} + \frac{1}{L}\nabla_k \widetilde{\mathcal{L}}_\lambda(\theta^{[m+0.5]}) + \frac{\lambda}{L}\xi_k = 0$

for some $\xi_k \in \partial|w_k^{[m+1]}|$. Combining (C.50) with (C.49), we have

(C.52) $\quad \mathcal{Q}_{\lambda,k,L}(w_k^{[m+1]}; \theta^{[m+0.5]}) - \mathcal{F}_\lambda(\theta^{[m+0.5]})$
$\quad \leq (w_k^{[m+1]} - \theta_k^{[m+0.5]})(\nabla_k \widetilde{\mathcal{L}}_\lambda(\theta^{[m+0.5]}) + \lambda\xi_k) + \frac{L}{2}(w_k^{[m+1]} - \theta_k^{[m+0.5]})^2$
$\quad \stackrel{(i)}{=} -\frac{L}{2}(w_k^{[m+1]} - \theta_k^{[m+0.5]})^2 \stackrel{(ii)}{\leq} -\frac{\delta^2\lambda^2}{2L},$

where (i) comes from (C.51) and (ii) comes from Lemma 7.8 and (C.47).

Assume that there exists another coordinate $j$ with $\theta_j^{[m+0.5]} = 0$ such that

(C.53) $\quad |\nabla_k \widetilde{\mathcal{L}}_\lambda(\theta^{[m+0.5]})| > |\nabla_j \widetilde{\mathcal{L}}_\lambda(\theta^{[m+0.5]})|.$

Similarly, we conduct a proximal coordinate gradient descent iteration over the coordinate $j$, and obtain an auxiliary solution $w_j^{[m+1]}$. By definition of the soft thresholding function, we rewrite $w_k^{[m+1]}$ and $w_j^{[m+1]}$ as

$$w_k^{[m+1]} = -\frac{z_k}{L}\nabla_k\widetilde{\mathcal{L}}_\lambda(\theta^{[m+0.5]}) \quad \text{and} \quad w_j^{[m+1]} = -\frac{z_j}{L}\nabla_j\widetilde{\mathcal{L}}_\lambda(\theta^{[m+0.5]}),$$

where $z_k$ and $z_j$ are defined as

$$z_k = 1 - \frac{\lambda}{|\nabla_k\widetilde{\mathcal{L}}_\lambda(\theta^{[m+0.5]})|} \quad \text{and} \quad z_j = 1 - \frac{\lambda}{|\nabla_j\widetilde{\mathcal{L}}_\lambda(\theta^{[m+0.5]})|}.$$

By (C.53), we know $z_k \geq z_j$. Moreover, we define

$$(\text{C.54}) \quad z = \frac{|\nabla_j\widetilde{\mathcal{L}}_\lambda(\theta^{[m+0.5]})|}{|\nabla_k\widetilde{\mathcal{L}}_\lambda(\theta^{[m+0.5]})|} \cdot z_j \quad \text{and} \quad \widetilde{w}_k^{[m+1]} = -\frac{z}{L}\nabla_k\widetilde{\mathcal{L}}_\lambda(\theta^{[m+0.5]}).$$

Note that we have $|\widetilde{w}_k^{[m+1]}| = |w_j^{[m+1]}|$. We then consider

$$\mathcal{Q}_{\lambda,k,L}(\widetilde{w}_k^{[m+1]}; \theta^{[m+0.5]}) - \widetilde{\mathcal{L}}_\lambda(\theta^{[m+0.5]})$$
$$= -\frac{z}{L}|\nabla_k\widetilde{\mathcal{L}}_\lambda(\theta^{[m+0.5]})|^2 + \frac{L}{2}|\widetilde{w}_k^{[m+1]}|^2 + \lambda|\widetilde{w}_k^{[m+1]}| + \lambda\|\theta_{\setminus k}^{[m+0.5]}\|_1$$
$$\stackrel{(i)}{=} -\frac{z_j}{L}|\nabla_k\widetilde{\mathcal{L}}_\lambda(\theta^{[m+0.5]})| \cdot |\nabla_j\widetilde{\mathcal{L}}_\lambda(\theta^{[m+0.5]})| + \frac{L}{2}|\widetilde{w}_k^{[m+1]}|^2 + \lambda|\widetilde{w}_k^{[m+1]}| + \lambda\|\theta_{\setminus k}^{[m+0.5]}\|_1$$
$$\stackrel{(ii)}{<} -\frac{z_j}{L}|\nabla_j\widetilde{\mathcal{L}}_\lambda(\theta^{[m+0.5]})|^2 + \frac{L}{2}|w_j^{[m+1]}|^2 + \lambda|w_j^{[m+1]}| + \lambda\|\theta_{\setminus k}^{[m+0.5]}\|_1$$
$$= \mathcal{Q}_{\lambda,k,L}(w_j^{[m+1]}; \theta^{[m+0.5]}) - \widetilde{\mathcal{L}}_\lambda(\theta^{[m+0.5]}),$$

where (i) comes from (C.54) and (ii) comes from (C.47). We then have

$$(\text{C.55}) \quad \mathcal{Q}_{\lambda,k,L}(w_k^{[m+1]}; \theta^{[m+0.5]}) \leq \mathcal{Q}_{\lambda,k,L}(\widetilde{w}_k^{[m+1]}; \theta^{[m+0.5]})$$
$$\leq \mathcal{Q}_{\lambda,j,L}(w_j^{[m+1]}; \theta^{[m+0.5]}),$$

where the last inequality comes from (C.48). Thus, (C.55) guarantees

$$(\text{C.56}) \quad \mathcal{Q}_{\lambda,k_m,L}(w_{k_m}^{[m+0.5]}; \theta^{[m+0.5]}) = \min_{j \in \overline{\mathcal{A}}_m} \mathcal{Q}_{\lambda,j,L}(w_j^{[m+1]}; \theta^{[m+0.5]}),$$

where $k_m = \mathrm{argmax}_{k \in \overline{\mathcal{A}}_m} |\nabla\widetilde{\mathcal{L}}_k(\theta)^{[m+0.5]}|$.

For any $j \in \mathcal{A}_m$, we construct two auxiliary solutions $w^{[m+1]}$ and $v^{[m+1]}$,

$$w_j^{[m+1]} = \underset{v_j}{\mathrm{argmin}}\, \mathcal{Q}_{\lambda,j,L}(v_j; \theta^{[m+0.5]}) \quad \text{and} \quad v_j^{[m+1]} = \underset{v_j}{\mathrm{argmin}}\, \mathcal{F}_\lambda(v_j, \theta_{\setminus j}^{[m+0.5]}).$$

Recall $\theta^{[m+0.5]}$ is the output solution of the previous inner loop, i.e., $\theta^{[m+0.5]} = \theta^{(t+1)}$. By the restricted convexity of $\mathcal{F}_\lambda(\theta)$, we have

$$\mathcal{F}_\lambda(\theta^{(t+1)}) - \mathcal{F}_\lambda(v_j^{[m+1]}, \theta_{\setminus j}^{(t+1)})$$
$$\leq \frac{(\nabla_j \widetilde{\mathcal{L}}_\lambda(\theta^{(t+1)}) + \lambda \xi_j)^2}{2\widetilde{\rho}_-(1)} \leq \frac{\|\nabla_\mathcal{A} \widetilde{\mathcal{L}}_\lambda(\theta^{(t+1)}) + \lambda \xi_\mathcal{A}\|_2^2}{2\widetilde{\rho}_-(1)},$$

for some $\xi_\mathcal{A} \in \partial \|\theta_\mathcal{A}^{(t+1)}\|_1$. Since the inner loop terminates when $\|\theta^{(t+1)} - \theta^{(t)}\|_2^2 \leq \tau^2 \lambda^2$, we have

$$(\text{C.57}) \quad \mathcal{F}_\lambda(\theta^{(t+1)}) - \mathcal{F}_\lambda(v_j^{[m+1]}, \theta_{\setminus j}^{(t+1)})$$
$$\leq \frac{(s^* + 2\widetilde{s})\rho_+^2(s^* + 2\widetilde{s})\|\theta^{(t+1)} - \theta^{(t)}\|_2^2}{2\widetilde{\rho}_-(1)} \leq \frac{\delta^2 \lambda^2}{2L},$$

where the last equality comes from Assumption 3.7. Thus, (C.57) implies

$$(\text{C.58}) \quad \mathcal{Q}_{\lambda, j, L}(w_j^{[m+1]}; \theta^{[m+0.5]}) - \mathcal{F}_\lambda(\theta^{[m+0.5]})$$
$$\geq \mathcal{F}_\lambda(\theta^{(t+1)}) - \mathcal{F}_\lambda(v_j^{[m+1]}, \theta_{\setminus j}^{(t+1)}) \geq -\frac{\delta^2 \lambda^2}{2L}.$$

Since $j$ is arbitrarily selected from $\mathcal{A}_m$, by (C.52) and (C.58), we have

$$(\text{C.59}) \qquad \mathcal{Q}_{\lambda, k_m, L}(w_{k_m}^{[m+0.5]}; \theta^{[m+0.5]}) \leq \min_{j \in \mathcal{A}_m} \mathcal{Q}_{\lambda, j, L}(w_j^{[m+1]}; \theta^{[m+0.5]}).$$

Combining (C.56) with (C.59), we have

$$\mathcal{Q}_{\lambda, k_m, L}(w_{k_m}^{[m+0.5]}; \theta^{[m+0.5]}) = \min_j \mathcal{Q}_{\lambda, j, L}(w_j^{[m+1]}; \theta^{[m+0.5]}).$$

□

### C.12. Proof of Lemma 7.7.

PROOF. Define $\mathcal{D}_m = \{w \mid w \in \mathbb{R}^d, w_{\overline{\mathcal{B}}_m} = 0\}$, we have

$$\mathcal{J}_{\lambda, L}(w^{[m+1]}; \theta^{[m+0.5]}) = \min_{w \in \mathcal{D}_m} \mathcal{J}_{\lambda, L}(w; \theta^{[m+0.5]}) = \min_{w \in \mathcal{D}_m} \widetilde{\mathcal{L}}_\lambda(\theta^{[m+0.5]})$$
$$+ (w - \theta^{[m+0.5]})^\top \nabla \widetilde{\mathcal{L}}_\lambda(\theta^{[m+0.5]}) + \lambda \|w\|_1 + \frac{L}{2}\|w - \theta^{[m+0.5]}\|_2^2$$

$$\leq \min_{w \in \mathcal{D}_m} \mathcal{F}_\lambda(w) + \frac{(L - \rho_-(s^* + 2\widetilde{s}))}{2}\|w - \theta^{[m+0.5]}\|_2^2,$$

where the last inequality coms from the restricted convexity of $\widetilde{\mathcal{L}}_\lambda(\theta)$, i.e.,

$$\widetilde{\mathcal{L}}_\lambda(w) \leq \widetilde{\mathcal{L}}_\lambda(\theta^{[m+0.5]}) + (w - \theta^{[m+0.5]})^\top \nabla \widetilde{\mathcal{L}}_\lambda(\theta^{[m+0.5]}) + \frac{\rho_-(s^* + 2\widetilde{s})}{2}\|w - \theta^{[m+0.5]}\|_2^2.$$

Let $w = z\overline{\theta}^\lambda + (1-z)\theta^{[m+0.5]})$ for $z \in [0,1]$. Then we have

$$(\text{C.60}) \quad \mathcal{J}_{\lambda,L}(w^{[m+1]}; \theta^{[m+0.5]})$$
$$\leq \min_{z \in [0,1]} \mathcal{F}_\lambda(z\overline{\theta}^\lambda + (1-z)\theta^{[m+0.5]}) + \frac{z^2(L - \rho_-(s^* + 2\widetilde{s}))}{2}\|\overline{\theta}^\lambda - \theta^{[m+0.5]}\|_2^2$$
$$\leq \mathcal{F}_\lambda(\theta^{[m+0.5]}) + \min_{z \in [0,1]} z[\mathcal{F}_\lambda(\overline{\theta}^\lambda) - \mathcal{F}_\lambda(\theta^{[m+0.5]})]$$
$$+ \frac{(z^2 L - z\rho_-(s^* + 2\widetilde{s}))}{2}\|\overline{\theta}^\lambda - \theta^{[m+0.5]}\|_2^2,$$

where the last inequality comes from the restricted convexity of $\mathcal{F}_\lambda(\theta)$, i.e.,

$$\mathcal{F}_\lambda(z\overline{\theta}^\lambda + (1-z)\theta^{[m+0.5]}) + \frac{z(1-z)\rho_-(s^* + 2\widetilde{s})}{2}\|\overline{\theta}^\lambda - \theta^{[m+0.5]}\|_2^2$$
$$\leq z\mathcal{F}_\lambda(\overline{\theta}^\lambda) + (1-z)\mathcal{F}_\lambda(\theta^{[m+0.5]}).$$

By the restricted convexity of $\mathcal{F}_\lambda(\theta)$, we have

$$(\text{C.61}) \quad \|\overline{\theta}^\lambda - \theta^{[m+0.5]}\|_2^2 \leq \frac{2[\mathcal{F}_\lambda(\theta^{[m+0.5]}) - \mathcal{F}_\lambda(\overline{\theta}^\lambda)]}{\rho_-(s^* + 2\widetilde{s})}.$$

Combining (C.61) with (C.60), we obtain

$$(\text{C.62}) \quad \mathcal{J}_{\lambda,L}(w^{[m+1]}; \theta^{(t)}) - \mathcal{F}_\lambda(\theta^{[m+0.5]})$$
$$\leq \min_{z \in [0,1]} \left(\frac{z^2 L}{\rho_-(s^* + 2\widetilde{s})} - 2z\right)[\mathcal{F}_\lambda(\theta^{[m+0.5]}) - \mathcal{F}_\lambda(\overline{\theta}^\lambda)].$$

By setting $z = \widetilde{\rho}_-(s^* + 2\widetilde{s})/L$, we minimize the R.H.S of (C.62) and obtain

$$\mathcal{F}_\lambda(\theta^{[m+0.5]}) - \mathcal{J}_{\lambda,L}(w^{[m+1]}; \theta^{(t)}) \geq \frac{\widetilde{\rho}_-(s^* + 2\widetilde{s})}{L}[\mathcal{F}_\lambda(\theta^{[m+0.5]}) - \mathcal{F}_\lambda(\overline{\theta}^\lambda)].$$

$\square$

### C.13. Proof of Lemma C.1.

PROOF. We prove the uniqueness of $\overline{\theta}^\lambda$ by contradiction. Assume that there exist two different local optima $\overline{\theta}^\lambda$ and $\widetilde{\theta}^\lambda$. Let $\bar{\xi} \in \partial\|\overline{\theta}^\lambda\|_1$ and $\widetilde{\xi} \in \partial\|\widetilde{\theta}^\lambda\|_1$ be two subgradient vectors satisfying

(C.63) $\qquad \nabla\widetilde{\mathcal{L}}_\lambda(\overline{\theta}^\lambda) + \lambda\bar{\xi} = 0 \quad \text{and} \quad \nabla\widetilde{\mathcal{L}}_\lambda(\widetilde{\theta}^\lambda) + \lambda\widetilde{\xi} = 0.$

By the restricted strong convexity of $\mathcal{F}_\lambda(\theta)$, we obtain

$$\mathcal{F}_\lambda(\overline{\theta}^\lambda) \geq \mathcal{F}_\lambda(\widetilde{\theta}^\lambda) + (\overline{\theta}^\lambda - \widetilde{\theta}^\lambda)^\top(\nabla\widetilde{\mathcal{L}}_\lambda(\widetilde{\theta}^\lambda) + \lambda\widetilde{\xi}) + \frac{\widetilde{\rho}_-(s^* + 2\widetilde{s})}{2}\|\overline{\theta}^\lambda - \widetilde{\theta}^\lambda\|_2^2,$$

$$\mathcal{F}_\lambda(\widetilde{\theta}^\lambda) \geq \mathcal{F}_\lambda(\overline{\theta}^\lambda) + (\widetilde{\theta}^\lambda - \overline{\theta}^\lambda)^\top(\nabla\widetilde{\mathcal{L}}_\lambda(\overline{\theta}^\lambda) + \lambda\bar{\xi}) + \frac{\widetilde{\rho}_-(s^* + 2\widetilde{s})}{2}\|\overline{\theta}^\lambda - \widetilde{\theta}^\lambda\|_2^2,$$

since $\|\overline{\theta}_{\overline{\mathcal{S}}}^\lambda\|_0 \leq \widetilde{s}$ and $\|\widetilde{\theta}_{\overline{\mathcal{S}}}^\lambda\|_0 \leq \widetilde{s}$. Combining the above two inequalities with (C.63), we have $\|\overline{\theta}^\lambda - \widetilde{\theta}^\lambda\|_2^2 = 0$ implying $\overline{\theta}^\lambda = \widetilde{\theta}^\lambda$. That is contradicted by our assumption. Thus, the local optimum $\overline{\theta}^\lambda$ is unique. $\qquad \square$

### C.14. Proof of Lemma 3.11.

PROOF. For notational simplicity, we define $\Delta = \theta - \theta^*$. Let $\widetilde{\xi} \in \partial\|\theta\|_1$ be a subgradient vector satisfying

$$\mathcal{K}_{\lambda_{K-1}}(\theta) = \|\nabla\widetilde{\mathcal{L}}_{\lambda_{K-1}}(\theta) + \lambda_{K-1}\widetilde{\xi}\|_\infty.$$

We then consider the following decomposition

(C.64) $\quad \mathcal{K}_{\lambda_K}(\theta) \leq \|\nabla\widetilde{\mathcal{L}}_{\lambda_K}(\theta) + \lambda_K\widetilde{\xi}\|_\infty$
$\qquad \leq \|\nabla\widetilde{\mathcal{L}}_{\lambda_{K-1}}(\theta) + \lambda_{K-1}\widetilde{\xi}\|_\infty + \|\lambda_K\widetilde{\xi} - \lambda_{K-1}\widetilde{\xi}\|_\infty$
$\qquad + \|\nabla\mathcal{H}_{\lambda_K}(\theta) - \nabla\mathcal{H}_{\lambda_{K-1}}(\theta)\|_\infty \overset{(i)}{\leq} \delta_{K-1}\lambda_{K-1} + 3(1-\eta)\lambda_{K-1} \overset{(ii)}{\leq} \frac{\lambda_K}{4},$

where (i) comes from (R.4) in Lemma B.1, and (ii) comes from $\delta_{K-1} \leq 1/8$ and $1 - \eta \leq 1/24$ in Assumption 3.1.

We then proceed to characterize the statistical error of $\theta$ in terms of $\lambda_K$. For notational simplicity, we omit the index $K$ and denote $\lambda_K$ by $\lambda$. Since (C.64) implies that $\theta$ satisfies the approximate KKT condition for $\lambda$, then by the restricted convexity of $\widetilde{\mathcal{L}}_\lambda(\theta)$, we have

(C.65) $\quad \mathcal{F}_\lambda(\theta^*) - \frac{\widetilde{\rho}_-(s^* + \widetilde{s})}{2}\|\Delta\|_2^2 \geq \mathcal{F}_\lambda(\theta) - \Delta^\top(\nabla\widetilde{\mathcal{L}}_\lambda(\theta) + \lambda\widetilde{\xi})$
$\qquad \overset{(i)}{\geq} \mathcal{F}_\lambda(\theta) - \|\nabla\widetilde{\mathcal{L}}_\lambda(\theta) + \lambda\widetilde{\xi}\|_\infty \cdot \|\Delta\|_1 \overset{(ii)}{\geq} \mathcal{F}_\lambda(\theta) - \frac{\lambda}{4}\|\Delta\|_1,$

where (i) comes from Hölder's inequality and (ii) comes from (C.64). We then rewrite (C.65) as

$$\text{(C.66)} \quad \lambda\|\theta^*\|_1 - \lambda\|\theta\|_1 + \frac{\lambda}{4}\|\Delta\|_1 \geq \widetilde{\mathcal{L}}_\lambda(\theta) - \widetilde{\mathcal{L}}_\lambda(\theta^*) + \frac{\widetilde{\rho}_-(s^* + \widetilde{s})}{2}\|\Delta\|_2^2.$$

By the restricted convexity of $\widetilde{\mathcal{L}}_\lambda(\theta)$ again, we have

$$\text{(C.67)} \quad \widetilde{\mathcal{L}}_\lambda(\theta) - \widetilde{\mathcal{L}}_\lambda(\theta^*) - \frac{\widetilde{\rho}_-(s^* + \widetilde{s})}{2}\|\Delta\|_2^2 \geq \Delta^\top \nabla \widetilde{\mathcal{L}}_\lambda(\theta^*)$$

$$\stackrel{(i)}{=} \Delta_\mathcal{S}^\top \nabla_\mathcal{S} \mathcal{L}(\theta^*) + \Delta_{\overline{\mathcal{S}}}^\top \nabla_{\overline{\mathcal{S}}} \mathcal{L}(\theta^*) + \Delta_\mathcal{S}^\top \nabla_\mathcal{S} \mathcal{H}_\lambda(\theta^*)$$

$$\stackrel{(ii)}{\geq} -\|\Delta_\mathcal{S}\|_1 \|\nabla \mathcal{L}(\theta^*)\|_\infty - \|\Delta_{\overline{\mathcal{S}}}\|_1 \|\nabla \mathcal{L}(\theta^*)\|_\infty - \|\Delta_\mathcal{S}\|_1 \|\nabla_\mathcal{S} \mathcal{H}_\lambda(\theta^*)\|_\infty,$$

where (i) comes from $\nabla_{\overline{\mathcal{S}}} \mathcal{H}_\lambda(\theta^*) = 0$ by (R.3) of Lemma B.1, and (ii) comes from Hölder's inequality. Assumption 3.1 and (R.2) of Lemma B.1 imply

$$\text{(C.68)} \quad \|\nabla \mathcal{L}(\theta^*)\|_\infty \leq \lambda/4 \quad \text{and} \quad \|\nabla_\mathcal{S} \mathcal{H}_\lambda(\theta^*)\|_\infty \leq \lambda.$$

Combining (C.67) with (C.68), we obtain

$$\text{(C.69)} \quad \widetilde{\mathcal{L}}_\lambda(\theta) - \widetilde{\mathcal{L}}_\lambda(\theta^*) \geq -\frac{3}{2}\lambda\|\Delta_\mathcal{S}\|_1 - \frac{\lambda}{2}\|\Delta_{\overline{\mathcal{S}}}\|_1 + \widetilde{\rho}_-(s^* + \widetilde{s})\|\Delta\|_2^2.$$

Plugging (C.69) and

$$\|\theta^*\|_1 - \|\theta\|_1 = \|\theta^*_\mathcal{S}\|_1 - (\|\theta_\mathcal{S}\|_1 + \|\Delta_{\overline{\mathcal{S}}}\|_1) \leq \|\Delta_\mathcal{S}\|_1 - \|\Delta_{\overline{\mathcal{S}}}\|_1$$

into (C.66), we obtain

$$\text{(C.70)} \quad \frac{11\lambda}{4}\|\Delta_\mathcal{S}\|_1 \geq \frac{\lambda}{4}\|\Delta_{\overline{\mathcal{S}}}\|_1 + \widetilde{\rho}_-(s^* + \widetilde{s})\|\Delta\|_2^2.$$

By simple manipulation, (C.70) implies

$$\text{(C.71)} \quad \widetilde{\rho}_-(s^* + \widetilde{s})\|\Delta\|_2^2 \leq \frac{11\lambda}{4}\|\Delta_\mathcal{S}\|_1 \leq \frac{11\lambda}{4}\sqrt{s^*}\|\Delta_\mathcal{S}\|_2 \leq \frac{11\lambda}{4}\sqrt{s^*}\|\Delta\|_2,$$

where the second inequality comes from the fact that $\Delta_\mathcal{S}$ only contains $s^*$ entries. By simple manipulation again, (C.71) implies

$$\text{(C.72)} \quad \|\Delta\|_2 \leq \frac{11\lambda\sqrt{s^*}}{4\widetilde{\rho}_-(s^* + \widetilde{s})}.$$

Meanwhile, (C.70) also implies

$$\text{(C.73)} \quad \|\Delta_{\overline{\mathcal{S}}}\|_1 \leq 11\|\Delta_\mathcal{S}\|_1.$$

Combining (C.72) with (C.73), we obtain

$$\text{(C.74)} \quad \|\Delta\|_1 \leq 11\|\Delta_{\mathcal{S}}\|_1 \leq 11\sqrt{s^*}\|\Delta_{\mathcal{S}}\|_2 \leq 11\sqrt{s^*}\|\Delta\|_2 \leq \frac{31\lambda s^*}{\widetilde{\rho}_-(s^* + \widetilde{s})}.$$

Plugging (C.74) and (C.72) into (C.65), we have

$$\mathcal{F}_\lambda(\theta) - \mathcal{F}_\lambda(\theta^*) \leq \delta\lambda\|\Delta\|_1 \leq \frac{4\lambda^2 s^*}{\widetilde{\rho}_-(s^* + \widetilde{s})}.$$

$\square$

## APPENDIX D: LEMMAS FOR GENERAL LOSS FUNCTIONS

### D.1. Proof of Lemma 4.3.

PROOF. For notational simplicity, we denote $\theta^{\mathsf{relax}}$ by $\theta$ and write $\widetilde{\mathcal{F}}_\lambda(\theta) = \mathcal{L}(\theta) + \lambda\|\theta\|_1$. Let $\widetilde{\xi} \in \partial\|\theta\|_1$ be a subgradient vector satisfying

$$\|\nabla\mathcal{L}(\theta) + \lambda\widetilde{\xi}\|_\infty = \min_{\xi \in \partial\|\theta\|_1} \|\nabla\mathcal{L}(\theta) + \lambda\xi\|_\infty.$$

For notational simplicity, we define $\Delta = \theta^* - \theta$. Since $\widetilde{\mathcal{F}}_\lambda(\theta)$ is a convex function, we have

$$\text{(D.1)} \quad \widetilde{\mathcal{F}}_\lambda(\theta^*) \geq \widetilde{\mathcal{F}}_\lambda(\theta) - \Delta^\top(\nabla\mathcal{L}(\theta) + \lambda\widetilde{\xi})$$
$$\geq \widetilde{\mathcal{F}}_\lambda(\theta) - \|\Delta\|_1\|\nabla\mathcal{L}(\theta) + \lambda\widetilde{\xi}\|_\infty \geq \widetilde{\mathcal{F}}_\lambda(\theta) - \frac{\lambda}{8}\|\Delta\|_1,$$

where the second inequality comes from Hölder's inequality, and the last inequality comes from (4.6).

To establish the statistical properties of $\theta$, we need to verify that $\theta$ satisfies $\|\theta - \theta^*\|_2 \leq R$ such that the restricted strong convexity holds for $\theta$. We prove it by contradiction. We first assume $\|\theta - \theta^*\|_2 \geq R$. Then there exists some $z \in (0, 1)$ such that

$$\text{(D.2)} \quad \widetilde{\theta} = (1-z)\theta + z\theta^* \quad \text{and} \quad \|\widetilde{\theta} - \theta^*\|_2 = R.$$

Then by the convexity of $\widetilde{\mathcal{F}}_\lambda(\theta)$ again, (D.1) and (D.2) imply

$$\text{(D.3)} \quad \widetilde{\mathcal{F}}_\lambda(\widetilde{\theta}) \leq (1-z)\widetilde{\mathcal{F}}_\lambda(\theta) + z\widetilde{\mathcal{F}}_\lambda(\theta^*)$$
$$\leq (1-z)\widetilde{\mathcal{F}}_\lambda(\theta^*) + \frac{(1-z)\lambda}{8}\|\Delta\|_1 + z\widetilde{\mathcal{F}}_\lambda(\theta^*) \leq \widetilde{\mathcal{F}}_\lambda(\theta^*) + \frac{\lambda}{8}\|\widetilde{\Delta}\|_1,$$

where the last inequality comes from the fact

$$\|\widetilde{\Delta}\|_1 = \|\widetilde{\theta} - \theta^*\|_1 = \|(1-z)\theta + z\theta^* - \theta^*\|_1 = (1-z)\|\Delta\|_1.$$

By simple manipulation, we can rewrite (D.3) as

(D.4) $$\mathcal{L}(\widetilde{\theta}) - \mathcal{L}(\theta^*) \leq \lambda\|\theta^*\|_1 - \lambda\|\widetilde{\theta}\|_1 + \frac{\lambda}{8}\|\widetilde{\Delta}\|_1.$$

By the convexity of $\mathcal{L}(\theta)$, we have

(D.5) $$\mathcal{L}(\widetilde{\theta}) - \mathcal{L}(\theta^*) \geq \widetilde{\Delta}^\top \nabla \mathcal{L}(\theta^*)$$
$$\geq -\|\widetilde{\Delta}\|_1 \|\nabla \mathcal{L}(\theta^*)\|_\infty \geq -\frac{\lambda}{8}\|\widetilde{\Delta}_\mathcal{S}\|_1 - \frac{\lambda}{8}\|\widetilde{\Delta}_{\overline{\mathcal{S}}}\|_1,$$

where the last inequality comes from our assumption $\lambda \geq 8\|\nabla \mathcal{L}(\theta^*)\|_\infty$. By the decomposability of the $\ell_1$ norm, we have

(D.6) $$\|\theta^*\|_1 - \|\widetilde{\theta}\|_1 + \frac{1}{8}\|\widetilde{\Delta}\|_1$$
$$= \|\theta^*_\mathcal{S}\|_1 - (\|\widetilde{\theta}_\mathcal{S}\|_1 + \|\widetilde{\Delta}_{\overline{\mathcal{S}}}\|_1) + \frac{1}{8}\|\widetilde{\Delta}_\mathcal{S}\|_1 + \frac{1}{8}\|\widetilde{\Delta}_{\overline{\mathcal{S}}}\|_1$$
$$\leq \frac{9}{8}\|\widetilde{\Delta}_\mathcal{S}\|_1 - (1-\delta)\|\widetilde{\Delta}_{\overline{\mathcal{S}}}\|_1 \leq \frac{9}{8}\|\widetilde{\Delta}_\mathcal{S}\|_1 - \frac{7}{8}\|\widetilde{\Delta}_{\overline{\mathcal{S}}}\|_1.$$

Combining (D.4) with (D.5) and (D.6), we obtain

(D.7) $$\|\widetilde{\Delta}_{\overline{\mathcal{S}}}\|_1 \leq \frac{5}{3}\|\widetilde{\Delta}_\mathcal{S}\|_1.$$

To establish the statistical properties of $\widetilde{\theta}$, we define the following sets:

$$\mathcal{S}_0 = \left\{ j \mid j \in \overline{\mathcal{S}},\ \sum_{k \in \overline{\mathcal{S}}} \mathbb{1}_{\{|\widetilde{\theta}_k| \geq |\widetilde{\theta}_j|\}} \leq \widetilde{s} \right\},$$
$$\mathcal{S}_1 = \left\{ j \mid j \in \overline{\mathcal{S}} \setminus \mathcal{S}_0,\ \sum_{k \in \overline{\mathcal{S}} \setminus \mathcal{S}_0} \mathbb{1}_{\{|\widetilde{\theta}_k| \geq |\widetilde{\theta}_j|\}} \leq \widetilde{s} \right\},$$
$$\mathcal{S}_2 = \left\{ j \mid j \in \overline{\mathcal{S}} \setminus (\mathcal{S}_0 \cup \mathcal{S}_1),\ \sum_{k \in \overline{\mathcal{S}} \setminus (\mathcal{S}_0 \cup \mathcal{S}_1)} \mathbb{1}_{\{|\widetilde{\theta}_k| \geq |\widetilde{\theta}_j|\}} \leq \widetilde{s} \right\},$$
$$\mathcal{S}_3 = \left\{ j \mid j \in \overline{\mathcal{S}} \setminus (\mathcal{S}_0 \cup \mathcal{S}_1 \cup \mathcal{S}_2),\ \sum_{k \in \overline{\mathcal{S}} \setminus (\mathcal{S}_0 \cup \mathcal{S}_1 \cup \mathcal{S}_2)} \mathbb{1}_{\{|\widetilde{\theta}_k| \geq |\widetilde{\theta}_j|\}} \leq \widetilde{s} \right\}, \ldots$$

Before we proceed with the proof, we introduce the following lemma.

**Lemma D.1** (Lemma 6.9 in Bühlmann and van de Geer (2011)). *Let* $b_1 \geq b_2 \geq \ldots \geq 0$. *For* $s \in \{1, 2, \ldots\}$, *we have*

$$\sqrt{\sum_{j \geq s+1} b_j^2} \leq \sum_{k=1}^\infty \sqrt{\sum_{j=ks+1}^{(k+1)s} b_j^2} \leq \sqrt{s} \sum_{k=1}^\infty b_j.$$

The proof of Lemma D.1 is provided in Bühlmann and van de Geer (2011), and therefore is omitted. By Lemma D.1 and (D.7), we have

$$\sum_{j\geq 1}\|\widetilde{\Delta}_{\mathcal{S}_j}\|_1 \leq \frac{1}{\sqrt{\widetilde{s}}}\|\widetilde{\Delta}_{\overline{\mathcal{S}}}\|_1 \leq \frac{5}{3}\sqrt{\frac{s^*}{\widetilde{s}}}\|\widetilde{\Delta}_{\mathcal{S}}\|_2 \leq \frac{5}{3}\sqrt{\frac{s^*}{\widetilde{s}}}\|\widetilde{\Delta}_{\mathcal{A}}\|_2,$$

where $\mathcal{A} = \mathcal{S} \cup \mathcal{S}_0$. By definition of the largest sparse eigenvalue and Assumption 3.5, given $\ddot{\theta} = z\widetilde{\theta} + (1-z)\theta^*$ for any $z \in [0,1]$ and $j \geq 1$, we have

$$\left|\widetilde{\Delta}_{\mathcal{S}_j}^\top \nabla^2_{\mathcal{S}_j \mathcal{A}} \mathcal{L}(\ddot{\theta}) \widetilde{\Delta}_{\mathcal{A}}\right| \leq \rho_+(s^* + \widetilde{s})\|\widetilde{\Delta}_{\mathcal{S}_j}\|_2 \|\widetilde{\Delta}_{\mathcal{A}}\|_2,$$

which further implies

$$(D.8) \quad \begin{aligned} |\widetilde{\Delta}_{\overline{\mathcal{A}}}^\top \nabla^2_{\overline{\mathcal{A}}\mathcal{A}} \mathcal{L}(\ddot{\theta}) \widetilde{\Delta}_{\mathcal{A}}| &\leq \sum_{j\geq 1} |\widetilde{\Delta}_{\mathcal{S}_j}^\top \nabla^2_{\mathcal{S}_j \mathcal{A}} \mathcal{L}(\ddot{\theta}) \widetilde{\Delta}_{\mathcal{A}}| \\ &= \frac{5\rho_+(s^* + 2\widetilde{s})}{3}\|\widetilde{\Delta}_{\mathcal{A}}\|_2^2 \sqrt{\frac{s^*}{\widetilde{s}}}. \end{aligned}$$

By definition of the smallest sparse eigenvalue and Assumption 3.5, we have

$$(D.9) \quad \frac{\widetilde{\Delta}_{\mathcal{A}}^\top \nabla^2_{\mathcal{A}\mathcal{A}} \mathcal{L}(\ddot{\theta}) \widetilde{\Delta}_{\mathcal{A}}}{\|\widetilde{\Delta}_{\mathcal{A}}\|_2^2} \geq \rho_-(s^* + \widetilde{s}).$$

Combining (D.8) with (D.9), we have

$$|\widetilde{\Delta}_{\overline{\mathcal{A}}}^\top \nabla^2_{\overline{\mathcal{A}}\mathcal{A}} \mathcal{L}(\ddot{\theta}) \widetilde{\Delta}_{\mathcal{A}}| \leq \frac{5\rho_+(s^* + \widetilde{s})}{3\rho_-(s^* + \widetilde{s})} \sqrt{\frac{s^*}{\widetilde{s}}} \widetilde{\Delta}_{\mathcal{A}}^\top \nabla^2_{\mathcal{A}\mathcal{A}} \mathcal{L}(\ddot{\theta}) \widetilde{\Delta}_{\mathcal{A}},$$

which further implies

$$\frac{|\widetilde{\Delta}_{\overline{\mathcal{A}}}^\top \nabla^2_{\overline{\mathcal{A}}\mathcal{A}} \mathcal{L}(\ddot{\theta}) \widetilde{\Delta}_{\mathcal{A}}|}{|\widetilde{\Delta}_{\mathcal{A}}^\top \nabla^2_{\mathcal{A}\mathcal{A}} \mathcal{L}(\ddot{\theta}) \widetilde{\Delta}_{\mathcal{A}}|} \leq \frac{5\rho_+(s^* + \widetilde{s})}{3\rho_-(s^* + \widetilde{s})} \sqrt{\frac{s^*}{\widetilde{s}}}.$$

Eventually, we have

$$\begin{aligned} \frac{\Delta^\top \nabla^2 \mathcal{L}(\ddot{\theta}) \Delta}{\|\Delta_{\mathcal{A}}\|_2^2} &\geq \left(1 - \frac{|\widetilde{\Delta}_{\overline{\mathcal{A}}}^\top \nabla^2_{\overline{\mathcal{A}}\mathcal{A}} \mathcal{L}(\ddot{\theta}) \widetilde{\Delta}_{\mathcal{A}}|}{|\widetilde{\Delta}_{\mathcal{A}}^\top \nabla^2_{\mathcal{A}\mathcal{A}} \mathcal{L}(\ddot{\theta}) \widetilde{\Delta}_{\mathcal{A}}|}\right) \rho_-(s^* + \widetilde{s}) \\ (D.10) \quad &\geq \left(1 - \frac{9\rho_+(s^* + \widetilde{s})}{7\rho_-(s^* + \widetilde{s})}\sqrt{\frac{s^*}{\widetilde{s}}}\right) \rho_-(s^* + \widetilde{s}) \geq \frac{7\rho_-(s^* + \widetilde{s})}{8}, \end{aligned}$$

where the last inequality comes from Assumption 3.5. Then by the mean value theorem, we choose some $z$ such that

$$\mathcal{L}(\widetilde{\theta}) - \mathcal{L}(\theta^*) - \widetilde{\Delta}^\top \nabla \mathcal{L}(\theta^*) = \frac{1}{2}\widetilde{\Delta}^\top \nabla^2 \mathcal{L}(\ddot{\theta})\widetilde{\Delta} \geq \frac{7\rho_-(s^* + \widetilde{s})}{16}\|\widetilde{\Delta}_\mathcal{S}\|_2^2,$$

which implies

$$\mathcal{L}(\widetilde{\theta}) - \mathcal{L}(\theta^*) \geq \widetilde{\Delta}^\top \nabla \mathcal{L}(\theta^*) + \frac{7\rho_-(s^* + \widetilde{s})}{16}\|\widetilde{\Delta}_\mathcal{A}\|_2^2$$

$$\geq \frac{7\rho_-(s^* + \widetilde{s})}{16}\|\widetilde{\Delta}_\mathcal{A}\|_2^2 - \frac{\lambda}{8}\|\widetilde{\Delta}_\mathcal{S}\|_1 - \frac{\lambda}{8}\|\widetilde{\Delta}_{\overline{\mathcal{S}}}\|_1.$$

Then by (D.4) and (D.6), we have

$$\rho_-(s^* + \widetilde{s})\|\widetilde{\Delta}_\mathcal{S}\|_2^2 \leq \rho_-(s^* + \widetilde{s})\|\widetilde{\Delta}_\mathcal{A}\|_2^2 \leq \frac{20}{7}\lambda\|\widetilde{\Delta}_\mathcal{S}\|_1$$

$$\leq \frac{20}{7}\sqrt{s^*}\lambda\|\widetilde{\Delta}_\mathcal{S}\|_2 \leq \frac{20}{7}\sqrt{s^*}\lambda\|\widetilde{\Delta}_\mathcal{A}\|_2,$$

which further implies

(D.11) $\quad \|\widetilde{\Delta}_\mathcal{S}\|_2 \leq \|\widetilde{\Delta}_\mathcal{A}\|_2 \leq \dfrac{20\sqrt{s^*}\lambda}{7\rho_-(s^* + \widetilde{s})} \quad$ and $\quad \|\widetilde{\Delta}_\mathcal{S}\|_1 \leq \dfrac{20s^*\lambda}{7\rho_-(s^* + \widetilde{s})}.$

By Lemma D.1, (D.11) implies

$$\|\widetilde{\Delta}_{\overline{\mathcal{A}}}\|_2 \leq \frac{\|\widetilde{\Delta}_{\overline{\mathcal{S}}}\|_1}{\sqrt{s^*}} \leq \frac{5\|\widetilde{\Delta}_\mathcal{S}\|_1}{3\sqrt{s^*}} = \frac{24\sqrt{s^*}\lambda}{5\rho_-(s^* + \widetilde{s})}.$$

Combining the above results, we have

$$\|\widetilde{\Delta}\|_2 = \sqrt{\|\widetilde{\Delta}_\mathcal{A}\|_2^2 + \|\widetilde{\Delta}_{\overline{\mathcal{A}}}\|_2^2} \leq \frac{17\sqrt{s^*}\lambda}{3\rho_-(s^* + \widetilde{s})} < R,$$

where the last inequality comes from the initial condition of $\theta$. This conflicts with our assumption $\|\widetilde{\Delta}\|_2 = R$. Therefore we must have $\|\theta - \theta^*\|_2 \leq R$. Consequently, we repeat the above proof for $\theta$, and obtain

$$\|\Delta\|_2 \leq \frac{17\sqrt{s^*}\lambda}{3\rho_-(s^* + \widetilde{s})} \quad \text{and} \quad \|\Delta\|_1 = \|\Delta_\mathcal{S}\|_1 + \|\Delta_{\overline{\mathcal{S}}}\|_1 \leq \frac{23\sqrt{s^*}\lambda}{3\rho_-(s^* + \widetilde{s})}.$$

We now characterize the sparsity of $\theta$. By Assumption 3.1 and the initial condition of $\theta$, we have $\lambda = 2\lambda_N \geq 8\|\nabla \mathcal{L}(\theta^*)\|_\infty$, which further implies

(D.12) $\qquad \left|\{j \mid |\nabla_j \mathcal{L}(\theta^*)| \geq \lambda/8, \ j \in \overline{\mathcal{S}}\}\right| = 0.$

We then consider an arbitrary set $\mathcal{S}'$ such that

$$\mathcal{S}' = \{j \mid |\nabla_j \mathcal{L}(\theta) - \nabla_j \mathcal{L}(\theta^*)| \geq 5\lambda/8, \ j \in \overline{\mathcal{S}}\}.$$

Let $s' = |\mathcal{S}'|$. Then there exists $v$ such that

$$\|v\|_\infty = 1, \quad \|v\|_0 \leq s', \quad \text{and} \quad 5s'\lambda/8 \leq v^\top(\nabla \mathcal{L}(\theta) - \nabla \mathcal{L}(\theta^*)).$$

Since $\mathcal{L}(\theta)$ is twice differentiable, then by the mean value theorem, there exists some $z_1 \in [0, 1]$ such that

$$\ddot{\theta} = z_1 \theta + (1 - z_1)\theta^* \quad \text{and} \quad \nabla \mathcal{L}(\theta) - \nabla \mathcal{L}(\theta^*) = \nabla^2 \mathcal{L}(\ddot{\theta}) \Delta.$$

Then we have

$$\frac{5s'\lambda}{8} \leq v^\top \nabla^2 \mathcal{L}(\ddot{\theta}) \Delta \leq \sqrt{v^\top \nabla^2 \mathcal{L}(\ddot{\theta}) v} \sqrt{\Delta^\top \nabla^2 \mathcal{L}(\ddot{\theta}) \Delta}.$$

Since we have $\|v\|_0 \leq s'$, then we obtain

$$\begin{aligned}
\frac{3s'\lambda}{4} &\leq \sqrt{\rho_+(s')}\sqrt{s'}\sqrt{\Delta^\top(\nabla\mathcal{L}(\theta) - \nabla\mathcal{L}(\theta^*))} \\
&\leq \sqrt{\rho_+(s')}\sqrt{s'}\sqrt{\|\Delta\|_1 \cdot \|\nabla\mathcal{L}(\theta) - \nabla\mathcal{L}(\theta^*)\|_\infty} \\
&\leq \sqrt{\rho_+(s')}\sqrt{s'}\sqrt{\|\Delta\|_1 (\|\nabla\mathcal{L}(\theta)\|_\infty + \|\nabla\mathcal{L}(\theta^*)\|_\infty)} \\
&\leq \sqrt{\rho_+(s')}\sqrt{s'}\sqrt{\|\Delta\|_1 (\|\nabla\mathcal{L}(\theta) - \lambda\xi\|_\infty + \lambda\|\widetilde{\xi}\|_\infty + \|\nabla\mathcal{L}(\theta^*)\|_\infty)} \\
&\leq \sqrt{\rho_+(s')}\sqrt{s'}\sqrt{\frac{115 s^* \lambda^2}{12\rho_-(s^* + \widetilde{s})}}.
\end{aligned}$$

By simple manipulation, we have

$$\frac{5\sqrt{s'}}{8} \leq \sqrt{\rho_+(s')} \sqrt{\frac{115 s^*}{12\rho_-(s^* + \widetilde{s})}},$$

which implies

$$s' \leq \frac{184\rho_+(s')}{15\rho_-(s^* + \widetilde{s})} \cdot s^*.$$

Since $s' = |cS'|$ attains the maximum value such that $s' \leq \widetilde{s}$ for arbitrary defined subset $\mathcal{S}'$, we obtain $s' \leq \widetilde{s}$. Then by simple manipulation, we have

(D.13) $\quad \left|\{j \mid |\nabla_j\mathcal{L}(\theta) - \nabla_j\mathcal{L}(\theta^*)| \geq 5\lambda/8, \ j \in \overline{\mathcal{S}}\}\right| \leq 13\kappa s^* < \widetilde{s}.$

Thus, (D.12) and (D.13) imply

$$\big|\{j \mid |\nabla_j \mathcal{L}(\widehat{\theta}) + \tfrac{\lambda}{8} u_j| \geq 7\lambda/8,\ j \in \overline{\mathcal{S}} \cap \mathcal{A}\}\big| \leq \widetilde{s}$$

for any $u \in \mathbb{R}^d$ satisfying $\|u\|_\infty \leq 1$. Then there exists a $\xi_j \in \mathbb{R}$ satisfying

$$|\xi_j| \leq 1 \quad \text{and} \quad \nabla_j \widetilde{\mathcal{L}}_\lambda(\widehat{\theta}) + \lambda u_j/8 + \lambda \xi_j = 0,$$

for any $j \in \overline{\mathcal{S}} \cap \mathcal{A}$ satisfying $|\nabla_j \mathcal{L}(\widehat{\theta}) + \lambda u_j/8| \leq 7\lambda/8$. This further implies $\theta_j = 0$. Thus, we have $\|\theta_{\overline{\mathcal{A}}}\|_0 \leq \widetilde{s}$.

Since $\theta$ is sufficiently sparse, we know that the restricted convexity holds for $\theta$ and $\theta^*$. Then we refine our analysis for $\theta$. By the restricted convexity of $\widetilde{\mathcal{F}}_\lambda(\theta)$, we have

$$(D.14) \qquad \widetilde{\mathcal{F}}_\lambda(\theta^*) - \frac{\rho_-(s^* + \widetilde{s})}{2} \|\Delta\|_2^2 \\ \geq \widetilde{\mathcal{F}}_\lambda(\theta) - \Delta^\top (\nabla \mathcal{L}(\theta) + \lambda \widetilde{\xi}) \geq \widetilde{\mathcal{F}}_\lambda(\theta) - \frac{\lambda}{8}\|\Delta\|_1.$$

By simple manipulation, we rewrite (D.14) as

$$\mathcal{L}(\theta) - \mathcal{L}(\theta^*) \leq \lambda \|\theta^*\|_1 - \lambda \|\theta\|_1 + \frac{\lambda}{8}\|\Delta\|_1.$$

By the restricted convexity of $\mathcal{L}(\theta)$, we have

$$(D.15) \qquad \mathcal{L}(\theta) - \mathcal{L}(\theta^*) - \rho_-(s^* + \widetilde{s})\|\Delta\|_2^2 \geq -\frac{\lambda}{8}\|\Delta_\mathcal{S}\|_1 - \frac{\lambda}{8}\|\Delta_{\overline{\mathcal{S}}}\|_1,$$

where the last inequality comes from our assumption $\lambda \geq 8\|\nabla \mathcal{L}(\theta^*)\|_\infty$. By the decomposability of the $\ell_1$ norm, we have

$$(D.16) \qquad \|\theta^*\|_1 - \|\theta\|_1 + \frac{1}{8}\|\Delta\|_1 \\ = \|\theta^*_\mathcal{S}\|_1 - (\|\theta_\mathcal{S}\|_1 + \|\Delta_{\overline{\mathcal{S}}}\|_1) + \frac{1}{8}\|\Delta_\mathcal{S}\|_1 + \frac{1}{8}\|\Delta_{\overline{\mathcal{S}}}\|_1 \\ \leq \frac{9}{8}\|\Delta_\mathcal{S}\|_1 - (1-\delta)\|\Delta_{\overline{\mathcal{S}}}\|_1 \leq \frac{9}{8}\|\Delta_\mathcal{S}\|_1 - \frac{7}{8}\|\Delta_{\overline{\mathcal{S}}}\|_1,$$

where the last inequality comes from $\delta < 1/8$ in Assumption 3.1. Combining (D.7) and (D.4) with (D.15) and (D.16), we obtain

$$\rho_-(s^* + \widetilde{s})\|\Delta\|_2^2 \leq \frac{5\lambda}{4}\|\Delta_\mathcal{S}\|_1 \leq \frac{5\lambda\sqrt{s^*}}{4}\|\Delta_\mathcal{S}\|_2 \leq \frac{5\lambda\sqrt{s^*}}{4}\|\Delta_\mathcal{S}\|_2,$$

which implies that

$$\|\Delta\|_2 \leq \frac{5\lambda\sqrt{s^*}}{4\rho_-(s^* + \widetilde{s})} \quad \text{and} \quad \|\Delta_{\mathcal{S}}\|_1 \leq \sqrt{s^*}\|\Delta_{\mathcal{S}}\|_2 \leq \frac{5\lambda s^*}{4\rho_-(s^* + \widetilde{s})}.$$

By (D.7), we further have

(D.17) $$\|\Delta\|_1 \leq \frac{8}{3}\|\Delta_{\mathcal{S}}\|_1 \leq \frac{10\lambda s^*}{3\rho_-(s^* + \widetilde{s})}.$$

Plugging (D.17) into (D.14), we have

$$\widetilde{\mathcal{F}}_\lambda(\theta^*) \geq \widetilde{\mathcal{F}}_\lambda(\theta) + \frac{8\lambda^2 s^*}{7\rho_-(s^* + \widetilde{s})}.$$

By the concavity of $\mathcal{H}_\lambda(\theta)$ and Hölder's inequality, we have

$$\mathcal{H}_\lambda(\theta^{\text{relax}}) \leq \mathcal{H}_\lambda(\theta^*) + (\theta^{\text{relax}} - \theta^*)^\top \nabla \mathcal{H}_\lambda(\theta^*)$$
$$\leq \mathcal{H}_\lambda(\theta^*) + \|\theta^{\text{relax}} - \theta^*\|_1 \|\nabla \mathcal{H}_\lambda(\theta^*)\|_\infty.$$

Since we have $\|\mathcal{H}_\lambda(\theta)\|_\infty \leq \lambda$, by Lemma 4.3, we have

$$\mathcal{H}_\lambda(\theta^{\text{relax}}) \leq \mathcal{H}_\lambda(\theta^*) + \lambda \|\theta^{\text{relax}} - \theta^*\|_1 \leq \mathcal{H}_\lambda(\theta^*) + \triangle_{\lambda_0}.$$

Since $\mathcal{F}_{\lambda_0}(\theta) = \widetilde{\mathcal{F}}_{\lambda_0}(\theta) + \mathcal{H}_{\lambda_0}(\theta)$, by Lemma 4.3 again, we have $\mathcal{F}_{\lambda_0}(\theta^{\text{relax}}) \leq \mathcal{F}_\lambda(\theta^*) + \triangle_{\lambda_0}$. Thus, $\theta^{\text{relax}}$ is a proper initial solution for solving (1.1) with $\lambda_0$ using PICASSO. □

### D.2. Proof of Theorem 7.9.

PROOF. Let $\widetilde{\xi} \in \partial \|\theta\|_1$ be a subgradient vector satisfying $\mathcal{K}_\lambda(\theta) = \|\nabla \widetilde{\mathcal{L}}_\lambda(\theta) + \lambda \widetilde{\xi}\|_\infty$. By the restricted convexity of $\widetilde{\mathcal{L}}_{\lambda'}(\theta)$, we have

(D.18) $$\mathcal{F}_{\lambda'}(\theta) - \mathcal{F}_{\lambda'}(\overline{\theta}^{\lambda'}) \leq (\theta - \overline{\theta}^{\lambda'})^\top (\nabla \mathcal{L}(\theta) + \nabla \mathcal{H}_{\lambda'}(\theta) + \lambda'\widetilde{\xi})$$
$$= (\theta - \overline{\theta}^{\lambda'})^\top (\nabla \mathcal{L}(\theta) + \nabla \mathcal{H}_\lambda(\theta)$$
$$+ \lambda\widetilde{\xi} - \lambda\widetilde{\xi} + \lambda'\widetilde{\xi} - \nabla\mathcal{H}_\lambda(\theta) + \nabla\mathcal{H}_{\lambda'}(\theta))$$
$$\stackrel{(i)}{\leq} \|\theta - \overline{\theta}^{\lambda'}\|_1 (\|\nabla\mathcal{L}(\theta) + \nabla\mathcal{H}_\lambda(\theta) + \lambda\widetilde{\xi}\|_\infty$$
$$+ (\lambda - \lambda') + \|\nabla\mathcal{H}_\lambda(\theta) - \nabla\mathcal{H}_{\lambda'}(\theta)\|_\infty)$$
$$\stackrel{(ii)}{\leq} (\mathcal{K}_\lambda(\theta) + 3(\lambda - \lambda'))\|\theta - \overline{\theta}^{\lambda'}\|_1,$$

where (i) comes from Hölder's inequality and $\|\widetilde{\xi}\|_\infty \leq 1$, and (ii) comes from (R.3) of Lemma B.1. Meanwhile, since we have

$$\|\overline{\theta}_{\overline{\mathcal{S}}}^{\lambda'}\|_0 \leq \widetilde{s}, \quad \mathcal{K}_{\lambda'}(\overline{\theta}^{\lambda'}) = 0 \leq \lambda'/4, \quad \|\theta_{\overline{\mathcal{S}}}\|_0 \leq \widetilde{s}, \text{ and } \mathcal{K}_\lambda(\theta) \leq \lambda/4,$$

following similar lines to the proof of Theorem 3.11, we have

$$\|\overline{\theta}^{\lambda'} - \theta^*\|_1 \leq \frac{25\lambda' s^*}{\widetilde{\rho}_-(s^* + \widetilde{s})} \quad \text{and} \quad \|\theta - \theta^*\|_1 \leq \frac{25\lambda s^*}{\widetilde{\rho}_-(s^* + \widetilde{s})},$$

which further implies

$$(D.19) \qquad \|\theta - \overline{\theta}^{\lambda'}\|_1 \leq \|\theta^* - \theta\|_1 + \|\theta^* - \overline{\theta}^{\lambda'}\|_1 \leq \frac{50(\lambda + \lambda')s^*}{\widetilde{\rho}_-(s^* + \widetilde{s})}.$$

Plugging (D.19) into (D.18), we obtain

$$\mathcal{F}_{\lambda'}(\theta) - \mathcal{F}_{\lambda'}(\overline{\theta}^{\lambda'}) \leq \frac{50(\mathcal{K}_\lambda(\theta) + 3(\lambda - \lambda'))(\lambda + \lambda')s^*}{\widetilde{\rho}_-(s^* + \widetilde{s})}.$$

$\square$

## APPENDIX E: LEMMAS FOR STATISTICAL THEORY

**E.1. Proof of Theorem 3.14.** Before we proceed with the main proof, we first introduce the following lemmas.

**Lemma E.1.** Suppose Assumptions 3.1, 3.5, and 3.7 hold. Then we have

$$\|\widehat{\theta}^{\{N\}} - \theta^*\|_2 = \mathcal{O}\left(\underbrace{\frac{\|\nabla_{\mathcal{S}_1}\mathcal{L}(\theta^*)\|_2}{\widetilde{\rho}_-(s^* + 2\widetilde{s})}}_{V_1} + \underbrace{\frac{\lambda\sqrt{|\mathcal{S}_2|}}{\widetilde{\rho}_-(s^* + \widetilde{s})}}_{V_2} + \underbrace{\frac{\delta_N \lambda \sqrt{s^*}}{\widetilde{\rho}_-(s^* + \widetilde{s})}}_{V_3}\right),$$

where $\mathcal{S}_1 = \{j \mid |\theta_j^*| \geq \gamma\lambda_N\}$ and $\mathcal{S}_2^* = \{j \mid 0 < |\theta_j^*| < \gamma\lambda_N\}$.

The proof of Lemma E.1 is provided in Appendix E.3. Lemma E.1 divides the estimation error of $\widehat{\theta}^{\{N\}}$ into three parts: $V_1$ is the error for strong signals; $V_2$ is the error for weak signals; $V_3$ is the optimization error.

**Lemma E.2.** Suppose Assumption 3.5 holds, $X$ satisfies the column normalization condition, and the observation noise $\varepsilon \sim N(0, \sigma^2 I_n)$ is Gaussian. We then have

$$\mathbb{P}\left(\frac{1}{n}\|X_{*\mathcal{S}_1}^\top \varepsilon\|_2 \geq 3\sigma\sqrt{\frac{\rho_+(|\mathcal{S}_1|) \cdot |\mathcal{S}_1|}{n}}\right) \leq 2\exp\left(-2|\mathcal{S}_1|\right).$$

Lemma E.2 is a direct result of Hanson-Wright inequality (Rudelson and Vershynin, 2013), and therefore its proof is omitted. Lemma E.2 characterizes the large deviation properties of $\|\nabla_{\mathcal{S}_1}\mathcal{L}(\theta^*)\|_2$ in Lemma E.1 for sparse linear regression.

We the proceed with the main proof. For notational simplicity, we omit the index $N$ and denote $\widehat{\theta}^{\{N\}}$, $\lambda_N$, and $\delta_N$ by $\widehat{\theta}$, $\lambda$, and $\delta$ respectively. If we choose a sufficiently small $\delta$ such that $\delta \leq \frac{1}{40\sqrt{s^*}}$, then we apply Lemmas E.1 and E.2, and obtain

$$\|\widehat{\Delta}\|_2 \leq \frac{3\sqrt{|\mathcal{S}_1|}\sigma}{\widetilde{\rho}_-(s^*+2\widetilde{s})}\sqrt{\frac{\rho_+(|\mathcal{S}_1|)\cdot|\mathcal{S}_1|}{n}} + \frac{3\lambda\sqrt{|\mathcal{S}_2|}}{\widetilde{\rho}_-(s^*+2\widetilde{s})} + \frac{0.3\lambda}{\widetilde{\rho}_-(s^*+2\widetilde{s})}.$$

Since all above results rely on Assumptions 3.1 and 3.5, by Lemma 3.13, we have

$$\|\widehat{\Delta}\|_2 \leq \frac{15\sqrt{|\mathcal{S}_1|}\sigma}{\psi_\ell}\sqrt{\frac{\rho_+(|\mathcal{S}_1|)|\mathcal{S}_1|}{n}} + \frac{(96\sqrt{|\mathcal{S}_2|}+10)\sigma}{\psi_\ell}\sqrt{\frac{\log d}{n}}$$

with probability at least $1 - 2\exp(-2\log d) - 2\exp(-2\cdot|\mathcal{S}_1|)$.

### E.2. Proof of Lemma 3.13.

PROOF. By Lemma 7.10, we have

$$(\text{E.1}) \qquad \|\nabla\mathcal{L}(\theta^*)\|_\infty = \|\frac{1}{n}X^\top(y - X\theta^*)\|_\infty = \frac{1}{n}\|X^\top\varepsilon\|_\infty.$$

Since we take $\lambda = 8\sigma\sqrt{\log d/n}$, combining (E.1) with Lemma 7.10, we obtain

$$\mathbb{P}\left(\lambda \geq 4\|\nabla\mathcal{L}(\theta^*)\|_\infty\right) \leq 1 - \frac{2}{d^2}.$$

Moreover, for any $v \in \mathbb{R}^d$ with $\|v\|_0 \leq s$ and $\|v\|_1 \leq \sqrt{s}\|v\|_2$, (3.5) implies

$$(\text{E.2}) \qquad \frac{\|Xv\|_2^2}{n} \geq \psi_\ell\|v\|_2^2 - \gamma_\ell\frac{s\log d}{n}\|v\|_2^2.$$

By simple manipulation, (E.2) implies

$$(\text{E.3}) \qquad \frac{\|Xv\|_2^2}{n} \geq \frac{3\psi_\ell}{4}\|v\|_2^2$$

for $n$ large enough such that $\gamma_\ell\frac{s\log d}{n} \leq \frac{\psi_\ell}{4}$. Similarly, (3.5) implies

$$(\text{E.4}) \qquad \frac{\|Xv\|_2^2}{n} \leq \frac{5\psi_u}{4}\|v\|_2^2$$

for $n$ large enough such that $\gamma_u \frac{s \log d}{n} \leq \frac{\psi_u}{4}$. Since $v$ is an arbitrary sparse vector, for $\alpha \leq \psi_\ell/4$, (E.3) and (E.4) guarantee

$$(\text{E.5}) \qquad \widetilde{\rho}_-(s) = \rho_-(s) - \alpha \geq \psi_\ell/2 \quad \text{and} \quad \rho_+(s) = \rho_-(s) \leq 5\psi_u/4.$$

Let $s = s^* + 2\widetilde{s}$. (E.5) implies

$$484\kappa^2 + 100\kappa \leq 484 \cdot \frac{25\psi_u^2}{4\psi_\ell^2} + 100 \cdot \frac{5\psi_u}{2\psi_\ell}.$$

Then we can choose $C_1$ as $C_1 = 3025 \cdot \frac{\psi_u^2}{\psi_\ell^2} + 250 \cdot \frac{\psi_u}{\psi_\ell}$ such that $\widetilde{s} = C_1 s^* \geq (484\kappa^2 + 100\kappa)s^*$. Meanwhile, we need a large enough $n$ satisfying

$$\frac{\log d}{n} \leq \frac{\psi_\ell}{4\gamma_\ell(s^* + 2C_1 s^*)} \quad \text{and} \quad \frac{\log d}{n} \leq \frac{\psi_u}{4\gamma_u s^* + 2C_1 s^*}.$$

Moreover, we have

$$\lambda_0 = \|\frac{1}{n}Xy\|_\infty \leq \|\frac{1}{n}X^\top X\theta^*\|_\infty + \|\frac{1}{n}X^\top \varepsilon\|_\infty$$

$$\leq \|\frac{1}{n}X^\top X\|_1 \|\theta^*\|_\infty + \mathcal{O}\left(\sigma\sqrt{\frac{\log d}{n}}\right).$$

Given $\|\frac{1}{n}X^\top X\|_1 = \mathcal{O}(d)$ and $\|\theta^*\|_\infty = \mathcal{O}(d)$, for large enough $n$, we have

$$\lambda_0 = \mathcal{O}_P(d^2) \quad \text{and} \quad N = \frac{\log \lambda_0/\lambda_N}{\log \eta} = \mathcal{O}_P\left(\log\left(\frac{d^2}{\sigma}\sqrt{\frac{n}{\log d}}\right)\right) = \mathcal{O}_P(\log d).$$

□

### E.3. Proof of Lemma E.1.

PROOF. For notational simplicity, we omit the index $N$ and denote $\widehat{\theta}^{\{N\}}$, $\lambda_N$, and $\delta_N$ by $\widehat{\theta}$, $\lambda$, and $\delta$ respectively. We define $\widehat{\Delta} = \widehat{\theta} - \theta^*$. Let $\widehat{\xi} \in \partial\|\widehat{\theta}\|_1$ be a subgradient vector satisfying $\mathcal{K}_\lambda(\widehat{\theta}) = \|\nabla\widetilde{\mathcal{L}}_\lambda(\widehat{\theta}) + \lambda\widehat{\xi}\|_\infty \leq \delta\lambda$. Then by the restricted convexity of $\mathcal{F}_\lambda(\theta)$, we have

$$(\text{E.6}) \qquad \mathcal{F}_\lambda(\widehat{\theta}) \geq \mathcal{F}_\lambda(\theta^*) + \widehat{\Delta}^\top (\nabla\widetilde{\mathcal{L}}_\lambda(\theta^*) + \lambda\widetilde{\xi}) + \frac{\widetilde{\rho}_-(s^* + \widetilde{s})}{2}\|\widehat{\Delta}\|_2^2,$$

$$(\text{E.7}) \qquad \mathcal{F}_\lambda(\theta^*) \geq \mathcal{F}_\lambda(\widehat{\theta}) - \widehat{\Delta}^\top (\nabla\widetilde{\mathcal{L}}_\lambda(\widehat{\theta}) + \lambda\widehat{\xi}) + \frac{\widetilde{\rho}_-(s^* + \widetilde{s})}{2}\|\widehat{\Delta}\|_2^2,$$

where $\widetilde{\xi} \in \partial\|\theta^*\|_1$. Combining (E.6) with (E.7), we have

$$\widetilde{\rho}_-(s^* + 2\widetilde{s})\|\widehat{\Delta}\|_2^2 \leq \|\widehat{\Delta}\|_1 \|\nabla\widetilde{\mathcal{L}}_\lambda(\widehat{\theta}) + \lambda\widehat{\xi}\|_\infty - \widehat{\Delta}^\top(\nabla\mathcal{L}(\theta^*) + \nabla\mathcal{H}_\lambda(\theta^*) + \lambda\widetilde{\xi})$$

$$(E.8) \qquad \leq \underbrace{|\widehat{\Delta}^\top(\nabla\mathcal{L}(\theta^*) + \nabla\mathcal{H}_\lambda(\theta^*) + \lambda\widetilde{\xi})|}_{V_0} + \underbrace{\delta\lambda\|\widehat{\Delta}\|_1}_{V4}.$$

[Bounding $V_0$] We consider the following decomposition

$$|\widehat{\Delta}^\top(\nabla\mathcal{L}(\theta^*) + \nabla\mathcal{H}_\lambda(\theta^*) + \lambda\widetilde{\xi})|$$
$$\leq \sum_{\mathcal{A}\in\{\mathcal{S}_1,\mathcal{S}_2,\overline{\mathcal{S}}\}} |\widehat{\Delta}_{\mathcal{A}}^\top(\nabla_{\mathcal{A}}\mathcal{L}(\theta^*) + \nabla_{\mathcal{A}}\mathcal{H}_\lambda(\theta^*) + \lambda\widetilde{\xi}_{\mathcal{A}})|,$$

where $\mathcal{S}_1 = \{j \mid |\theta_j^*| \geq \gamma\lambda\}$ and $\mathcal{S}_2 = \{j \mid 0 < |\theta_j^*| < \gamma\lambda\}$. For $\overline{\mathcal{S}}$, we have $\|\nabla_{\overline{\mathcal{S}}}\mathcal{L}(\theta^*)\|_\infty \leq \lambda/4$ and $\nabla_{\overline{\mathcal{S}}}\mathcal{H}_\lambda(\theta^*) = 0$. Thus, there exists some $\widetilde{\xi}_{\overline{\mathcal{S}}} \in \partial\|\theta_{\overline{\mathcal{S}}}^*\|_1$ such that $\nabla_{\overline{\mathcal{S}}}\mathcal{L}(\theta^*) + \nabla_{\overline{\mathcal{S}}}\mathcal{H}_\lambda(\theta^*) + \lambda\widetilde{\xi}_{\overline{\mathcal{S}}} = 0$, which implies

$$(E.9) \qquad |\widehat{\Delta}^\top(\nabla_{\overline{\mathcal{S}}}\mathcal{L}(\theta^*) + \nabla_{\overline{\mathcal{S}}}\mathcal{H}_\lambda(\theta^*) + \lambda\widetilde{\xi}_{\overline{\mathcal{S}}})| = 0.$$

For all $j \in \mathcal{S}_1$, we have $|\theta_j^*| > \gamma\lambda$ and $|\theta_j|$ is smooth at $\theta_j = \theta_j^*$. Thus, by (R.2) of Lemma B.1, we have $\nabla_{\mathcal{S}_1}\mathcal{H}_\lambda(\theta^*) + \lambda\widetilde{\xi}_{\mathcal{S}_1} = 0$, which implies

$$(E.10) \quad |\widehat{\Delta}_{\mathcal{S}_1}^\top(\nabla_{\mathcal{S}_1}\mathcal{L}(\theta^*) + \nabla_{\mathcal{S}_1}\mathcal{H}_\lambda(\theta^*) + \lambda\widetilde{\xi}_{\mathcal{S}_1})| = |\widehat{\Delta}_{\mathcal{S}_1}^\top \nabla_{\mathcal{S}_1}\mathcal{L}(\theta^*)|$$
$$\leq \|\widehat{\Delta}_{\mathcal{S}_1}\|_2 \|\nabla_{\mathcal{S}_1}\mathcal{L}(\theta^*)\|_2 \leq \|\widehat{\Delta}\|_2 \|\nabla_{\mathcal{S}_1}\mathcal{L}(\theta^*)\|_2.$$

We then consider $\mathcal{S}_2$. Then we have

$$(E.11) \quad |\widehat{\Delta}_{\mathcal{S}_2}^\top(\nabla_{\mathcal{S}_2}\mathcal{L}(\theta^*) + \nabla_{\mathcal{S}_2}\mathcal{H}_\lambda(\theta^*) + \lambda\widetilde{\xi}_{\mathcal{S}_2})|$$
$$\leq \|\widehat{\Delta}_{\mathcal{S}_2}\|_1(\|\nabla_{\mathcal{S}_2}\mathcal{L}(\theta^*)\|_\infty + \|\nabla_{\mathcal{S}_2}\mathcal{H}_\lambda(\theta^*)\|_\infty + \|\lambda\widetilde{\xi}_{\mathcal{S}_2}\|_\infty) \leq 3\lambda\sqrt{|\mathcal{S}_2|}\|\widehat{\Delta}\|_2.$$

Combining (E.9) and (E.10) with (E.11), we have

$$(E.12) \qquad V_0 \leq \|\nabla_{\mathcal{S}_1}\mathcal{L}(\theta^*)\|_2 \|\widehat{\Delta}\|_2 + 3\lambda\sqrt{|\mathcal{S}_2|}\|\widehat{\Delta}\|_2.$$

[Bounding $V_4$] We then proceed to bound $V_4$. Since $\theta$ satisfies the approximate KKT condition, by Theorem 3.11, we have $\|\widehat{\Delta}\|_1 \leq 11\sqrt{s^*}\|\widehat{\Delta}\|_2$. Thus, by (E.12) into (E.8), we have

$$\widetilde{\rho}_-(s^* + \widetilde{s})\|\widehat{\Delta}\|_2^2 \leq \|\nabla_{\mathcal{S}_1}\mathcal{L}(\theta^*)\|_2 \|\widehat{\Delta}\|_2 + 3\lambda\sqrt{|\mathcal{S}_2|}\|\widehat{\Delta}\|_2 + 11\delta\lambda\sqrt{s^*}\|\widehat{\Delta}\|_2.$$

Solving the above inequality, we complete the proof. $\square$

**E.4. Proof of Lemma 7.11.**

PROOF. We then proceed to establish the error bound of the oracle estimator under the $\ell_\infty$ norm. Since Lemma 3.13 guarantees $\rho_-(s) > 0$, (3.7) is a strongly convex problem over $\theta_{\mathcal{S}}$ with a unique optimum

$$\widehat{\theta}^{\text{o}}_{\mathcal{S}} = (X_{*\mathcal{S}}^\top X_{*\mathcal{S}})^{-1} X_{*\mathcal{S}}^\top y. \tag{E.13}$$

Then conditioning on the event $\mathcal{E}_1 = \{\|X^\top \varepsilon\|_\infty / n \leq 2\sigma\sqrt{\log d/n}\}$, we rewrite (E.13) as

$$\|\widehat{\theta}^{\text{o}}_{\mathcal{S}} - \theta^*_{\mathcal{S}}\|_\infty = \|(X_{*\mathcal{S}}^\top X_{*\mathcal{S}})^{-1} X_{*\mathcal{S}}^\top (y - \mathbb{E}y)\|_\infty \tag{E.14}$$

$$= \|(X_{*\mathcal{S}}^\top X_{*\mathcal{S}})^{-1} X_{*\mathcal{S}}^\top \varepsilon\|_\infty \leq \frac{1}{\rho_-(s^*)n} \|X_{*\mathcal{S}}^\top \varepsilon\|_\infty \leq \frac{2\sigma}{\rho_-(s^*)} \sqrt{\frac{\log d}{n}}.$$

Since $\theta^*$ satisfies (3.8), (E.14) implies

$$\min_{j \in \mathcal{S}} |\widehat{\theta}^{\text{o}}_j| = \min_{j \in \mathcal{S}} |\widehat{\theta}^{\text{o}}_j - \theta^*_j + \theta^*_j| \geq \min_{j \in \mathcal{S}} |\theta^*_j| - \|\widehat{\theta}^{\text{o}}_{\mathcal{S}} - \theta^*_{\mathcal{S}}\|_\infty \tag{E.15}$$

$$\geq \left(C_5 \gamma - \frac{2}{\rho_-(s^*)}\right) \sigma \sqrt{\frac{\log d}{n}} \geq \left(C_5 \gamma - \frac{4}{\psi_\ell}\right) \sigma \sqrt{\frac{\log d}{n}},$$

where the last inequality comes from Lemma 3.13. Taking $C_5 = 8 + \frac{4}{\gamma \psi_\ell}$, (E.15) implies

$$\min_{j \in \mathcal{S}} |\widehat{\theta}^{\text{o}}_j| \geq \left(C_5 \gamma - \frac{4}{\psi_\ell}\right) \sigma \sqrt{\frac{\log d}{n}} \geq 8\gamma \sigma \sqrt{\frac{\log d}{n}} = \gamma \lambda,$$

where the last equality comes from $\gamma \geq 4/\psi_\ell$. Then by (R.2) of Lemma B.1, we have

$$\nabla_{\mathcal{S}} \mathcal{H}_\lambda(\widehat{\theta}^{\text{o}}) + \lambda \nabla \|\widehat{\theta}^{\text{o}}_{\mathcal{S}}\|_1 = 0. \tag{E.16}$$

Combining (E.16) with the optimality condition of (3.7), we have

$$\frac{1}{n} X_{*\mathcal{S}}(y - X\widehat{\theta}^{\text{o}}) + \nabla_{\mathcal{S}} \mathcal{H}_\lambda(\widehat{\theta}^{\text{o}}) + \lambda \nabla \|\widehat{\theta}^{\text{o}}_{\mathcal{S}}\|_1 = 0. \tag{E.17}$$

□

**E.5. Proof of Lemma 7.12.**

PROOF. We consider the decomposition

$$
\begin{aligned}
\text{(E.18)} \quad \|X_{*\overline{\mathcal{S}}}^\top(y - X\widehat{\theta}^{\text{o}})\|_\infty &= \|X_{*\overline{\mathcal{S}}}^\top(y - X_{*\mathcal{S}}\widehat{\theta}_{\mathcal{S}}^{\text{o}})\|_\infty \\
&= \|X_{*\overline{\mathcal{S}}}^\top[X_{*\mathcal{S}}\theta_{\mathcal{S}}^* + \varepsilon + X_{*\mathcal{S}}(X_{*\mathcal{S}}^\top X_{*\mathcal{S}})^{-1}X_{*\mathcal{S}}^\top(X_{*\mathcal{S}}\theta_{\mathcal{S}}^* + \varepsilon)]\|_\infty \\
&= \|X_{*\overline{\mathcal{S}}}^\top(I_n - X_{*\mathcal{S}}(X_{*\mathcal{S}}^\top X_{*\mathcal{S}})^{-1}X_{*\mathcal{S}}^\top)\varepsilon\|_\infty \leq \|U_{*\overline{\mathcal{S}}}^\top \varepsilon\|_\infty,
\end{aligned}
$$

where $U = X^\top(I_n - X_{*\mathcal{S}}(X_{*\mathcal{S}}^\top X_{*\mathcal{S}})^{-1}X_{*\mathcal{S}}^\top)$. Conditioning on the event $\mathcal{E}_2 = \{\|U^\top\varepsilon\|_\infty/n \leq 2\sigma\sqrt{\log d/n}\}$, (E.18) implies

$$
\text{(E.19)} \quad \frac{1}{n}\|X_{*\overline{\mathcal{S}}}^\top(y - X\widehat{\theta}^{\text{o}})\|_\infty \leq \frac{\lambda}{4}.
$$

By (R.3) of Lemma B.1, we have $\nabla \mathcal{H}_\lambda(\widehat{\theta}_{\mathcal{S}}^{\text{o}}) = 0$. Since $|\theta_j|$ is non-differentiable at $\theta_j = 0$, then (E.19) implies that there exists some $\widehat{\xi}_{\overline{\mathcal{S}}}^{\text{o}} \in \partial\|\widehat{\theta}_{\overline{\mathcal{S}}}^{\text{o}}\|_1$ such that

$$
\text{(E.20)} \quad \frac{1}{n}X_{*\overline{\mathcal{S}}}^\top(y - X\widehat{\theta}^{\text{o}}) + \nabla_{\overline{\mathcal{S}}}\mathcal{H}_\lambda(\widehat{\theta}^{\text{o}}) + \lambda\widehat{\xi}_{\overline{\mathcal{S}}}^{\text{o}} = 0.
$$

$\square$

## APPENDIX F: EXTENSION TO SPARSE ROBUST REGRESSION

PICASSO can be extended to solve the sparse robust regression problem. The analysis is similar to sparse logistic regression. We need to verify a few slightly different assumptions. Particularly, we denote the response vector by $y = (y_1, ..., y_n)^\top \in \mathbb{R}^n$, and the design matrix by $X \in \mathbb{R}^{n\times d}$. We consider a sparse linear regression model with heavy tail random noise

$$y = X\theta^* + \varepsilon,$$

where $\varepsilon_i$'s are independent sampled from a distribution with $\mathbb{E}\varepsilon_i = 0$ and $\mathbb{E}\varepsilon_i^2 < \infty$. Let $\ell_\zeta$ denote the huber function defined as

$$\ell_\zeta(a) = \frac{a^2}{2} \cdot \mathbb{1}_{\{|a|\leq\zeta\}} + \left(\zeta|a| - \frac{\zeta^2}{2}\right) \cdot \mathbb{1}_{\{|a|>\zeta\}}.$$

When $\theta^*$ is sparse, we consider the optimization problem

$$
\text{(F.1)} \quad \min_{\theta\in\mathbb{R}^d} \mathcal{L}(\theta) + \mathcal{R}_\lambda(\theta), \quad \text{where } \mathcal{L}(\theta) = \frac{1}{n}\sum_{i=1}^n \ell_\zeta(y_i - X_{i*}^\top\theta^*).
$$

For notational simplicity, we denote the huber loss function in (F.1) as $\mathcal{L}(\theta)$, and define $\widetilde{\mathcal{L}}_\lambda(\theta) = \mathcal{L}(\theta) + \mathcal{H}_\lambda(\theta)$. Then similar to sparse linear regression, we also write $\mathcal{F}_\lambda(\theta)$ as

$$\mathcal{F}_\lambda(\theta) = \mathcal{L}(\theta) + \mathcal{R}_\lambda(\theta) = \widetilde{\mathcal{L}}_\lambda(\theta) + \lambda \|\theta\|_1.$$

The huber loss function is differentiable with

$$\nabla \mathcal{L}(\theta) = \frac{1}{n}\sum_{i=1}^n \ell'_\zeta(X_{i*}^\top \theta - y_i) X_{i*}, \ \ell'_\zeta(a) = a \cdot \mathbb{1}_{\{|a| \leq \zeta\}} + \zeta \cdot \text{sign}(a) \cdot \mathbb{1}_{\{|a| > \zeta\}}.$$

Similar to sparse linear regression, we also assume that the design matrix $X$ satisfies the column normalization condition $\|X_{*j}\|_2 = \sqrt{n}$ for all $j = 1, ..., d$.

We apply the proximal coordinate gradient algorithm to solve (F.1). Though the huber loss function is not twice differentiable everywhere, its coordinate gradient is Lipschitz continuous, i.e., for any $\theta$ and $\theta'$, we have

$$|\nabla_j \mathcal{L}(\theta_j, \theta_{\setminus j}) - \nabla_j \mathcal{L}(\theta_j,' \theta_{\setminus j})| \leq \frac{1}{\zeta}|\theta_j - \theta'_j|,$$

which further implies

$$\mathcal{L}(\theta'_j, \theta_{\setminus j}) \leq \mathcal{L}(\theta) + (\theta'_j - \theta_j)\nabla_j \mathcal{L}(\theta) + \frac{1}{2\zeta}(\theta_j - \theta'_j)^2.$$

Thus, similar to sparse logistic regression, if we choose $L = 1/\zeta$, we guarantee

$$\mathcal{Q}_{\lambda, j, L}(\theta_j; \theta^{(t)}) \geq \mathcal{F}_\lambda(\theta_j, \theta^{(t)}_{\setminus j})$$

for all $j = 1, ...d$. Thus, the proximal coordinate gradient algorithm is applicable to sparse robust regression.

We then verify Assumption 3.1 by the following lemma.

**Lemma F.1.** Given $\lambda_N = 16\sqrt{\log d/n}$, we have

$$\mathbb{P}(\lambda_N \geq \|\nabla \mathcal{L}(\theta^*)\|_\infty) \geq 1 - d^{-3}.$$

The proof of Lemma F.1 is provided in Fan et al. (2016), and therefore omitted. Lemma F.1 guarantees that Assumption 3.1 holds with high probability for sparse robust regression.

Different from the logistic loss function, the huber loss function is not twice differentiable everywhere. Therefore, the largest and smallest sparse eigenvalues become invalid when the Hessian matrix does not not exist. To address this issue, Fan et al. (2016) propose a second order approximation method. Particularly, they define the reminder term of the first order approximation of the huber loss function as

$$\mathcal{Z}(\theta', \theta) = \mathcal{L}(\theta') - \mathcal{L}(\theta) - (\theta' - \theta)^\top \nabla \mathcal{L}(\theta).$$

As shown in the proof of Lemma 2 in Fan et al. (2016), they exploit the Lipschitz continuity of $\ell'_\zeta(\cdot)$, and construct twice differentiable functions to approximate $\mathcal{Z}(\theta', \theta)$ from upper and below. Eventually, they characterize the restricted strong smoothness and convexity of the huber loss function by analyzing sparse eigenvalue properties of the Hessian matrices of the twice differentiable approximations. Specifically, they consider a sub-Gaussian random design, and show that given $\|\theta^*\|_0 \leq s^*$, for any $\theta$ satisfying $\|\theta - \theta^*\|_2 \leq R$, with high probability, we have

$$(\text{F.2}) \qquad \frac{\psi_\ell}{2}\|v\|_2^2 - \gamma_\ell \frac{\log d}{n}\|v\|_1^2 \leq \mathcal{Z}(\theta + v, \theta) \leq \frac{\psi_u}{2}\|v\|_2^2 + \gamma_u \frac{\log d}{n}\|v\|_1^2,$$

where $\psi_\ell$, $\psi_u$, $\gamma_\ell$, and $\gamma_u$ are positive constants, and do not scale with $(s^*, n, d)$. Please refer to Fan et al. (2016) for more technical details.

Thus, following similar lines to Lemma 4.2, we can show that given (F.2), the huber loss satisfies the restricted strong convexity and smoothness within a neighborhood of $\theta^*$. This implies that all our following analysis holds, and PICASSO works for sparse robust regression.

## APPENDIX G: OTHER ACTIVE SET UPDATING RULES

We present the other two rules for updating active set: The first one is the randomized selection rule, and the second one is the truncated selection rule. Recall that the iteration index of the middle loop is $[m]$, where $m = 0, 1, 2, \ldots$.

**G.1. Randomized Selection.** At the m-th iteration of the middle loop, we randomly select a coordinate $k_m$ from

$$(\text{G.1}) \qquad \mathcal{M}_{m+0.5} = \{k \mid k \in \overline{\mathcal{A}}_{m+0.5}, \ |\nabla_k \mathcal{L}(\theta^{[m+0.5]})| \geq (1+\delta)\lambda\}$$

with equal probability, where $\delta$ is defined in (2.6). We terminate the IteActUpd algorithm if $\mathcal{M}_{m+0.5}$ is an empty set, i.e., $\mathcal{M}_{m+0.5} = \emptyset$. Otherwise, we obtain $\theta^{[m+1]}$ by

$$\theta^{[m+1]}_{k_m} = \mathcal{T}_{\lambda, k_m}(\theta^{[m+0.5]}) \quad \text{and} \quad \theta^{[m+1]}_{\setminus k_m} = \theta^{[m+0.5]}_{\setminus k_m},$$

and take the new active and inactive sets as

$$\mathcal{A}_{m+1} = \mathcal{A}_{m+0.5} \cup \{k_m\} \quad \text{and} \quad \overline{\mathcal{A}}_{m+1} = \overline{\mathcal{A}}_{m+0.5} \setminus \{k_m\}.$$

We summarize the IteActUpd algorithm using the randomized selection rule in Algorithm 4.

**Remark G.1.** The randomized selection procedure has an equivalent and efficient implementation as follows: (i) We generate a randomly shuffled order of all inactive coordinates of $\overline{\mathcal{A}}_{m+0.5}$; (ii) We then check all inactive coordinates in a cyclic order according to the order generated in (i), and select the first inactive coordinate $k$ satisfying $|\nabla_k \mathcal{L}(\theta^{[m+0.5]})| \geq (1+\delta)\lambda$. If no inactive coordinate satisfies the requirement, i.e.,

$$\max_{k \in \overline{\mathcal{A}}_{m+0.5}} |\nabla_k \mathcal{L}(\theta)^{[m+0.5]}| \leq (1+\delta)\lambda,$$

then we have $\mathcal{M}_{m+0.5} = \emptyset$ and terminate the middle loop.

---

**Algorithm 4:** Similar to the greedy selection rule, the randomized selection rule also moves only one inactive coordinate to the active set in each iteration to encourage the sparsity of the active set.

**Algorithm:** $\widehat{\theta} \leftarrow \mathsf{IteActUpd}(\lambda, \theta^{[0]}, \delta, \tau, \varphi)$
**Initialize:** $m \leftarrow 0$, $\mathcal{A}_0 \leftarrow \{j \mid \theta_j^{[0]} = 0, \ |\nabla_j \mathcal{L}(\theta^{[0]})| \geq (1-\varphi)\lambda\} \cup \{j \mid \theta_j^{[0]} \neq 0\}$
**Repeat**
    $\theta^{[m+0.5]} \leftarrow \mathsf{ActCooMin}(\lambda, \theta^{[m]}, \mathcal{A}_m, \tau)$
    $\mathcal{A}_{m+0.5} \leftarrow \{j \mid \theta_j^{[m+0.5]} \neq 0\}$
    $\overline{\mathcal{A}}_{m+0.5} \leftarrow \{j \mid \theta_j^{[m+0.5]} = 0\}$
    Randomly sample a coordinate $k_m$ from

$$\mathcal{M}_{m+0.5} = \{k \mid k \in \overline{\mathcal{A}}_{m+0.5}, \ |\nabla_k \mathcal{L}(\theta^{[m+0.5]})| \geq (1+\delta)\lambda\}$$

    with equal probability
    $\theta_{k_m}^{[m+1]} \leftarrow \mathcal{T}_{\lambda, k_m}(\theta^{[m+0.5]})$
    $\theta_{\backslash k_m}^{[m+1]} \leftarrow \theta_{\backslash k_m}^{[m+0.5]}$
    $\mathcal{A}_{m+1} \leftarrow \mathcal{A}_{m+0.5} \cup \{k_m\}$
    $\overline{\mathcal{A}}_{m+1} \leftarrow \overline{\mathcal{A}}_{m+0.5} \setminus \{k_m\}$
    $m \leftarrow m + 1$
**Until** $\max_{k \in \overline{\mathcal{A}}_{m+0.5}} |\nabla_k \mathcal{L}(\theta^{[m+0.5]})| \leq (1+\delta)\lambda$
**Output:** $\widehat{\theta} \leftarrow \theta^{[m]}$

---

**G.2. Truncated Cyclic Selection.** At the m-th iteration of the middle loop, we choose to iterate over all coordinates of $\overline{\mathcal{A}}_{m+0.5}$ in a cyclic order. But different from the cyclic selection rule in Friedman et al. (2007); Mazumder et al. (2011), we conduct exact coordinate minimization over an inactive coordinate and add it into the active set only if the corresponding coordinate gradient has a sufficiently large magnitude. Otherwise, we make this coordinate stay in the inactive set.

More specifically, without loss of generality, we assume

$$|\overline{\mathcal{A}}_{m+0.5}| = g \quad \text{and} \quad \overline{\mathcal{A}}_{m+0.5} = \{j_1, ..., j_g\} \subseteq \{1, ..., d\},$$

where $j_1 \leq j_2 \leq ... \leq j_g$. We terminate the IteActUpd algorithm if

$$\max_{k \in \overline{\mathcal{A}}_{m+0.5}} |\nabla_k \mathcal{L}(\theta)^{[m+0.5]}| \leq (1+\delta)\lambda.$$

Otherwise, we construct a sequence of auxiliary solutions $\{w^{[m+1,k]}\}_{k=0}^g$ as follows: For $k = 0$, we set $w^{[m+1,0]} = \theta^{[m+0.5]}$; For $k = 1, ..., g$, we take $w_{\backslash j_k}^{[m+1,k]} = w_{\backslash j_k}^{[m+1,k-1]}$ and

$$w_{j_k}^{[m+1,k]} = \begin{cases} \mathcal{T}_{\lambda,j_k}(w^{[m+1,k-1]}) & \text{if } |\nabla_{j_k} \mathcal{L}(w^{[m+1,k-1]})| \geq (1+\delta)\lambda, \\ w_{j_k}^{[m+1,k-1]} & \text{otherwise}, \end{cases}$$

where $\delta$ is defined in (2.6). Note that when $\delta = 0$, the truncated cyclic selection rule is reduced to the cyclic selection rule in Friedman et al. (2007); Mazumder et al. (2011). Once we obtain all auxiliary solutions, we set $\theta^{[m+1]} = w^{[m+1,g]}$, and take the new active and inactive sets based on the sparsity pattern of $\theta^{[m+1]}$, i.e.,

$$\mathcal{A}_{m+1} = \{j \mid \theta_j^{[m+1]} \neq 0\} \quad \text{and} \quad \overline{\mathcal{A}}_{m+1} = \{j \mid \theta_j^{[m+1]} = 0\}.$$

We summarize the iterative active set updating algorithm using the truncated cyclic selection rule in Algorithm 5.

### G.3. Computational Theory.

**Theorem G.2.** Suppose Assumptions 3.1, 3.5, and 3.7 hold. For any $\lambda \geq \lambda_N$, if the initial solution $\theta^{[0]}$ in Algorithms 4 and 5 satisfies $\|\theta_{\overline{\mathcal{S}}}^{[0]}\|_0 \leq \widetilde{s}$ and $\mathcal{F}_\lambda(\theta^{[0]}) \leq \mathcal{F}_\lambda(\theta^*) + \triangle_\lambda$, then regardless the active set initialized by either the strong rule or simple rule, we have $|\mathcal{A}_0 \cap \overline{\mathcal{S}}| \leq \widetilde{s}$. Meanwhile, for $m = 0, 1, 2, ...$, we have $\|\theta_{\overline{\mathcal{S}}}^{[m]}\|_0 \leq \widetilde{s} + 1$, $\|\theta_{\overline{\mathcal{S}}}^{[m+0.5]}\|_0 \leq \widetilde{s}$. Moreover, for the randomized selection rule, we have

$$\mathbb{E}\mathcal{F}_\lambda(\theta^{[m]}) - \mathcal{F}_\lambda(\bar{\theta}^\lambda) \leq \left(1 - \frac{\widetilde{\rho}_-(s^* + 2\widetilde{s})}{(s^* + 2\widetilde{s})\rho_+(1)}\right)^m [\mathcal{F}_\lambda(\theta^{(0)}) - \mathcal{F}_\lambda(\bar{\theta}^\lambda)];$$

For the truncated cyclic selection rule, we have

$$\mathcal{F}_\lambda(\theta^{[m]}) - \mathcal{F}_\lambda(\theta^{[m-1]}) \leq \frac{\delta^2 \lambda^2}{2\rho_+(1)}.$$

**Algorithm 5:** To encourage the sparsity of the active set, the truncated cyclic selection rule only selects coordinates only when their corresponding coordinate gradients are sufficiently large in magnitude. Without loss of generality, we assume $|\overline{\mathcal{A}}_m| = g$ and $\overline{\mathcal{A}}_m = \{j_1, ..., j_g\}$, where $j_1 \leq j_2 \leq ... \leq j_g$.

**Algorithm:** $\widehat{\theta} \leftarrow \mathsf{IteActUpd}(\lambda, \theta^{[0]}, \delta, \tau, \varphi)$
**Initialize:** $m \leftarrow 0$, $\mathcal{A}_0 \leftarrow \{j \mid \theta_j^{[0]} = 0, \ |\nabla_j \mathcal{L}(\theta^{[0]})| \geq (1-\varphi)\lambda\} \cup \{j \mid \theta_j^{[0]} \neq 0\}$
**Repeat**
$\quad \theta^{[m+0.5]} \leftarrow \mathsf{ActCooMin}(\lambda, \theta^{[m]}, \mathcal{A}_m, \tau)$
$\quad \mathcal{A}_{m+0.5} \leftarrow \{j \mid \theta_j^{[m+0.5]} \neq 0\}$
$\quad \overline{\mathcal{A}}_{m+0.5} \leftarrow \{j \mid \theta_j^{[m+0.5]} = 0\}$
$\quad w^{[m+1,1]} \leftarrow \theta^{[m+0.5]}$
$\quad$ **For** $k \leftarrow 1, ..., g$
$\quad\quad$ **If** $|\nabla_{j_k} \mathcal{L}(w^{[m+1,k-1]})| \geq (1+\delta)\lambda$
$\quad\quad\quad w_{j_k}^{[m+1,k]} \leftarrow \mathcal{T}_{\lambda, j_k}(w^{[m+1,k-1]})$
$\quad\quad$ **Else**
$\quad\quad\quad w_{j_k}^{[m+1,k]} \leftarrow w_{j_k}^{[m+1,k-1]}$
$\quad\quad w_{\setminus j_k}^{[m+1,k]} \leftarrow w_{\setminus j_k}^{[m+1,k-1]}$
$\quad \theta^{[m+1]} \leftarrow w^{[m+1,g]}$
$\quad \mathcal{A}_{m+1} \leftarrow \{j \mid \theta_j^{[m+1]} \neq 0\}$
$\quad \overline{\mathcal{A}}_{m+1} \leftarrow \{j \mid \theta_j^{[m+1]} = 0\}$
$\quad m \leftarrow m + 1$
**Until** $\max_{k \in \overline{\mathcal{A}}_{m+0.5}} |\nabla_k \mathcal{L}(\theta^{[m+0.5]})| \leq (1+\delta)\lambda$
**Output:** $\widehat{\theta} \leftarrow \theta^{[m]}$

Moreover, when we terminate the IteActUpd algorithm with

$$\max_{k \in \overline{\mathcal{A}}_{m+0.5}} |\nabla_k \mathcal{L}(\theta^{[m+0.5]})| \leq (1+\delta)\lambda,$$

the following results hold:

(1) The output solution $\widehat{\theta}^\lambda$ satisfies $\mathcal{K}_\lambda(\widehat{\theta}^\lambda) \leq \delta\lambda$;
(2) For the randomized selection, given a constant $\vartheta \in (0,1)$, the number of active set updating iterations is at most

$$\frac{(s^* + 2\widetilde{s})\rho_+(1)}{\widetilde{\rho}_-(s^* + 2\widetilde{s})} \cdot \log\left(\frac{3\rho_+(1)\left[\mathcal{F}_\lambda(\theta^{(0)}) - \mathcal{F}_\lambda(\bar{\theta}^\lambda)\right]}{\vartheta\delta\lambda}\right)$$

with probability at least $1 - \vartheta$;

(3) For the truncated cyclic selection, given $\delta \geq \sqrt{73/(484\kappa + 100)}$, the number of active set updating iterations is at most

$$\frac{2\rho_+(1)[\mathcal{F}_\lambda(\theta^{[0]}) - \mathcal{F}_\lambda(\bar{\theta}^\lambda)]}{\delta^2 \lambda^2}.$$

Theorem G.2 shows that the randomized selection rule and the truncated cyclic selection attain similar convergence properties to the greedy selection rule. We then present the technical details. We first start with the proof for the randomized selection rule.

PROOF. The proof of the randomized selection rule is similar to that of the greedy selection rule. We first follow similar lines to show $\|\theta_{\overline{\mathcal{S}}}^{[m]}\|_0 \leq \widetilde{s}$ and $\|\theta_{\overline{\mathcal{S}}}^{[m]}\|_0 \leq \widetilde{s} + 1$ for all $m = 0, 1, 2, ...$, since the randomized selection also moves only one inactive coordinate into the active set in each iteration.

For the proximal coordinate gradient descent, we construct the same auxiliary solution $w^{[m+1]} = (w_1^{[m+0.5]}, ..., w_d^{[m+0.5]})^\top$, where

$$w_k^{[m+1]} = \underset{\theta_k}{\operatorname{argmin}}\, \mathcal{Q}_{\lambda,k,L}(\theta_k; \theta^{[m+0.5]}).$$

We define $\mathcal{B} = \{j \mid w_k^{[m+1]} \neq \theta_k^{[m+0.5]}\}$. Similarly, Lemma 7.4 guarantees $|\mathcal{B}| \leq s^* + 2\widetilde{s}$. We then divide all coordinates into three subsets

$$\mathcal{M}_1 = \mathcal{B} \cap \{j \mid j \in \overline{\mathcal{A}}_m,\ |\nabla_j \widetilde{\mathcal{L}}_\lambda(\theta^{[m+0.5]})| \geq (1+\delta)\lambda\},$$
$$\mathcal{M}_2 = \mathcal{B} \cap \{j \mid j \in \overline{\mathcal{A}}_m,\ |\nabla_j \widetilde{\mathcal{L}}_\lambda(\theta^{[m+0.5]})| \leq (1+\delta)\lambda\},$$
$$\mathcal{M}_3 = \mathcal{B} \cap \{j \mid j \in \mathcal{A}_m\}.$$

Following similar lines to the proof of Lemma C.4, we have

$$\max_{k \in \mathcal{M}_1} \mathcal{Q}_{\lambda,k,L}(w_k^{[m+1]}; \theta^{[m+0.5]}) \leq \min_{j \in \mathcal{M}_2 \cup \mathcal{M}_3} \mathcal{Q}_{\lambda,j,L}(w_j^{[m+1]}; \theta^{[m+0.5]}),$$

which implies that

$$(\text{G.2}) \quad \sum_{j \in |\mathcal{B}|} \mathcal{Q}_{\lambda,j,L}(w_j^{[m+1]}; \theta^{[m+0.5]}) \geq \frac{|\mathcal{B}|}{|\mathcal{M}_1|} \sum_{k \in |\mathcal{M}_1|} \mathcal{Q}_{\lambda,k,L}(w_k^{[m+1]}; \theta^{[m+0.5]}).$$

Then conditioning on $\theta^{[m+0.5]}$, we have

$$(\text{G.3}) \quad \mathbb{E}\mathcal{F}_\lambda(\theta^{[m+1]})|\theta^{[m+0.5]} = \frac{1}{|\mathcal{M}_1|} \sum_{k \in \mathcal{M}_1} \mathcal{F}_\lambda(w_k^{[m+1]}, \theta_{\setminus k}^{[m+0.5]})$$
$$\leq \frac{1}{|\mathcal{M}_1|} \sum_{k \in |\mathcal{M}_1|} \mathcal{Q}_{\lambda,k,L}(w_k^{[m+1]}; \theta^{[m+0.5]}) \leq \frac{1}{|\mathcal{B}|} \sum_{j \in |\mathcal{B}|} \mathcal{Q}_{\lambda,j,L}(w_j^{[m+1]}; \theta^{[m+0.5]}),$$

where the last inequality comes from (G.2). By rearranging (G.3), we have

$$\mathcal{F}_\lambda(\theta^{[m+0.5]}) - \mathbb{E}[\mathcal{F}_\lambda(\theta^{[m+1]})|\theta^{[m+0.5]}] \geq \frac{1}{s^* + 2\widetilde{s}}[\mathcal{J}_{\lambda,L}(w^{[m+1]}; \theta^{[m+0.5]}) - \mathcal{F}_\lambda(\theta^{[m+0.5]})],$$

where $\mathcal{J}_{\lambda,L}(w^{[m+1]}; \theta^{[m+0.5]})$ is defined in Appendix C.12. We then follow similar lines to show

$$(G.4) \quad \mathbb{E}\mathcal{F}_\lambda(\theta^{[m+1]}) - \mathcal{F}_\lambda(\bar{\theta}^\lambda) \leq \left(1 - \frac{\widetilde{\rho}_-(s^* + 2\widetilde{s})}{(s^* + 2\widetilde{s})L}\right)[\mathcal{F}_\lambda(\theta^{[m]}) - \mathcal{F}_\lambda(\bar{\theta}^\lambda)].$$

For the exact coordinate minimization, we construct a similar auxiliary solution $\theta^{[m+0.75]}$ obtain by the proximal coordinate gradient descent using $L = \rho_+(1)$, then follow similar lines to show

$$(G.5) \quad \mathbb{E}\mathcal{F}_\lambda(\theta^{[m+1]}) - \mathcal{F}_\lambda(\bar{\theta}^\lambda) \leq \left[\mathbb{E}\mathcal{F}_\lambda(\theta^{[m+0.75]}) - \mathcal{F}_\lambda(\bar{\theta}^\lambda)\right]$$

$$\leq \left(1 - \frac{\widetilde{\rho}_-(s^* + 2\widetilde{s})}{(s^* + 2\widetilde{s})\rho_+(1)}\right)\left[\mathcal{F}_\lambda(\theta^{[m]}) - \mathcal{F}_\lambda(\bar{\theta}^\lambda)\right].$$

Combine (G.4) with (G.5) and taking the expectation over $m = 0, 1, 2, ...$, we obtain

$$(G.6) \quad \frac{\mathbb{E}\mathcal{F}_\lambda(\theta^{[m+1]}) - \mathcal{F}_\lambda(\bar{\theta}^\lambda)}{\mathcal{F}_\lambda(\theta^{[m]}) - \mathcal{F}_\lambda(\bar{\theta}^\lambda)} \leq \left(1 - \frac{\widetilde{\rho}_-(s^* + 2\widetilde{s})}{(s^* + 2\widetilde{s})\rho_+(1)}\right)^m.$$

The iteration complexity can also be derived in a similar fashion. When we have

$$(G.7) \quad \mathcal{F}_\lambda(\theta^{[m]}) - \mathcal{F}_\lambda(\bar{\theta}^\lambda) \leq \frac{\delta^2 \lambda^2}{3\nu_+(1)},$$

we must have $|\nabla_{k_m} \widetilde{\mathcal{L}}_\lambda(\theta^{[m+0.5]})| \leq (1+\delta)\lambda$. That implies that the algorithm must terminate when (G.7) holds. Applying the Markov inequality to (G.6), we have

$$\mathbb{P}\left(\mathcal{F}_\lambda(\theta^{[m]}) - \mathcal{F}_\lambda(\bar{\theta}^\lambda) \geq \frac{\delta^2 \lambda^2}{3\nu_+(1)}\right) \leq \frac{3\nu_+(1)}{\delta^2 \lambda^2}[\mathbb{E}\mathcal{F}_\lambda(\theta^{[m]}) - \mathcal{F}_\lambda(\bar{\theta}^\lambda)]$$

$$\leq \frac{3\nu_+(1)}{\delta^2 \lambda^2}\left(1 - \frac{\widetilde{\rho}_-(s^* + 2\widetilde{s})}{(s^* + 2\widetilde{s})\nu_+(1)}\right)^m \left[\mathcal{F}_\lambda(\theta^{(0)}) - \mathcal{F}_\lambda(\bar{\theta}^\lambda)\right] \leq \vartheta.$$

By simple manipulation, we need

$$m \geq \log^{-1}\left(1 - \frac{\widetilde{\rho}_-(s^* + 2\widetilde{s})}{(s^* + 2\widetilde{s})\nu_+(1)}\right) \log\left(\frac{\vartheta \delta^2 \lambda^2}{3\nu_+(1)\left[\mathcal{F}_\lambda(\theta^{(0)}) - \mathcal{F}_\lambda(\bar{\theta}^\lambda)\right]}\right)$$

iterations such that

$$\mathbb{P}\left(\mathcal{F}_\lambda(\theta^{[m]}) - \mathcal{F}_\lambda(\bar{\theta}^\lambda) \leq \frac{\delta^2\lambda^2}{3\nu_+(1)}\right) \geq 1 - \vartheta.$$

$\square$

We then proceed with the proof for the truncated cyclic selection rule.

PROOF. To guarantee the sparsity of the active set, we need to show that the truncated cyclic selection moves no more than $\widetilde{s}$ inactive coordinates to the active set in each active set updating step. We prove this by contradiction.

We assume that the truncated cyclic selection adds exactly $\widetilde{s}+1$ inactive coordinates into the active set, and obtain a new active set $\mathcal{A}_{m+1}$. We define an auxiliary set $\mathcal{B} = \mathcal{A}_{m+1} \cup \mathcal{S}$. Since the objective value always decreases in each middle loop, we have

(G.8) $$\mathcal{F}_\lambda(\theta^{[m+1]}) \leq \mathcal{F}_\lambda(\theta^*) + \frac{4\lambda^2 s^*}{\widetilde{\rho}_-(s^* + \widetilde{s})}.$$

We then define $w^{[m+1]}$ as a local optimum to the following optimization problem,

(G.9) $$\min_{\theta \in \mathbb{R}^d} \mathcal{F}_\lambda(\theta) \quad \text{subject to} \quad \theta_{\overline{\mathcal{B}}} = 0.$$

By Assumption 3.5, we know that (G.9) is a strongly convex optimization problem. Therefore $w^{[m+1]}$ is a unique global optimum. Moreover, since $|\mathcal{B} \cap \overline{\mathcal{S}}| \leq 2\widetilde{s} + 1$, by Lemma C.2, we have

$$\|w^{[m+1]} - \theta^*\|_1 \leq \frac{25\lambda s^*}{\widetilde{\rho}_-(s^* + 2\widetilde{s} + 1)}.$$

By the restricted strong convexity of $\mathcal{F}_\lambda(\theta)$, for any $\xi \in \partial \|\theta^*\|_1$ such that

$$\mathcal{F}_\lambda(\theta^*) - \mathcal{F}_\lambda(w^{[m+1]}) \leq -(w^{[m+1]} - \theta^*)^\top (\nabla \widetilde{\mathcal{L}}_\lambda(\theta^*) + \lambda\xi)$$
$$\stackrel{(i)}{\leq} \|w^{[m+1]} - \theta^*\|_1 \cdot \|\nabla \widetilde{\mathcal{L}}_\lambda(\theta^*) + \lambda\xi\|_\infty$$
(G.10) $$\stackrel{(ii)}{\leq} \frac{25\lambda s^*}{\widetilde{\rho}_-(s^* + 2\widetilde{s} + 1)}\left(\frac{\lambda}{4} + \lambda\right) \leq \frac{125\lambda^2 s^*}{4\widetilde{\rho}_-(s^* + 2\widetilde{s} + 1)},$$

where (i) comes from Hölder's inequality, and (ii) comes from Assumption 3.1 and the fact $\|\xi\|_\infty \leq 1$. Since $w^{[m+1]}$ is the global optimum to (G.9),

(G.10) further implies

$$\mathcal{F}_\lambda(\theta^*) \leq \mathcal{F}_\lambda(w^{[m+1]}) + \frac{125\lambda^2 s^*}{4\widetilde{\rho}_-(s^* + 2\widetilde{s} + 1)} \tag{G.11}$$

$$\leq \mathcal{F}_\lambda(\theta^{[m+1]}) + \frac{125\lambda^2 s^*}{4\widetilde{\rho}_-(s^* + 2\widetilde{s} + 1)}.$$

Combining (G.11) with (G.8), we have

$$\mathcal{F}_\lambda(\theta^{[m]}) \leq \mathcal{F}_\lambda(\theta^{[m+1]}) + \frac{36\lambda^2 s^*}{\widetilde{\rho}_-(s^* + 2\widetilde{s} + 1)}.$$

Since the truncated cyclic selection only selects an inactive coordinate when its corresponding coordinate gradient is sufficiently large in magnitude, by Lemma 7.8, we know that adding $\widetilde{s} + 1$ inactive coordinates leads to

$$\mathcal{F}_\lambda(\theta^{[m+1]}) \leq \mathcal{F}_\lambda(\theta^{[m+0.5]}) \leq \mathcal{F}_\lambda(\theta^{[m]}) - \frac{(\widetilde{s} + 1)\delta^2 \lambda^2}{2\nu_+(1)}.$$

Combining the above results, we have

$$\widetilde{s} \leq \frac{72\nu_+(1) s^*}{\delta^2 \widetilde{\rho}_-(s^* + 2\widetilde{s}) + 1} \leq \frac{72\kappa}{\delta^2} \cdot s^*. \tag{G.12}$$

Thus if we have $\delta \geq \sqrt{73/(484\kappa + 100)}$, then (G.12) implies $\widetilde{s} < (484\kappa^2 + 100\kappa)s^*$, which is contradicted by Assumption 3.5. Thus the truncated cyclic selection cannot add more than $\widetilde{s}$ inactive coordinates into the active set.

We then proceed to bound the empirical iteration complexity. Since the algorithm guarantees the solution sparsity, i.e., $\|\theta_{\overline{S}}\|_0 \leq \widetilde{s}$, by the restricted strong convexity of $\mathcal{F}_\lambda(\theta)$, we have

$$\mathcal{F}_\lambda(\theta) - \mathcal{F}_\lambda(\bar{\theta}^\lambda) \geq (\theta - \bar{\theta}^\lambda)^\top (\nabla \widetilde{\mathcal{L}}_\lambda(\bar{\theta}^\lambda) + \lambda \xi) + \frac{\rho_-(s^* + 2\widetilde{s})}{2} \|\theta - \bar{\theta}^\lambda\|_2^2 \geq 0,$$

where $\xi \in \partial \|\bar{\theta}^\lambda\|_1$ satisfies the optimality condition of (1.1), i.e.,

$$\nabla \widetilde{\mathcal{L}}_\lambda(\bar{\theta}^\lambda) + \lambda \xi = 0.$$

Thus we know that $\mathcal{F}_\lambda(\theta)$ is always lower bounded by $\mathcal{F}_\lambda(\bar{\theta}^\lambda)$. Moreover, by Lemma 7.8, we have

$$\mathcal{F}_\lambda(\theta^{[m]}) - \mathcal{F}_\lambda(\theta^{[m+1]}) \geq \frac{\delta^2 \lambda^2}{2\nu_+(1)}.$$

Therefore, the number of the active set updating iterations is at most

$$\frac{2\nu_+(1)[\mathcal{F}_\lambda(\theta^{[0]}) - \mathcal{F}_\lambda(\bar{\theta}^\lambda)]}{\delta^2 \lambda^2}.$$

$\square$

To guarantee the truncated cyclic selection rule to output solutions with similar precisions to the other selection rules, our theoretical analysis needs $\kappa$ to linearly scale with $\sqrt{s^*}$. This eventually leads to a slightly worse sample complexity than the greedy and randomized selection rules. We suspect that this is an artifact of our proof, since we do not observe such a difference of the sample complexity among all three selection rules in our numerical experiments.

Theorem G.2 can be further integrated with Theorem 3.12. Since the extension is straight forward, we omit the details.

Tuo Zhao  
School of Industrial and Systems Engineering  
Georgia Institute of Technology  
755 Ferst Drive NW  
Atlanta, GA 30332, US  
E-mail: tourzhao@gatech.edu

Han Liu  
Department of Operations Research  
and Financial Engineering  
Sherred Hall 224, Princeton University  
Princeton, NJ 08544, US  
E-mail: hanliu@princeton.edu

Tong Zhang  
Tencent AI Lab  
Shennan Ave, Nanshan District  
Shen Zhen, Guangdong Province 518057, China  
E-mail: tzhang@stat.rutgers.edu